\PassOptionsToPackage{table}{xcolor}
\documentclass[draft]{agujournal2019}
\usepackage{amssymb} 
\usepackage{array}
\usepackage{booktabs}

\usepackage{amsmath}
\usepackage{textcomp}
\usepackage{makecell}
\usepackage{dsfont}
\usepackage{tikz}
\usetikzlibrary{positioning,shapes,fit,backgrounds}
\newcolumntype{R}[2]{%
    >{\adjustbox{angle=#1,lap=\width-(#2)}\bgroup}%
    l%
    <{\egroup}%
}
\usepackage{url} 
\usepackage{soul}
\usepackage{float}
\usepackage{xr}
\usepackage{adjustbox}
\usepackage{makecell}
\usepackage{pdfpages}

\usepackage{amssymb} 
\usepackage{array}
\usepackage{amsmath}
\usepackage{textcomp}
\newcolumntype{R}[2]{%
    >{\adjustbox{angle=#1,lap=\width-(#2)}\bgroup}%
    l%
    <{\egroup}%
}
\usepackage[inline]{trackchanges} 
\usepackage{lipsum}
\usepackage{enumitem}
\setlist{
  listparindent=\parindent,
  parsep=0pt,
}

\draftfalse

\journalname{Water Resources Research}

\begin{document}

%
%

\title {Evaluating Deep Learning Approaches for Predictions in Unmonitored Basins with Continental-scale Stream Temperature Models}
\authors{Jared D. Willard\affil{1}, Fabio Ciulla\affil{2}, Helen Weierbach\affil{2}, Vipin Kumar\affil{3}, Charuleka Varadharajan\affil{2}}

\affiliation{1}{Computing Sciences Area, Lawrence Berkeley National Laboratory}
\affiliation{2}{Earth and Environmental Sciences Area, Lawrence Berkeley National Laboratory}
\affiliation{3}{University of Minnesota Department of Computer Science and Engineering}





\correspondingauthor{Jared Willard}{jwillard@lbl.gov}
\correspondingauthor{Charuleka Varadharajan}{cvaradharajan@lbl.gov}





\begin{keypoints}
\item Top-down deep learning models outperform bottom-up and grouped methods for predicting daily stream temperatures in unmonitored basins. 
\item Streamflow is not necessary for accurate stream temperature prediction, as models using meteorology and site attributes perform comparably. 
\item Sites impacted by deep groundwater and dams have greater errors compared to sites where stream temperatures are atmospherically-driven.  
\end{keypoints}

%
%

%
%


\begin{abstract}
The prediction of streamflows and other environmental variables in unmonitored basins is a grand challenge in hydrology. Recent machine learning (ML) models can harness vast datasets for accurate predictions at large spatial scales. However, there are open questions regarding model design and data needed for inputs and training to improve performance. This study explores these questions while demonstrating the ability of deep learning models to make accurate stream temperature predictions in unmonitored basins across the conterminous United States. First, we compare top-down models that utilize data from a large number of basins with bottom-up methods that transfer ML models built on local sites, reflecting traditional regionalization techniques. We also evaluate an intermediary grouped modeling approach that categorizes sites based on regional co-location or similarity of catchment characteristics. Second, we evaluate trade-offs between model complexity, prediction accuracy, and applicability for more target locations by systematically removing inputs. We then examine model performance when additional training data becomes available due to reductions in input requirements. Our results suggest that top-down models significantly outperform bottom-up and grouped models. Moreover, it is possible to get acceptable accuracy by reducing both dynamic and static inputs enabling predictions for more sites with lower model complexity and computational needs. From detailed error analysis,  we determined that the models are more accurate for sites primarily controlled by air temperatures compared to locations impacted by groundwater and dams. By addressing these questions, this research offers a comprehensive perspective on optimizing ML model design for accurate predictions in unmonitored regions.

\end{abstract}

\section*{Plain Language Summary}
Predictions of river flows and water quality are needed globally to address increasing demands from population growth, land use change and projected impacts of climate change. However, many watersheds lack sufficient observations needed to build models, making accurate predictions difficult. Recent advances in machine learning (ML) have shown that more accurate predictions for environmental variables can be made by using large, diverse datasets. In this work, we use stream temperature as a case study to demonstrate the ability of ML models to make large-scale predictions in regions with no (or sparse) monitoring data. We found that "top-down" models trained on large amounts of data from different locations were the most accurate, compared to those built from smaller or more localized regional datasets. In particular these top-down models outperform traditional "bottom-up" approach where models are built at a monitored sites and transferred to similar sites.  Simpler models using only widely available meteorological data also performed well and can be used broadly at any location. However, predictions were more difficult for sites with dam or groundwater influence. Overall, our findings suggest that deep learning models using large-scale datasets can be effectively used to make widespread, accurate predictions of stream temperatures in unmonitored regions.


\addcontentsline{toc}{section}{Introduction}
\section{Introduction}
\label{sec:intro}

Accurate, spatially-distributed predictions of stream hydrology and water quality can provide essential information for sustainable water resources management \cite{chen2021integrating,dwivedi2022legacy}. Such predictions are needed to understand, plan for, and respond to the effects of anthropogenic stressors, global warming, and extreme events that are occurring with increasing frequency and intensity \cite{best2019anthropogenic,van2023global}. The prediction of river flows and other critical environmental variables in unmonitored basins has been a grand challenge in hydrology for decades \cite{sivapalan2003iahs,hrachowitz2013decade,guo2021regionalization,xiong_predicting_2022,rahmani2021deep}.  This challenge exists because only a small number of stream reaches around the world are monitored for important hydrological variables \cite{read_water_2017,seibert2009gauging}, and has persisted despite numerous community efforts to address the problem for several reasons. These include the complexity and variability of hydrological processes across catchments with diverse characteristics \cite{kannan2019some}, data sparsity \cite{schuol2006calibration,ang2022evaluation,kapangaziwiri2009towards}, an emphasis on modeling approaches based on transfer of knowledge from a few intensively monitored locations \cite{parajka2005comparison,zhang2009evaluation}, and computational limitations inhibiting the ability to run process-rich models at large scales \cite{azmi2021bit,vema2020towards}.  

Traditional approaches for prediction in ungauged basins (PUBs) involve the concept of "regionalization", wherein localized, process-based or statistical models are built at data-rich locations, and their parameters extrapolated to similar ungauged regions \cite{Bloschl1995Apr, he2011review,merz2004regionalisation}. In these "bottom-up" approaches, different methods have been used for regionalization that all use explicit measures of similarity such as those based on biophysical catchment characteristics, spatial proximity (i.e., using physical distance) or hydrological signatures \cite{guo_regionalization_2020} to decide which model or parameter set is best to transfer. Some studies have also attempted "grouped" approaches where models are trained or calibrated on a collection of similar or co-located sites prior to regionalization \cite{rao2006regionalization_fuzz,rao2006regionalization_hyb,hu2024parameter,bock2016parameter}. Recently, there have been significant breakthroughs in the PUBs challenge, with machine learning (ML) models showing remarkable promise at making accurate, large-scale predictions at unmonitored sites for various hydrological variables such as stream flows \cite{choi2022utilization,feng2021mitigating,koch2022long,kratzert2019toward_ung} including floods \cite{Nearing2024Mar}, evapotranspiration \cite{chen2020estimating}, stream and lake temperatures \cite{rahmani2021deep,tayal2022invertibility}, and other stream water quality variables \cite{kalin2010predicting, xiong2022predicting, zhi2021hydrometeorology}. These "top-down" models leverage long-term data from hundreds of monitored sites, often from diverse regions \cite{kratzert2019toward_ung,fang_data_2022,rahmani2021deep}, and have even been demonstrated to outperform process models calibrated for the catchments of comparison \cite{kratzert_toward_2019,chen2023improving,yin_runoff_2021}. A majority of ML studies utilize the long short-term memory (LSTM) neural network \cite{hochreiter_long_1997} due to its ability to represent system memory, with a set of inputs that comprise dynamic variables (e.g., meteorology) and a small number of biophysical characteristics that are treated as static variables to provide information for the model to distinguish between different catchment types \cite{willard2023water}.

An emerging debate in hydrology is whether advances in ML should effect a fundamental change in modeling philosophy shifting from a "bottom-up" to a "top-down" approach for applications that seek to maximize predictive accuracy \cite{kratzert2024hess}. Here, we use the same definitions of these modeling approaches (specified above) as in \citeA{kratzert2024hess} and not in the process modeling context as specified in \citeA{hrachowitz_decade_2013}, who referred to "top-down" approaches as spatially lumped, conceptual bucket models versus "bottom-up" as spatially distributed, continuum, physically based models. Recent studies have shown that deep learning models in general benefit from larger, more diverse training datasets in monitored (especially gauged) scenarios, where training and testing is done on the same set of sites by splitting the data into separate time periods. For instance, \citeA{fang2022data} found that deep learning models for streamflow and soil moisture in gauged basins achieve better performance when trained on diverse datasets from multiple regions compared to training on a homogeneous dataset from a single region, even when the latter is more hydrologically similar to the test conditions and the datasets are of equal size. The results from \citeA{kratzert2024hess} also suggest that LSTM models, at least for stream flow prediction in the United States, should never be trained on a single basin and using all available data should be the default approach. Both studies have demonstrated that training on smaller groupings of sites both by region and attribute similarity (determined using a k-means clustering) results in significantly worse performance than the top-down model, implying that deep learning models benefit from seeing more diverse, large datasets. 

Given recent advancements in ML for hydrological modeling, an important question that remains unresolved for unmonitored scenarios is how "bottom-up" approaches based on regionalization of site-specific models that rely on calculated measures of similarity or transferability compare with "top-down" deep learning (DL) models that presumably learn to generalize by seeing diversity within large-sample data from monitored sites. Moreover, there are other open questions to address regarding the implementation of DL models for predictions of dynamic hydrologic variables in unmonitored basins \cite{willard2023water}. First, the selection of input variables is typically done by domain experts without a rigorous process for feature selection, unlike in other applications of ML time series regression where feature selection is standard \cite{meisenbacher2022review,li2017feature}. For example, there are hundreds of catchment characteristics that are now available as large-scale data products \cite{falcone_gages-ii_2011,hill2016stream,linke2019global}, yet only a handful (typically 20-30) are used in hydrologic ML models without justification for feature selection.  Systematically evaluating these features and their representation could significantly improve the accuracy and applicability of predictive models by using techniques that address redundancy and relevance, such as those highlighted in \citeA{ciulla2023network}.  Second, imposing additional data requirements may limit the applicability of the model if those inputs are not available. For example, while some data such as meteorology are widely available at most locations particularly in the form of gridded data products, models that require inputs such as catchment characteristics that are not broadly available or co-located hydrologic variables such as stream flows and temperatures will have substantially lower number of sites available for training. Finally, most studies only use accuracy as the sole metric of model performance; however trade-offs between increasing complexity and data requirements versus accuracy need to be evaluated for optimal model choice \cite{varadharajan_can_2022}. 

Our objectives for this study are to address these knowledge gaps using a set of ML-based model experiments for predictions in unmonitored sites. A primary goal is to determine the best modeling approach by investigating whether models should be trained on large, diverse datasets covering wide geographic areas (top-down approach), focus on hydrologically-similar or co-located subsets (grouped approach), or use a bottom-up approach where site-specific models built for data-rich locations are regionalized to unmonitored locations using transfer learning. Another key objective is to identify which input variables are most crucial for accurate predictions. We aim to understand whether models need a broad range of catchment characteristics, or if they can achieve high accuracy with a more streamlined set of inputs. We also explore how the availability of different types of input data (e.g., meteorology, discharge, catchment characteristics) affects model performance. This involves exploring the trade-offs between requiring comprehensive input data versus utilizing a larger training dataset with fewer input constraints.

In this work, we have built several models to predict daily stream temperatures in unmonitored sites as a case study to demonstrate these objectives.  Water temperature is an important variable in stream ecosystems, influencing water chemistry and aquatic life \cite{brett1970environmental}. It is typically a monitored and regulated parameter for water resource management. However, observations are relatively sparse and an order of magnitude less available in comparison to stream flows \cite{normand2021us,usgs2016waterdata}, making it a good candidate for these research objectives. Factors influencing stream temperatures include climate (particularly air temperatures and solar radiation),  stream flows, snowmelt and surface runoff, surface-groundwater interactions,  riparian shading, and human activities (e.g., discharge from thermal power plants, dam releases) \cite{kedra2018climatic,zhang2020river,kelleher2012investigating,borman2003case}. A myriad of anthropogenic stressors---ranging from urbanization \cite{somers2013streams} and dam installations \cite{risley2010effects} to agriculture \cite{essaid2017evaluating} ---are reshaping established stream thermal regimes. These changes will be exacerbated due to the consequences of climate change including rising air temperatures, changing stream flow patterns, and an increase in intensity and frequency of extreme events such as heatwaves and droughts  \cite{ficklin2013effects,van_vliet_global_2013}. The ability to accurately predict stream temperatures is vital for the management of river ecosystem services including fisheries, hydroelectric power generation as well as management of aquatic habitats \cite{wilby2010evidence,van2016impacts,hansen2015learning}.

Ultimately, our goal is to enable accurate predictions of stream temperatures in unmonitored sites using broadly available input data, ensuring the models' applicability across diverse regions and conditions.   Hence, our study uses data from both pristine and human-impacted sites, in contrast to a vast majority of hydrological machine learning models that focus only on pristine sites \cite{willard2023water}.  Here, we aim to predict the entire stream temperature time series at a target location, which is distinct from making temporal predictions at monitored sites or from temporal forecasting applications, where past observations can be used to build a predictive model for future.  
By evaluating the best approaches for building accurate models and understanding the loss of accuracy when certain data are unavailable, we aim to provide insight into the design of practical ML models for predictions in unmonitored regions and for eventual use in water resource management. Additionally, we seek to identify the conditions under which these models are applicable to use given acceptable error thresholds. While this study focuses on stream temperature modeling, our methods and findings are broadly applicable for other hydrologic variables.

To our knowledge, this study is the first that compares a bottom-up regionalization approach to both top-down and grouped ML models. We use novel techniques for this comparison, namely a meta transfer learning framework \cite{willard_predicting_2021} for bottom-up modeling as well as a network-based method to cluster sites by attribute similarity, which has been demonstrated to outperform unsupervised k-means and hierarchical clustering \cite{ciulla2023network}. The network-based similarity method also serves to reduce model complexity by representing an extensive set of 274 catchment attributes with a smaller set of ~25 highly interpretable attribute categories that have minimal redundancy. We are also the first to examine trade-offs in data availability to enhance model generalizability. For this, we compare input data requirements that are more comprehensive resulting in fewer sites that have the necessary data versus requiring less inputs whose data are broadly available for more target locations.  Overall, our findings complement other large-sample deep learning studies that advocate for a shift in modeling philosophy towards  large-scale data-driven approaches, moving beyond the traditional method of regionalization of  localized models, for the purpose of making the most accurate spatiotemporal predictions in unmonitored basins.

\section{Model Experiment Overview}
\label{sec:model_exp_overview}
We conduct three experiments to independently test the effects of different modeling choices for prediction of daily stream temperatures. For this, we utilize a large dataset comprising over 4 million stream temperature observations in the continental United States (CONUS) at 1362 pristine and human-impacted sites (Section \ref{subsec:methods:data}). We use the LSTM architecture as the primary model for all experiments both because it is the most popular ML model used for hydrological time series prediction in unmonitored sites \cite{willard2023water}, and also to maintain a fair comparison of models across the different experiments. 

\begin{itemize}
\item \textbf{Experiment 1: Comparing top-down, bottom-up, and grouped approaches} 

This experiment compares four different approaches for modeling PUBs, including top-down ML models, along with ML analogs of traditional bottom-up and grouped approaches used in hydrological modeling (Figure \ref{fig:exp1_diagram}). For the top-down model, we follow the approach used in prior studies (e.g., \cite{kratzert2019toward_ung,wang2022exploring,nogueira2022deep,zhi2021hydrometeorology} using observations from diverse sites across the conterminous United States as training data. For the grouped models, we use two approaches. The first uses similarity based on spatial proximity, with groups defined as the 18 regions within the United States Geological Surveys Hydrologic Unit Codes (HUCs) categorization, which has been used in previous studies \cite{kratzert2024hess,wang2022exploring}. The second is based on similarity of catchment attributes as determined by a network-based approach that has been demonstrated to outperform other unsupervised clustering methods for multiple metrics \cite{ciulla2023network}. 

\begin{figure}[h!]
    \centering
    \includegraphics[width=\textwidth]{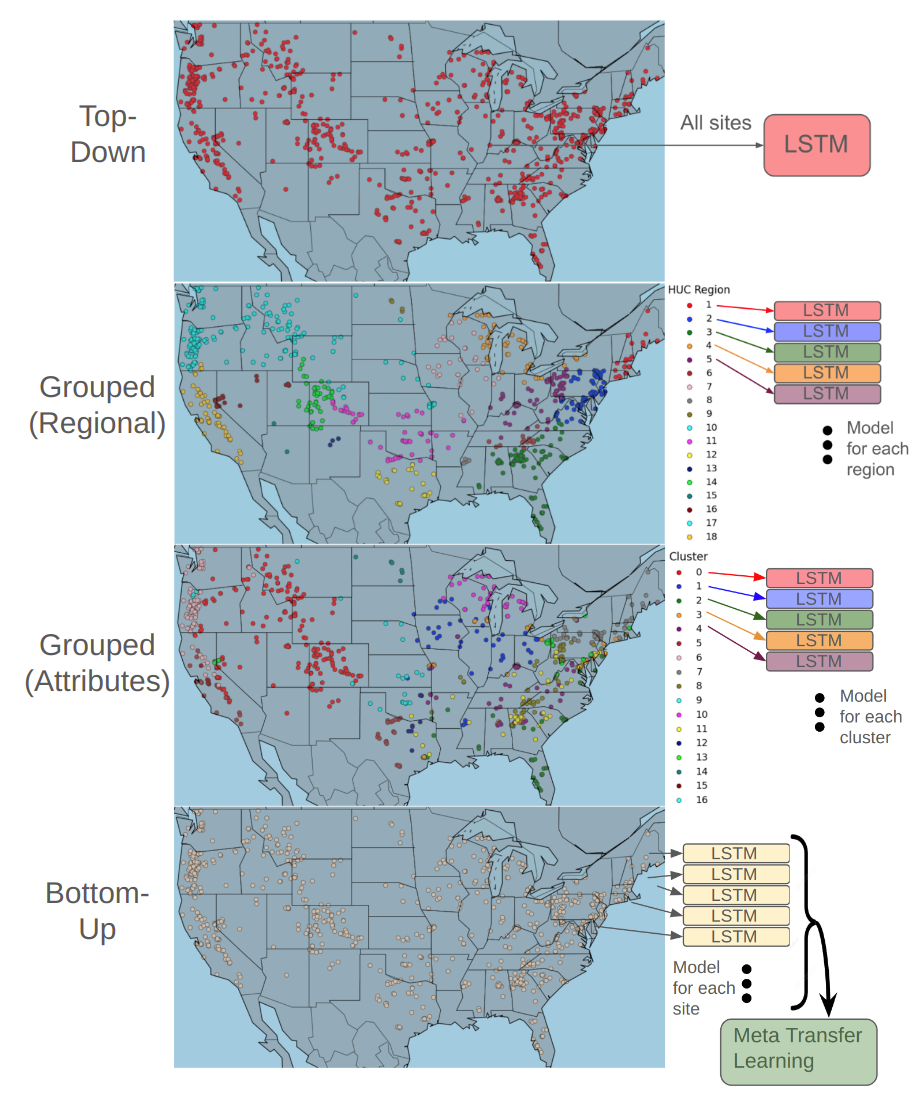}
    \caption{Depiction of the four modeling approaches used in Experiment 1. The top-down model is built on diverse  sites across the CONUS in the training dataset, the grouped models are built on subsets of those sites based on catchment attribute similarity or regional co-location, and the bottom-up model transfers models built on single sites using meta transfer learning.}
    \label{fig:exp1_diagram}
\end{figure}

Lastly, the "bottom-up" approach of transferring models built on data-rich sites for predictions in unmonitored sites uses a meta transfer learning (MTL) approach, which is analogous to calibrating process-based model parameters on a "donor" site and strategically transferring those parameters to a model for a target location \cite{willard_predicting_2021}. The MTL is instead a data-driven approach for selecting models to be transferred, where the base models can be either process-based, statistical or ML-based. The transfer is optimized using past performance metrics transferring models between sites. The MTL uses a classical ML model to predict the error of each pairwise combination of training and test sites with inputs (referred to as  meta-features) such as base model structures, catchment attributes, or input data statistics, to determine the best model for a given unmonitored site. We further describe how this model is used to select sites in Section \ref{subsubsec:architectures}. This practice of transfering parameters from a donor site to a target location has been the primary approach in PUBs modeling for decades \cite{bardossy2007calibration,patil2012controls,zelelew2014transferability,zhang2009relative}, and the well-trodden question of "How do we select the most ideal donor catchment?" for regionalization mirrors our ML question of "How do we select the ideal model for transfer learning?".

\item \textbf{Experiment 2: Examining model generalizability by reducing input data requirements} 

A key bottleneck to modeling stream temperatures in unmonitored sites is having sufficient co-located input data at the target locations of interest \cite{mcgrath2017statistical}. This experiment explores the trade-offs between model performance, data availability, and applicability when certain inputs are limited. By systematically reducing input requirements--such as stream flow and catchment attributes--it examines how this affects the number of training sites and the overall model performance, highlighting the potential of minimalist models that rely only on universally available meteorological and location data. Here, we build models based on the four different data availability groups (Figure \ref{fig:exp2_diagram}a). For each of these data availability groups, we built two versions of models to examine the effect of additional sites with stream temperature observations becoming available for training by dropping input data requirements (Figure \ref{fig:exp2_diagram}b).

\begin{figure}[h]
    \centering
    \includegraphics[width=\textwidth]{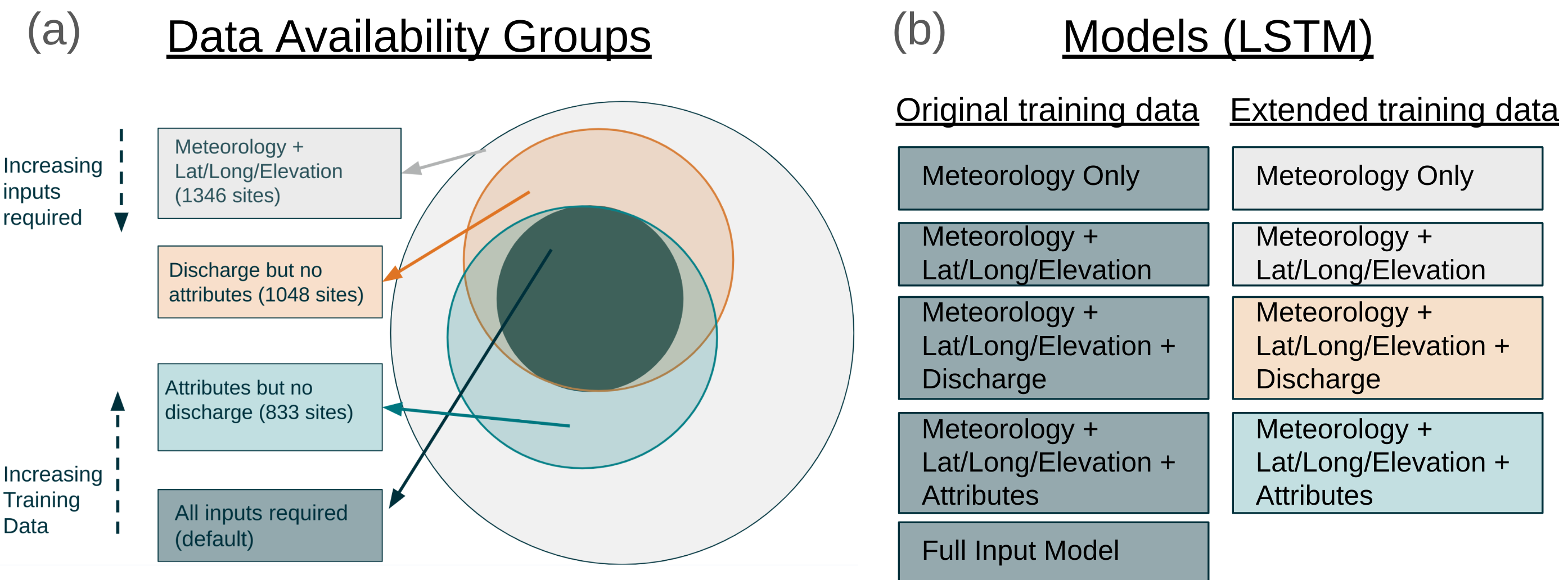}
    \caption{Depiction of the four data availability groups (a) and corresponding eight models approaches (b) in Experiment 2. The light grey data availability group represents all sites in the dataset since meteorology, latitude, longitude and elevation can be obtained at most locations with acceptable spatial resolution from gridded or digital elevation data products. The orange group  adds co-located discharge measurements as a constraint, and the light blue adds GAGES-II attributes, decreasing the amount of training data available in both cases. The dark blue group requires all three types of inputs like in Experiment 1, resulting in the least amount of training data. The models listed under "Original Training Data" are all trained on the same 782 training sites from Experiment 1, while the ones listed under "Extended Training Data" include up to 562 additional sites that become available due to having less input data requirements.}
    \label{fig:exp2_diagram}
\end{figure}

\item \textbf{Experiment 3: Assessing the appropriate set of catchment attributes to include in top-down models and associated trade-offs with model complexity} 

This experiment addresses the gap in comparisons of different sets of catchment characteristics used as inputs for modeling. With large geospatial datasets that have information on hundreds of attributes for thousands of catchments, it is currently possible to include a large number of these attributes as inputs to a deep learning model. However, doing so introduces a significant degree of complexity across the ML modeling pipeline including the neural network model and the optimization process, as well as increased computational time. Also, many catchment characteristics have overlapping information content \cite{ciulla2023network}, and redundant inputs to deep learning models can potentially reduce model performance \cite{peng2005feature,wang2021feature} and interpretability \cite{ismail2021improving,pan2023interpretability}. 

In this experiment, we test three different sets of attributes (Figure \ref{fig:exp3_diagram}):

\begin{itemize}
     \item \textbf{All available:} This is a comprehensive set  of 274 non-categorical GAGES-II attributes that describe the physical, climatic, and land-use characteristics of catchments \cite{falcone_gages-ii_2011}, and is used as the default inputs for the top-own and grouped models in Experiment 1. While this set maximizes the amount of catchment information provided to the model, it also introduces a higher degree of complexity and redundancy. This approach allows us to test whether increasing the breadth of input data leads to improvements in model performance or if the added complexity outweighs the benefits.
     
    \item \textbf{Expert-selected:} This set includes a small, targeted group of 21 attributes used in a prior work using LSTM for stream temperature prediction in unmonitored basins that was determined based on insights from hydrological experts \cite{rahmani2021deep}. This includes variables that typically influence stream temperatures such as catchment drainage area, mean annual precipitation, elevation, straightline distance to nearest dam, and percent forest cover.

    \item \textbf{Aggregated z-scores :} This is a set of catchment attribute categories derived from a network-based method to address redundancy in the attribute information and improve model efficiency. \cite{ciulla2023network}.  The inputs used are the averaged z-score values of all the attributes for each category. This approach provides an objective means to reduce the dimensionality of any dataset of catchment attributes,  balancing model interpretability and complexity, and potentially improving model performance by eliminating unnecessary or highly correlated attributes.
    
\end{itemize}

\end{itemize}

\begin{figure}[h]
    \centering
    \includegraphics[width=\textwidth]{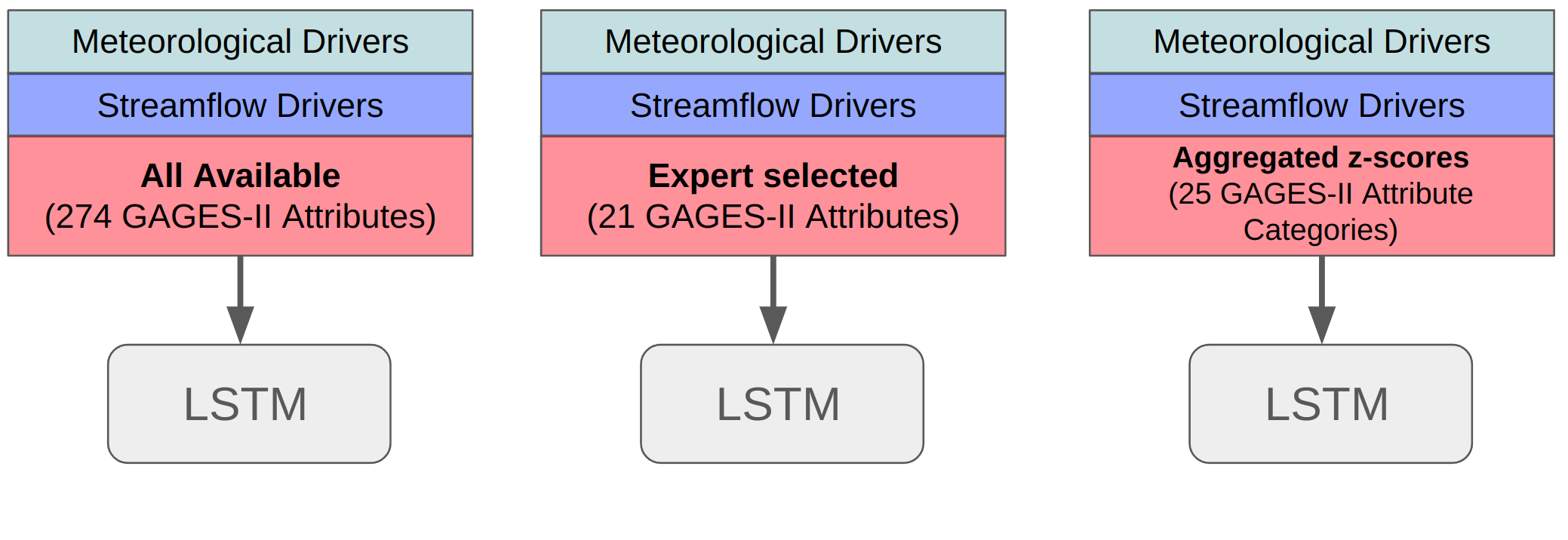}
    \caption{Depiction of the three different representations of catchment attributes used as inputs for the top-down models in Experiment 3.}
    \label{fig:exp3_diagram}
\end{figure}

\section{Materials and Methods}
\label{sec:mats_methods}

\subsection{Data}
\label{subsec:methods:data}

The following section describes the datasets used in this study and associated preprocessing steps. All the data used and generated in this study along with the code, models, model outputs, and results, are available at \citeA{willard2024dataset}.

\subsubsection{Data Sources}
\label{subsubsec:data_sources}

We used daily stream temperature data from all sites available in the United States Geological Survey's (USGS) National Water Information System (NWIS) that had at least 365 observations. All sites that had any water temperature observations below -1\textdegree C or over 40\textdegree C were removed to account for data errors and hot springs. This results in 2999 sites containing 8,347,577 observations across the conterminous United States. However, to obtain input data for the models, sites needed to have co-located continuous streamflow data as well as attributes in the USGS GAGES-II dataset (as described below). This results in 1362 sites containing 4,378,876 stream temperature observations. We use this subset of data for most experiments, with an exception being Experiment 2 which draws data from the 2999 sites, which we describe in Section \ref{subsubsec:exp_details}.

The dynamic input data to the models consists of both meteorological and streamflow data. The meteorological values were obtained from the Daymet product \cite{thornton_daymet_2016} and consist of day length, air temperature (mean, maximum, and minimum), snow water equivalent, vapor pressure, solar radiation, and precipitation . The Daymet variables were obtained at same location as the stream temperature sites, and were not basin-averaged since stream temperatures are more likely to be driven by local meteorology. The selection of these meteorological features is supported by previous research which indicates that these variables are critical drivers of stream temperature dynamics. For instance, numerous studies have examined the mechanistic effect of air temperature on stream temperature \cite{mohseni1998nonlinear,caissie2006thermal,webb2008recent}. Also, precipitation and snow water equivalent are known to influence streamflow and temperature particularly in mountainous regions \cite{isaak2012climate}. We also include streamflow since it is often considered a necessary input for stream temperature prediction \cite{piccolroaz_prediction_2016,naresh2017modeling,rahmani2021exploring,van2011global} due to  stream temperature being affected by processes such as advection from upstream reaches, snow melt and runoff, and groundwater exchange. The streamflow and precipitation inputs each consist of two components, the first being the cubic feet per second value and the second is the log-transformed version of the first value. This choice was made in order to account for the fact that these data can span several orders of magnitude and the model may need to capture both linear and non-linear relationships between flow and temperature. In general, we took an inclusive approach to input selection since deep learning models are generally robust to non-informative inputs \cite{goodfellow2016deep}. This yields 9 meteorology and 2 streamflow values in total for each sampling date. 

The complete set of site attributes used in this study consists of the 274 non-categorical site-specific catchment attributes from the USGS GAGES-II dataset \cite{falcone_gages-ii_2011} (for a detailed list see \url{https://doi.org/10.5066/F7HQ3XS4}). These attributes consist of a diverse set of physical and environmental site characteristics including their climatology, land cover, land use, reservoirs and other human activities. Though many of these attributes are actually dynamic (e.g. land use, climate variables), we treat them as static for the purposes of this study. As in \citeA{ciulla2023network}, we also transform attributes related to a site's distance to a dam (e.g. distance to nearest major dam, distance to nearest dam) that have values of $-999$ in the absence of dams. This is done by setting the $-999$ values to zero, and then taking the inverse of values greater than zero to maintain the monotonic relationship (for details see \citeA{ciulla2023network}). The site attributes chosen for use in models were 274 GAGES-II attributes. In Experiment 3, we also use  a smaller set of GAGES-II attributes based off a prior LSTM stream temperature modeling study \cite{rahmani2021deep} and  attribute categories as described in \citeA{ciulla2023network} and available at \citeA{ciulla2023data}.

\subsubsection{Data acquisition}
\label{subsubsec:data_acquisition}
Stream temperature and meteorological data for this study has been acquired using the BASIN-3D (Broker for Assimilation, Synthesis, and Integration of eNvironmental Diverse, Distributed Datasets) data integration tool \cite{varadharajan_basin-3d_2022}, and the catchment attributes (treated as static) were acquired from the Geospatial Attributes of Gages for Evaluating Streamflow (GAGES-II) dataset \cite{falcone_gages-ii_2011}.  BASIN-3D takes the queries returned from USGS and Daymet and harmonizes them into a single HDF5 file. All input and water temperature data was limited to the years from 1980 to 2021 for the purposes of this study. 
\subsubsection{Normalization}
\label{subsubsec:normalization}

Prior to modeling using deep learning, it is recommended to normalize the input and target data for improved convergence speed and both numeric and gradient stability \cite{goodfellow2016deep}. This is especially important when inputs are on different scales, which they are in this study. Therefore, we normalize all dynamic and static inputs and the stream temperature observations using z-score normalization to a mean of 0 and standard deviation of 1 based on a calculated global mean and standard deviation for each input across all stream sites in a given  training dataset. The z-score normalization is defined as $$x_{\text{norm}} = \frac{x - \mu}{\sigma}$$ where \( x \) is the original value, \( \mu \) is the mean, and \( \sigma \) is the standard deviation of the input or target variable.

\subsection{Model and Experimental Setup}
\label{subsec:model_exp_setup}
Here we describe the ML model frameworks and architectures used for stream temperature prediction, as well as any preprocessing, hyperparameter tuning, feature selection, and feature extraction that was done. Then, we describe the three experiment procedures in more detail. Lastly, we cover how the models are evaluated and how feature importance is calculated using the permutation feature importance technique. 

\subsubsection{Model descriptions}
\label{subsubsec:architectures}

\paragraph{Long short-term memory (LSTM)}
\label{para:lstm_architecture}
The LSTM is a type of recurrent neural network that includes dedicated memory that can store information over long time periods \cite{hochreiter_long_1997}. This memory function is analogous to a system state vector in dynamical systems and process-based modeling, making it a popular architecture for modeling watershed processes \cite{kratzert2019benchmarking,willard2023water}. Compared to other variants of recurrent neural networks, LSTMs are not as vulnerable to the problem of exploding and vanishing gradients during training.  For further details on the LSTM recurrent neural network architecture see \citeA{hochreiter_long_1997}, and we also include a brief description of the architecture in the supplementary material (Section S1). We used the Pytorch \cite{paszke2019pytorch} LSTM class for implementation, and all models were trained on the Perlmutter HPC system within the National Energy Research Scientific Computing Center (NERSC) using Nvidia A100 GPUs. 


\paragraph{Meta Transfer Learning}
\label{para:mtl_architecture}

The MTL framework used in the Experiment 1 for the "bottom-up" approach uses a metamodel in addition to the LSTM models previously described. While the LSTM models predict stream temperature itself, the metamodel predicts the expected performance an LSTM model built on one site and transferred to predict in another. For the metamodel we use an extreme gradient boosting (XGBoost) regression model architecture due to its prior success in an MTL model \citeA{willard_predicting_2021} and ease of implementation. A list of the 300 total candidate meta-features can be seen in supplementary Table S1, and we perform recursive feature elimination with cross validation (RFECV) to narrow it down to 234 listed in Table S2. We use the default XGBoost hyperparameters for both RFECV and also the final model. Here, we use the metamodel in the same manner as previous works on lake temperature and streamflow prediction \cite{willard_predicting_2021,ghosh2022meta} which proceeds as follows:

\begin{enumerate}
	\item Build and train five LSTM models for each of the well-monitored stream sites in the training dataset.
	\item For each site in the training dataset, use all models built on \textit{other} training sites to predict daily stream temperatures and evaluate prediction accuracy.  This yields $N*(N-1)$ prediction performance values where $N$ is the number of training sites.
	\item Train the meta-learning XGBoost regression model to predict the collected model RMSE performance values from (2) based on the meta-features found using RFECV.
	\item Given an artificially unmonitored stream site, where data is only used for final evaluation, and its meta-features, use the meta-learning model to predict model performance of each  model built on a training site. Use the models with the lowest predicted errors to model the target.
\end{enumerate}

Given an unmonitored target site, we select the 10 sites that contain the source models with the lowest predicted error by the meta-model. Selecting more than one source model and combining them in an ensemble was found in Willard et al. \cite{willard_predicting_2021} to significantly outperform the single source model transfer. These source models are combined in an ensemble and the final predictions are an average of the 10 models.



\subsubsection{Experiment Details}
\label{subsubsec:exp_details}

As described above, we conducted three experiments with different inputs or models (Table \ref{tab:model_list}) to assess their impact on the accuracy of stream temperature predictions in unmonitored locations. For this study, we split the 1362 sites as follows. The 782 sites with a total of at least five years  (1825 sampling dates) of stream temperature observations were used for training. The remaining 580 sites that had between one and five years of data were treated as representative of  "unmonitored" sites and used for testing in the final evaluation. The special case of Experiment 2 uses an extended dataset of 2999 sites, with 1346 sites having at least five years of data and used for training.  The observations are not required to be continuous in time so there are many missing sampling dates in the majority of sites, but these are simply masked out of the optimization and evaluation processes (Sections \ref{subsubsec:model_description:hpo}, \ref{subsubsec:model_eval}, S1), so it does not pose an issue with model training or evaluation. 

For all three experiments the LSTM models were constructed using the same hyperparameters and training protocols (Section \ref{subsubsec:model_description:hpo}), only differing in the choice of inputs and the breadth of training data. Note that when the training data changes, the normalization (Section \ref{subsubsec:normalization}) of the training and testing data in each case will also change.

\textbf{\textit{Experiment 1:}} In this experiment comparing top-down, bottom-up and grouped modeling approaches, we use the same meteorological and stream flow inputs for all models (Table \ref{tab:model_list}). However, only \textit{LSTM\_conus}, \textit{LSTM\_regional}, and \textit{LSTM\_cluster} use the 274 GAGES-II attributes as inputs. This is because the single site LSTM models used in the MTL have no use for inputs that are specific for each sites, since those values will be constant and non-informative. Instead, in the MTL approach, the attributes are used as meta-features eventually (Table S1) in the XGBoost regression model (Section \ref{para:mtl_architecture}). Also, for each of the grouped approaches, test sites are predicted solely by the model built on the region or cluster they belong to. 


\begin{table}[]
    \centering
    \begin{tabular}{|p{.5\linewidth} p{.5\linewidth}| }
        \hline
         Name & Inputs \\
         \hline
         \textbf{Experiment 1} & \\
         \textit{LSTM\_conus} & Meteorology, Location, Streamflow, and all 274 GAGES-II attributes  \\
         \textit{LSTM\_regional} & Meteorology, Location, Streamflow, and all 274 GAGES-II attributes  \\
         \textit{LSTM\_cluster} & Meteorology, Location, Streamflow, and all 274 GAGES-II attributes  \\
         \textit{MTL Source LSTM models (n=782)} & Meteorology and Streamflow \\
         \textit{MTL XGBoost Metamodel} & 234 Inputs shown in Table S2 \\

         \hline
         \textbf{Experiment 2 Default Training Data} & \\
         \textit{Meteo only} & Meteorology \\
         \textit{Meteo + Lat/Long/Alt} & Meteorology, Location \\
         \textit{Meteo + Lat/Long/Alt} + Flow & Meteorology, Location, Streamflow \\
         \textit{Meteo + Lat/Long/Alt + GAGES-II} & Meteorology, Location, all 274 GAGES-II attributes  \\
        \hline
        \textbf{Experiment 2 Extended Training Data} & \\
        \textit{Meteo only (extended data)}& Meteorology \\
        \textit{Meteo + Lat/Long/Alt (extended data)} & Meteorology, Location \\
        \textit{Meteo + Lat/Long/Alt + Flow (extended data)} & Meteorology, Location, Streamflow \\
        \textit{Meteo + Lat/Long/Alt + GAGES-II attr (extended data)} & Meteorology, Location, all 274 GAGES-II attributes  \\
         \hline
        \textbf{Experiment 3} & \\
        \textit{LSTM\_conus} (same as Experiment 1) & Meteorology, Location, Streamflow, and all 274 GAGES-II attributes  \\
        \textit{LSTM\_conus (expert-selected attributes)} & Meteorology, Location, Streamflow, and 21 expert-selected GAGES-II attributes  \\
        \textit{LSTM\_conus (z-score-based attributes)} & Meteorology, Location, Streamflow, and 25 attributes represented a condensed version of the total 274 GAGES-II attributes  \\
        \hline

    \end{tabular}
    \caption{List of LSTM models trained for all three experiments and the corresponding inputs used for each model. Here, location data consists of latitude, longitude, and elevation of each site. The list of all 274 non-categorical GAGES-II site attributes can be found in \citeA{falcone_gages-ii_2011}. The list of the 21 expert-selected attributes can be found in Table S3 and the list of 25 z-score-attributes can be found in Table S4}
    \label{tab:model_list}
\end{table}





\textbf{\textit{Experiment 2:}} We built eight models for Experiment 2 that are compared with the top-down model from Experiment 1 (\textit{LSTM\_conus}). The first set of models (left column of Figure \ref{fig:exp2_diagram}b)  uses the same amount of data originally used in Experiment 1, and only changes the inputs provided (see "Experiment 2 Default Training Data" in Table \ref{tab:model_list}). The second set of models (right column of Figure \ref{fig:exp2_diagram}b)  uses additional stream temperature observations that become available by reducing the set of inputs (see "Experiment 2 Extended Training Data" in Table \ref{tab:model_list}), effectively increasing the size of the training data. The models built using only meteorology and location data (latitude, longitude, and elevation) as inputs are able to use stream temperature data from 2999 sites since the inputs are universally available for any location using gridded meteorology and topography datasets. This amounts to \char`~ 6.13 million training observations spanning 1346 sites (light grey circle in Figure \ref{fig:exp2_diagram}) used as inputs for the extended set of models, in contrast to \char`~ 3.82 million available across 782 sites (dark blue filled circle in Figure \ref{fig:exp2_diagram}) used in Experiment 1. The second data availability group (shown in orange in Figure \ref{fig:exp2_diagram}) includes river discharge (streamflow) as an additional input. With this additional requirement the total available training data is reduced to \char`~ 4.65 million stream temperature observations spanning 1048 sites. The third data group includes GAGES-II attributes as a requirement but does not use river discharge, resulting in  \char`~ 4.13 million observations over 833 sites. The final data availability group is the same as Experiment 1 (Section \ref{subsec:methods:data}), where all default inputs are required including meteorology, river discharge, and GAGES-II attributes.

\textbf{\textit{Experiment 3:}} Experiment 3 compares the performance of the \textit{LSTM\_conus} model using two alternate sets of catchment attributes  with the default set used in Experiment 1 (Table \ref{tab:model_list}).  The first alternative consists of 21 expert-chosen values including properties like dam storage density and number of dams in the watershed, drainage area, stream density, straightline distance to nearest dam, land use, topographic features, location, and meteorological statistics (for further details see Table S3). The second alternative is a reduced set of attribute categories obtained using a network approach to minimize redundancies listed in Table S4 and can be downloaded from the dataset \citeA{ciulla2023data}. In the network, the nodes are attributes, and the edge weights represent their similarity. The attributes are then clustered  into different clusters using a community detection algorithm \cite{ciulla2023network}. The network-based attribute aggregation produces highly interpretable categories. For example, attributes associated with urban development are classified into one category that we refer to as "Developed Areas", while attributes related to temperature are classified under a different "Temperature"  category (for a detailed breakdown of all the attribute categories see \citeA{ciulla2023network}). By categorizing attributes, we average out their slight differences and redundancies, and can analyze their collective contribution. The values used as model inputs represent the averaged z-scores of all the attributes within a particular category. The models in Experiment 3 were constructed using the same hyperparameters and training protocols as in Experiment 1, and only differed in the choice of inputs.


\subsubsection{Hyperparameter optimization}
\label{subsubsec:model_description:hpo}


Hyperparameters for the LSTM were tuned using Bayesian  optimization available through the hyperparameter sweep framework from the Weights and Biases  platform (\url{http://wandb.ai}, \citeA{biewald2020experiment}). As opposed to the standard grid search, which exhaustively searches all possible combinations, Bayesian optimization guides the search by learning from previous evaluations and is  a more efficient way to find optimal parameters \cite{wandbWhatBayesian}. The ranges for the hyperparameter sweeps as well as hyperparameters set to a single value (Table S5 in Supplementary Information) were based on experiments from previous water temperature modeling studies \cite{willard2023machine,read_process-guided_2019,jia_physics-guided_2021-1}). The learning rate was set to 0.001, initial weights were set using the Xavier normal distribution \cite{glorot2010understanding}, and the AdamW optimizer \cite{loshchilov2017decoupled} was used with a mean squared error loss function. We chose a sequence length of 200 days, with a sliding window shift of 100 days at a time, which was also based on results from aforementioned previous studies. Predictions are only gathered from the second 100 days of the 200 day sequence, so the model is able to build up memory from previous timesteps. We tuned the batch size, number of hidden units, the number of hidden layers, the weight decay value, and the dropout rate. The runs in the sweep are scored by mean squared error (MSE) on the validation set, which consisted of 20\% of the entire training dataset. For each site, the least recent 20\% of water temperature observation were set aside and aggregated for validation. In total, the hyperparameter sweep completed 686 runs and the top five sets of hyperparameters were selected for use (Table S6). 

Models were trained with these hyperparameters until the error on the validation dataset was minimized using the early stopping technique with a patience value of 300 epochs \cite{prechelt1998early}. We undertook the same training procedure for all LSTM models, though hyperparameters were only tuned for the top-down model.   This choice was made due to the high computational expense of hyperparameter tuning, and when an additional hyperparameter optimization scheme was applied to one of the models in the attribute-based grouped modeling scenario we found the validation error was unchanged. 

We also conducted hyperparameter tuning for the XGBoost meta-model using random search and feature selection using the RFECV method in the \textit{scikit-learn} Python library \cite{pedregosa_scikit-learn_2011}. The meta-features consisted of the GAGES-II attribute differences, in addition to the mean and standard deviation of each dynamic meteorological and streamflow-based input feature, and observation statistics based off the previous MTL study \cite{willard_predicting_2021} (e.g. number of water temperature observations available for training data). These are listed in Table S1. The values found during hyperparameter tuning of the XGBoost metamodel were 152 estimators with a learning rate of 0.183. 

\subsubsection{Model realization ensembles}
ML models have uncertainty in model parameters after training due to a variety of factors including stochasticity in initialization of neural network weights and the shuffling of data between training epochs. It has been shown that ensemble results from multiple diverse model runs facilitates better overall ML model performance and robustness and also allow for the quantification of uncertainty \cite{ganaie2022ensemble}. Hence, we used an ensemble average of five realizations for all model types, except in the case of results used for the error analysis described in Section \ref{subsec:exp_err_analysis}, where we used an ensemble average of 20 model realizations to further reduce variance of model performance. We also spread out the top five hyperparameter sets found in the  sweep across this model realization ensemble to facilitate more diversity between ensemble members, which tends to improve performance over creating ensembles with more similar models \cite{krogh1995neural}. Furthermore, the validation MSE between the top five hyperparameters was negligible so an overall performance decrease was highly improbable.


\subsubsection{Model Evaluation}
\label{subsubsec:model_eval}
\paragraph{Performance Metrics}
\label{para:eval_metrics}
We assessed model performance using three primary metrics: (a) the median per-site RMSE calculated across the 580 test sites; (b) the median value of the average bias calculated for each of the test sites and (c)  the median per-site RMSE calculated on only the warmest 10\% of water temperature observations per site. All metrics were calculated on predictions averaged across the five model realizations in each case, with the exception being \textit{MTL} which has 10 predicting models. We preferred median values due to large outliers in RMSEs for some sites. However,  for comparison we have also included results for mean RMSE values instead of per-site median in supplementary material Tables S7 through S11.




\paragraph{Feature Importances}
\label{para:feat_imp}
Feature importance calculations help increase the transparency and interpretability of otherwise “black-box” modeling approaches like the LSTM and other deep learning models \cite{molnar2020interpretable}. We chose to use the permutation feature importance \cite{molnar2022interpretable} method since it is model-agnostic and straightforward in implementation and understanding. The main idea is to observe the change in model error after permuting the feature's values. This process effectively disrupts the relationship between the feature and the outcome, allowing us to determine how reliant the model's predictions are on each feature. A substantial increase in error suggests  the model heavily relies on that feature for making accurate predictions, and a negligible change in error implies the feature's relative unimportance. Though permutation feature importance can be calculated on either the training or testing data (or both), we chose to calculate it on the RMSE for all test observations in order to gauge the importance for model generalizability to unseen scenarios. 

Furthermore, we include three "combined" permutations of features that are strongly related or transformations of one another. So, "combined air temp" would be the combination of maximum, mean, and minimum daily temperature (\textit{tmax, tmean, tmin}), "combined discharge" would be the combination of river discharge (\textit{rdc}) and log-transformed river discharge (\textit{logrdc}), and "combined precipitation" would be the combination of precipitation \textit{(prcp)} and log-transformed precipitation (\textit{logprcp}), and "combined GAGES-II" is the combination of all GAGES-II attributes used as inputs. We include these combined features so that the correlation of features doesn't strongly affect the importance of a single permuted feature, when the other features are still present and informative.

\subsection{Model Comparison}
\label{subsec:model_comparison}
To compare results between model runs, we use the non-parametric two-sided Wilcoxon signed-rank statistical test \cite{rosner2006wilcoxon} with a significance level of $\alpha$ = 0.05 based on its implementation in the SciPy library \cite{2020SciPy-NMeth}. For this, we designate the top-down \textit{LSTM\_conus} model as the baseline against which all other models are compared across experiments. The significance test is performed for each test site, where we define the two populations as the absolute errors for each model calculated across every water temperature observation at that site. As a result, each test site will be categorized as "Significant (Better)", "No Significance", or "Significant (Worse)", and we report the counts for each category. We also calculate the difference in RMSE ($\Delta$ RMSE) by subtracting the baseline model RMSE from a given model RMSE, averaged over all sites in the respective category. This analysis not only allows us to compare results across experiments, but also provides an objective measure to determine the best models that would apply for an individual site.

\subsection{Error Analysis: Assessing predictability of sites across top performing models}
\label{subsec:exp_err_analysis}
We also perform additional analysis of per-site RMSEs to determine the attributes or processes that make a site more or less predictable (using the RMSE metric) as described below. For this analysis, we focused on the model errors from Experiment 3 alone. This was because our results (Section \ref{subsec:results:exp3}) showed equivalent performance across all the different sets of catchment attributes for the model that performed the best in Experiments 1 and 2, namely the \textit{LSTM\_conus} with the default inputs.

\subsubsection{Attribute-based Error Analysis}
\label{subsubsec:network_analysis}
One of the critical aspects of our error analysis is to investigate whether specific catchment attributes are associated with sites that have higher prediction errors. 
For this, we set up an XGBoost model \cite{chen_xgboost_2016} where the inputs were all the 274 GAGES-II attributes and the target variables were the per-site RMSE values resulting from each of the three methods. We conducted hyperparameter tuning on XGBoost for the following parameters: learning rate ($[0.0001,1]$ range), number of estimators ($[10,10,000]$ range) and max depth ($[1,10]$ range), and found that the best values minimizing the model RMSE predictions are 0.015, 200 and 3 respectively. We performed a Recursive Feature Elimination (RFE) of the attributes to reduce the redundancy present in the attribute information  \cite{ciulla2023network}, and used the set of features that resulted in lowest RMSE. The importance of the attributes for the predictions are computed as the average Shapley values \cite{molnar2022interpretable} of 100 independent realizations of the best performing XGBoost model. We then calculated the Spearman correlation coefficients of the Shapley values with the LSTM model errors to determine whether the corresponding attributes improve or detract from the performance of the models in Experiment 3.


\subsubsection{Paired Air-Stream Temperature Analysis}
\label{subsubsec:groundwater_analysis}

Stream temperatures in many locations are driven by variation in local air temperatures. However, groundwater flows disrupt this coupling between air and stream temperature regimes via mixing of older, cooler water from subsurface aquifers \cite{briggs2018inferring}. Recent studies have shown that the amplitude ratio and phase lag of air and stream temperature signals can be used identify the presence of groundwater, and differentiate between shallow and deep groundwater sources to streams \cite{johnson2020paired, briggs2018inferring, hare2021continental}. Upstream dams also disrupt the coupling of air and stream temperatures through mixing of reservoir releases \cite{kedra2018climatic}.

To understand how models are able to represent thermal dynamics influenced by different processes (equilibration with air temperatures or groundwater mixing), we applied these existing paired air-stream temperature methods to distinguish between sites that were primarily influenced by air temperatures, versus those that were influenced by shallow and deep groundwater. Specifically, we fit sine-wave linear regressions for both the Daymet mean daily air temperature and daily stream water temperature records at each site. We then use the sine wave regressions to calculate two comparative metrics as described in \citeA{johnson2020paired}-- amplitude ratio and phase lag. The amplitude ratio represents the variation between maxima in air and stream temperatures.  The phase lag represents the lag between the air and stream temperature signals. We used these two values to classify each station as influenced by atmospheric, shallow groundwater, or deep groundwater factors using thresholds defined by \citeA{hare2021continental}. Lastly, we also separate out stations within 25 km of a major dam using GAGES-II metadata as done by \citeA{hare2021continental}, to create a fourth class representing the influence of dams. We analyzed model errors across these four different groups to determine if there was a pattern in predictability of individual sites based on the primary factors influencing their stream temperatures.

\section{Results}
\label{sec:results}
\subsection{Experiment 1: Top-down, bottom-up, and grouped models}
\label{subsec:results:exp1}
\begin{figure}[h]
\centering
\includegraphics[width=0.85\textwidth]{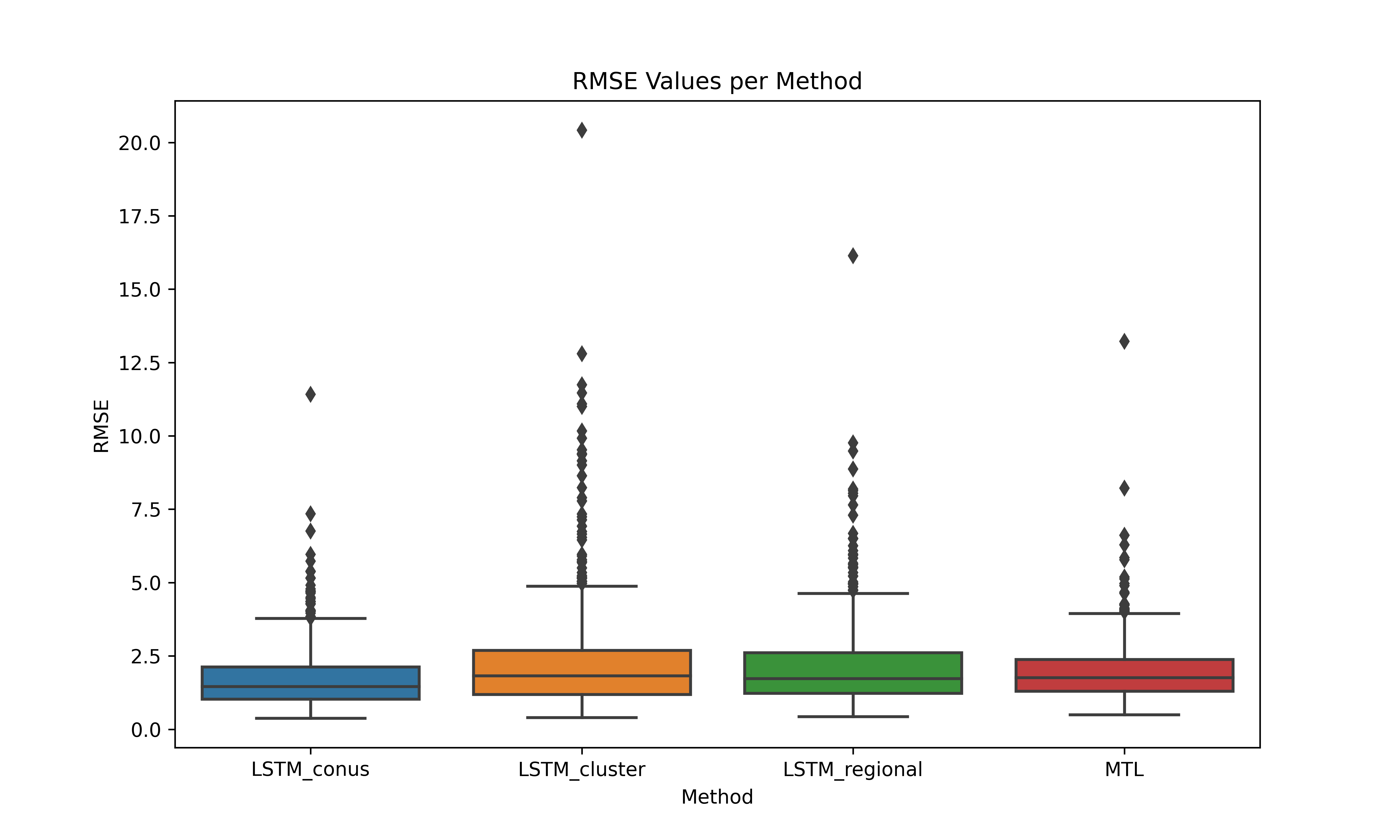}
\caption{Plot showing the distribution of per-site RMSE values for \textit{LSTM\_conus} (top-down method), \textit{LSTM\_regional} and \textit{LSTM\_cluster} (grouped methods), and \textit{MTL} (bottom-up method)}
\label{fig:boxplot}
\end{figure}

For Experiment 1, we compared model performance across per-site median values of overall RMSE, average bias, and RMSE on the warmest 10\% of observations. The \textit{LSTM\_conus} has the best overall performance (Figure \ref{fig:boxplot}, Tables \ref{tab:exp1_overall} and S7) based on the median per-site RMSE (1.43$^{\circ}C$) and the median RMSE for the warmest 10\% (1.7$^{\circ}C$). While the \textit{LSTM\_conus} is the best  model for the majority of sites, there are about some sites where the other models perform better (Table \ref{tab:wilcoxon_results}, Figure S16). However the sites that have significantly worse errors using the \textit{LSTM\_regional} and \textit{MTL} relative to the \textit{LSTM\_conus} experience a larger performance decline than the sites where those models perform significantly better, as indicated by the $\Delta$ RMSE values in Table \ref{tab:wilcoxon_results} and the outlier points in Figure \ref{fig:boxplot}. We also include example time series plots of the models predictions at four different sites in Figure S20. This result shows that overall the \textit{LSTM\_conus} performs better or just slightly worse than the other models for the vast majority of sites.


Performance metrics per USGS-defined hydrological region are shown in Tables \ref{tab:regional_results} and S8. For the 17 regions, the \textit{LSTM\_conus} consistently outperformed the grouped and bottom-up models, and had the lowest per-site median RMSE for 14 of the regions. The three exceptions where the \textit{MTL} had better performance were in the Lower Mississippi (HUC 08) and Souris-Red-Rainy (HUC 09), and California (HUC 18) regions, and in these cases the \textit{LSTM\_CONUS} was only marginally worse. The top-down \textit{LSTM\_conus} and bottom-up \textit{MTL} are much more consistent in performance, with their individual highest RMSE regions being California (HUC 18) at 2.34$^{\circ}$C for \textit{LSTM\_conus} and 3.58$^{\circ}$C for \textit{MTL} predicting in Lower Colorado (HUC 15). The grouped models have much higher errors, with the worst case scenarios for the \textit{LSTM\_regional} and  \textit{LSTM\_cluster} models in the Souris-Red-Rainy region (9.09$^{\circ}$C and 8.70$^{\circ}$C RMSE  respectively). In cases where the regional models performed meaningfully worse than the others (HUC 08, HUC09, HUC15), there were very few sites with available data. This indicates that for these cases, the regional models could have suffered from having very small groups. Regions that had overall higher errors ($>$2 $^{\circ}$C) are in the mountainous West, namely the Upper and Lower Colorado basins and California.

When comparing performance across the different clusters determined by attribute similarity, we find that the \textit{LSTM\_conus} had the lowest per-site median RMSE for 14 out of the 15 clusters (Table \ref{tab:cluster_results} and S9). The sole exception is cluster 12 where the \textit{MTL} had slightly better accuracy (RMSE =1.82$^{\circ}$C) compared to the \textit{LSTM\_conus} model (RMSE=2.00$^{\circ}$C). 



\begin{table}[h]
\centering
\begin{tabular}{|p{.18\linewidth} p{.22\linewidth} p{.16\linewidth} p{.16\linewidth} p{.16\linewidth}|} 
 \hline
 Method & Median per-site RMSE ($^{\circ}$C) (std\_dev) & Median Bias ($^{\circ}$C) & Median RMSE (warmest 10\%) ($^{\circ}$C) & n\_sites $<$ 2$^{\circ}$C RMSE\\ [0.5ex] 
 \hline
 \textit{\textbf{LSTM\_conus}} & \textbf{1.43($\pm$0.98)}  & -0.09($\pm$1.22) & \textbf{1.70($\pm$1.58)} & 406\\ \hline
 \textit{LSTM\_regional} & 1.98($\pm$1.78) & 0.04($\pm$1.76) & 2.22($\pm$2.89) & 293\\ \hline
 \textit{LSTM\_cluster} & 2.02($\pm$1.66) & 0.14($\pm$1.80) & 2.10($\pm$2.36) & 287 \\ \hline
 \textit{MTL} & 2.00($\pm$1.07) & -0.55($\pm$1.57) & 2.29($\pm$1.75) & 290 \\ 
 \hline
\end{tabular}
\caption{Experiment 1 Overall RMSE Results for the top-down \textit{LSTM\_conus} model, the grouped \textit{LSTM\_regional} and \textit{LSTM\_cluster} models, and the bottom-up \textit{MTL} model. All values are median per-site values across the 580 testing sites. Lowest RMSE values are shown in bold.}
\label{tab:exp1_overall}
\end{table}

\begin{table}[!ht]
\centering
\scalebox{0.73}{
\begin{tabular}{|l l p{1.5cm} l l l l|}
\hline
Region & n\_obs (train/test) & n\_sites (train/test) & \textit{LSTM\_conus} & \textit{LSTM\_regional} & \textit{LSTM\_cluster} & \textit{MTL} \\ \hline
01 (New England) & 101,768/9,784 & 10/23 & \textbf{1.08($\pm$0.74)} & 1.97($\pm$2.37) & 1.71($\pm$0.96) & 1.92($\pm$0.77) \\ \hline
02 (Mid-Atlantic) & 477,381/88,224 & 81/121 & \textbf{1.27($\pm$0.88)} & 1.61($\pm$1.17) & 1.72($\pm$1.73) & 1.80($\pm$1.02) \\ \hline
03 (South Atlantic-Gulf) & 537,631/64,533 & 97/77 & \textbf{1.41($\pm$0.81)} & 1.63($\pm$1.27) & 2.10($\pm$1.44) & 1.71($\pm$0.79) \\ \hline
04 (Great Lakes) & 282,868/18,522 & 58/22 & \textbf{1.04($\pm$0.29)} & 1.25($\pm$0.78) & 1.39($\pm$2.04) & 1.48($\pm$0.41) \\ \hline
05 (Ohio) & 273,588/28,431 & 66/41 & \textbf{1.39($\pm$0.83)} & 2.12($\pm$1.75) & 1.85($\pm$1.13) & 2.24($\pm$0.91) \\ \hline
06 (Tennessee) & 48,359/3,174 & 15/5 & \textbf{0.86($\pm$0.35)} & 1.72($\pm$1.64) & 0.94($\pm$0.70) & 1.69($\pm$0.50) \\ \hline
07 (Upper Mississippi) & 122,889/28,053 & 32/29 & \textbf{1.74($\pm$0.83)} & 2.75($\pm$1.68) & 2.35($\pm$1.09) & 1.89($\pm$0.87) \\ \hline
08 (Lower Mississippi) & 18,467/3,813 & 5/5 & 1.86($\pm$1.40) & 7.22($\pm$1.75) & 2.55($\pm$1.07) & \textbf{1.83($\pm$1.57)} \\ \hline
09 (Souris-Red-Rainy) & 19,440/4,188 & 4/5 & 1.61($\pm$0.98) & 9.07($\pm$1.30) & 8.70($\pm$3.65) & \textbf{1.43($\pm$0.64)} \\ \hline
10 (Missouri) & 220,070/46,083 & 52/54 & \textbf{1.46($\pm$0.91)} & 1.85($\pm$2.17) & 2.05($\pm$1.29) & 2.04($\pm$0.97) \\ \hline
11 (Arkansas-White-Red) & 322,319/21,989 & 56/23 & \textbf{1.44($\pm$0.51)} & 2.57($\pm$1.36) & 1.97($\pm$2.49) & 2.17($\pm$0.50) \\ \hline
12 (Texas-Gulf) & 172,894/15,377 & 26/16 & \textbf{1.57($\pm$1.28)} & 1.99($\pm$1.21) & 2.37($\pm$1.72) & 2.92($\pm$1.14) \\ \hline
13 (Rio Grande) & 14,533/0 & 3/0 & N/A & N/A & N/A & N/A \\ \hline
14 (Upper Colorado) & 234,865/27,723 & 47/31 & \textbf{2.03($\pm$0.89)} & 2.56($\pm$1.36) & 2.27($\pm$1.14) & 2.24($\pm$1.10) \\ \hline
15 (Lower Colorado) & 5,609/5,742 & 2/7 & \textbf{2.10($\pm$3.07)} & 6.38($\pm$1.68) & 4.24($\pm$2.45) & 3.58($\pm$3.87) \\ \hline
16 (Great Basin) & 115,347/17,914 & 25/20 & \textbf{1.64($\pm$1.03)} & 2.91($\pm$1.92) & 1.79($\pm$1.27) & 2.52($\pm$0.73) \\ \hline
17 (Pacific Northwest) & 901,098/67,182 & 142/70 & \textbf{1.22($\pm$0.89)} & 1.27($\pm$1.17) & 1.54($\pm$1.30) & 2.17($\pm$0.96) \\ \hline
18 (California) & 138,220/36,043 & 54/44 & 2.34($\pm$1.06) & 2.85($\pm$1.41) & 3.65($\pm$1.57) & \textbf{2.20($\pm$0.98)} \\ \hline
\end{tabular}}
\caption{RMSE statistics per USGS-defined hydrological region (\url{https://water.usgs.gov/GIS/regions.html}). Values are the median RMSE($^{\circ}$C) $\pm$standard deviation across sites in a region. Lowest median RMSE values per region are shown in bold.}
\label{tab:regional_results}
\end{table}

\begin{table}[!ht]
    \centering
    \scalebox{0.75}{
    \begin{tabular}{|l l p{1.5cm} l l l l|}
        \hline
        Cluster & n\_obs (train/test) & n\_sites (train/test) & \textit{LSTM\_conus} & \textit{LSTM\_regional} & \textit{LSTM\_cluster} & \textit{MTL} \\ \hline
        0 & 882,382/105,762 & 183/109 & \textbf{1.78($\pm$0.99)} & 2.10($\pm$1.66) & 1.92($\pm$1.09) & 2.05($\pm$0.92) \\ \hline
        1 & 182,149/46,399 & 46/45 & \textbf{1.44($\pm$1.29)} & 1.80($\pm$1.26) & 1.72($\pm$1.08) & 1.73($\pm$0.74) \\ \hline
        2 & 152,682/54,531 & 38/59 & \textbf{1.39($\pm$0.72)} & 1.64($\pm$2.14) & 1.63($\pm$0.79) & 1.39($\pm$0.77) \\ \hline
        3 & 476,469/41,125 & 89/47 & \textbf{1.39($\pm$0.89)} & 1.56($\pm$1.36) & 1.59($\pm$1.16) & 1.42($\pm$1.07) \\ \hline
        4 & 193,885/59,760 & 40/73 & \textbf{1.15($\pm$1.16)} & 1.69($\pm$1.07) & 1.69($\pm$0.97) & 1.38($\pm$0.51) \\ \hline
        5 & 159,958/30,618 & 38/38 & \textbf{1.85($\pm$1.81)} & 2.12($\pm$1.59) & 6.64($\pm$4.03) & 2.36($\pm$2.00) \\ \hline
        6 & 662,196/47,206 & 99/44 & \textbf{1.16($\pm$0.99)} & 1.31($\pm$1.06) & 1.46($\pm$1.37) & 1.86($\pm$0.90) \\ \hline
        7 & 315,696/27,276 & 56/33 & \textbf{1.11($\pm$0.69)} & 1.39($\pm$0.76) & 1.36($\pm$0.83) & 1.41($\pm$1.55) \\ \hline
        8 & 246,997/42,897 & 57/51 & \textbf{1.68($\pm$1.02)} & 1.93($\pm$1.28) & 1.78($\pm$1.18) & 1.76($\pm$0.93) \\ \hline
        9 & 98,175/21,157 & 22/23 & \textbf{1.71($\pm$0.75)} & 1.94($\pm$0.86) & 2.18($\pm$2.07) & 2.09($\pm$0.55) \\ \hline
        10 & 187,683/10,136 & 34/12 & \textbf{1.47($\pm$0.59)} & 2.32($\pm$2.37) & 2.35($\pm$0.85) & 1.95($\pm$0.45) \\ \hline
        11 & 206,921/23,653 & 34/22 & \textbf{1.11($\pm$0.72)} & 1.60($\pm$1.97) & 1.63($\pm$2.12) & 2.04($\pm$0.64) \\ \hline
        12 & 49,148/9,582 & 9/9 & 2.00($\pm$1.30) & 2.46($\pm$1.33) & 3.50($\pm$1.34) & \textbf{1.82($\pm$0.73)} \\ \hline
        13 & 92,009/9,525 & 21/8 & \textbf{1.78($\pm$1.15)} & 2.90($\pm$1.36) & 2.06($\pm$1.56) & 2.11($\pm$0.96) \\ \hline
        14 & 22,750/5,491 & 6/5 & \textbf{1.60($\pm$0.72)} & 2.19($\pm$2.42) & 3.72($\pm$1.52) & 2.06($\pm$0.90) \\ \hline
    \end{tabular}}
    \caption{RMSE statistics per cluster. Values are the median RMSE($^\circ$C) $\pm$ standard deviation across sites in a cluster, with per-site RMSE standard deviation in parentheses. Lowest median values per cluster are shown in bold.}
\label{tab:cluster_results}
\end{table}

\begin{figure}[h]
\centering
\includegraphics[width=\textwidth]{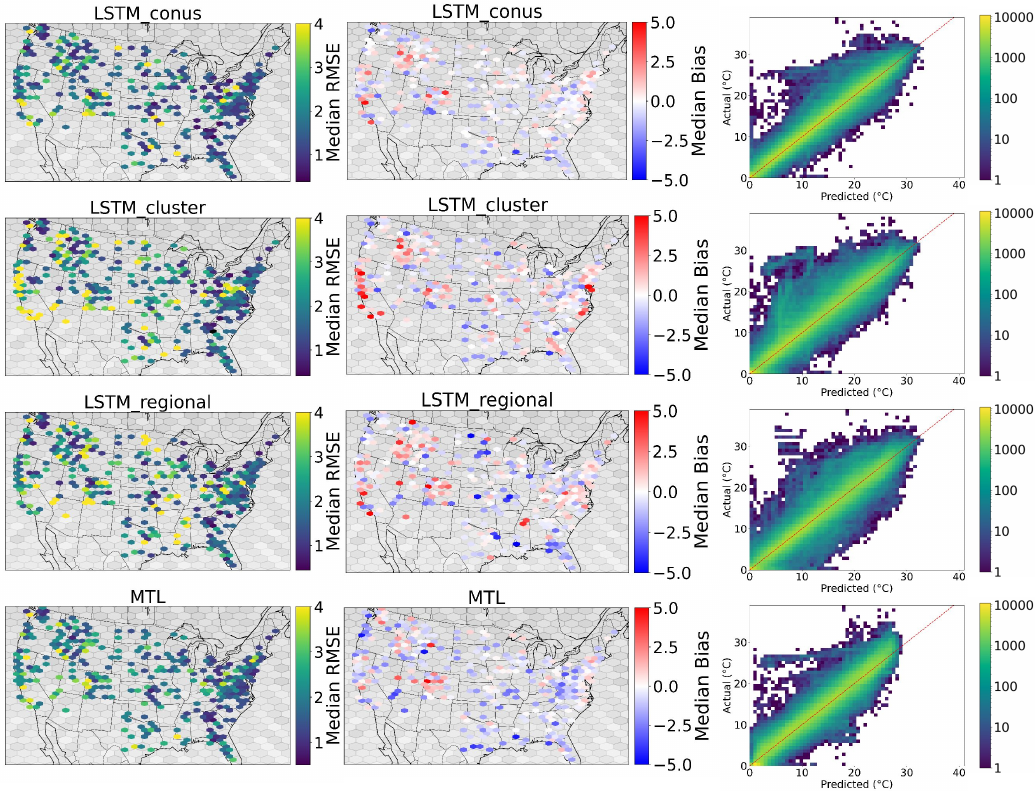}
\caption{Twelve panel plot showing the performance of the methods used in Experiment 1: \textit{LSTM\_conus} (top-down method), \textit{LSTM\_regional} and \textit{LSTM\_cluster} (grouped methods), and \textit{MTL} (bottom-up method). The left and middle columns show the hexbin spatial distribution of per-site RMSE and mean bias values in $^{\circ}$C, where the color represents the median within each hexbin. The right column is a two dimensional histogram showing the distribution of individual stream temperature predictions across all sites. The color represents the count within each bin.}
\label{fig:exp1_8panel}
\end{figure}

\subsubsection{Experiment 1 feature importances}
\label{subsubsec:results:exp1:featimp}

\begin{figure}
    \centering
    \includegraphics[width=0.95\textwidth]{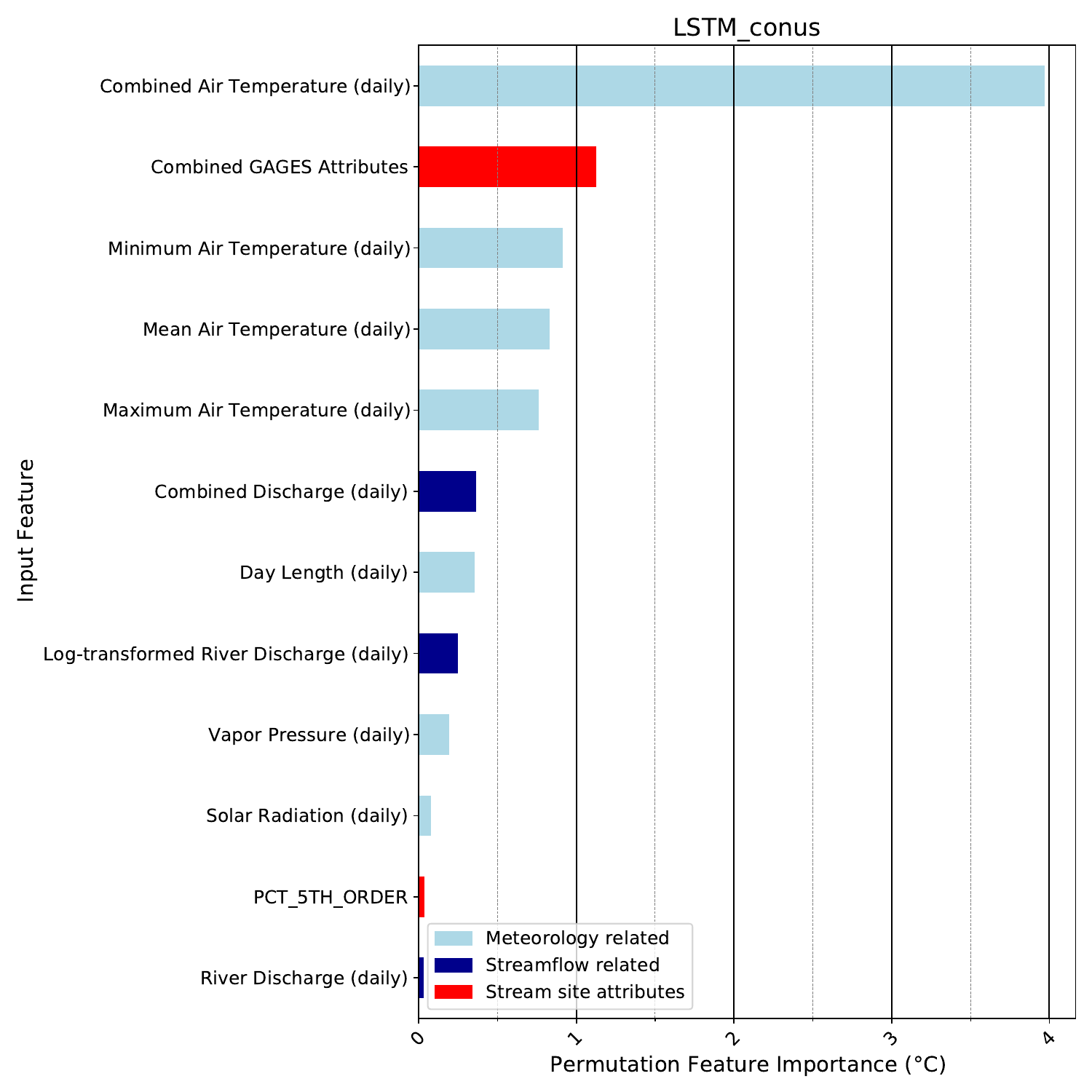}
    \caption{Bar chart displaying the permutation feature importance for various input features in the \textit{LSTM\_Conus} model from Experiment 1 measured by how much the RMSE increases relative to the RMSE calculated using all test observations (1.97$^{\circ}$C). Features are categorized into three groups: meteorology-related features (light blue), streamflow-related features (dark blue), and catchment attributes (red). Only importance values greater than 0.03$^{\circ}$C are shown to focus on features with a significant impact on the model's predictive accuracy. This threshold is set based on the standard deviation of the RMSE per individual member of the model ensemble, highlighting features whose impact is beyond the ensemble's inherent variability.}
    \label{fig:pfi_conus}
\end{figure}
The feature importance results for \textit{LSTM\_conus} show that the largest factor influencing prediction errors is air temperature with permutation importance values of 3.97$^{\circ}$C for all three air temperature inputs combined (Figure \ref{fig:pfi_conus}). The combined impact of the GAGES-II features is 1.12$^{\circ}$C, showing that catchment attributes are considered important by the model. Other significant importance values from daily meteorology include day length (0.36$^{\circ}$C) and to a lesser extent vapor pressure (0.19$^{\circ}$) and solar radiation (0.08$^{\circ}$). Features related to daily streamflows account for 0.36$^{\circ}$C combined. Only one individual GAGES-II attribute was deemed significant, which was the percentage of a stream segment that is of the 5th order (PCT\_5TH\_ORDER) at 0.03$^{\circ}$. All other features were less than the 0.03$^{\circ}$C threshold, as defined by the standard deviation of the RMSE per individual member of the model ensemble. Amongst the other models (Figures S2-S4) air temperatures was also the primary feature of importance. Only the \textit{LSTM\_cluster} had high importance values for the combined GAGES-II attributes while the \textit{LSTM\_regional} had negligible importance for the same attributes. As described above, the \textit{MTL} single-site models can't make use of attributes since they are trained on a single site.

\subsection{Experiment 2}
\label{subsec:results:exp2}

Experiment 2 explored the impact of omitting certain input variables, such as location data, streamflow, and catchment attributes, on model performance while also considering the influence of additional stream temperature observations  that become available for training by excluding requirements for their co-location with these variables. The following results are based on dividing the model performance into four groups based on availability (Figure \ref{fig:exp2_diagram} and \ref{subsubsec:exp_details}). We evaluated two model versions for each group: one using a fixed data quantity as in Experiment 1 and another utilizing an extended stream temperature dataset for the given inputs. 

\subsubsection{Experiment 2 Performance}
\label{subsubsec:results:exp2:perf}

The results of Experiment 2 indicate that the \textit{LSTM\_conus} with default inputs as in Experiment 1 has the lowest median errors, although performance for some of the other models with reduced inputs are close (Table \ref{tab:exp2_overall}). Surprisingly, median per-site RMSEs for the worst performing models (using meteorology only as minimalist inputs) were only about 25\% higher (median RMSE=1.77°C) than the best model \textit{LSTM\_conus} (median RMSE=1.43°C). Adding location information to the meteorology inputs did not meaningfully improve performance. In general, most of the sites,  approximately 75\% for the \textit{Meteo\_only} model and 70\% for the  \textit{Meteo+Lat/Long/Alt} model, had significantly worse performance in comparison to the \textit{LSTM\_conus} model (Table \ref{tab:wilcoxon_results}, Figure S17). When examining the effect of purely adding more inputs beyond meteorology and location while keeping training data the same, we see substantial performance gain and reduction in bias from including either river discharge (median site RMSE drop from 1.78°C to 1.57°C) or GAGES-II attributes (median site RMSE drop from 1.78°C to 1.47°C) individually. Notably, when adding just the GAGES-II attributes, only 55\% of the sites had significantly worse performance relative to the \textit{LSTM\_conus} and the average performance drop for those sites was the lowest compared to all other input combinations  ($\Delta$ RMSE=0.26°C; Table \ref{tab:wilcoxon_results}). This suggests that the model with the GAGES-II attributes that does not have streamflow as an input can be comparable to the best performing \textit{LSTM\_conus} model.

Surprisingly, the inclusion of additional stream temperature observations for enhanced training does not result in substantial performance improvements for any of the input combinations, indicating a saturation in performance for the training datasets. For example, the \textit{Meteo\_only} and \textit{Meteo+Lat/Long/Alt} models, we do not see an improvement in RMSE and bias after expanding the training dataset from approximately 3.82 million observations across 782 sites to about 6.13 million observations across 1346 sites. We see a small decrease in median site RMSE with the enhanced training for the models that included either discharge (median per-site RMSE drop from 1.57°C to 1.50°C) or the GAGES-II attributes (median per-site RMSE drop from 1.47°C to 1.43°C). The extended training for the GAGES-II attributes also results in model performance that is comparable to \textit{LSTM\_conus} in terms of overall RMSE  (1.43°C versus 1.46°C respectively) but slightly worse for the warmest 10\% of observations (1.71°C versus 1.64°C respectively). We also note all models have a similar, slight cold bias, with the median bias ranging from -0.10°C and -0.29°C across the eight models. Additional results for Experiment 2 are included in the supplementary information (Tables S10, Figures S17-18, Figures S21-22).

\begin{table}[h!]
\centering
    \scalebox{0.75}{
\begin{tabular}{|>{\centering\arraybackslash}p{.11\linewidth} 
                 >{\centering\arraybackslash}p{.12\linewidth}
                 >{\centering\arraybackslash}p{.08\linewidth}
                 >{\centering\arraybackslash}p{.11\linewidth} |
                 p{.15\linewidth} 
                 p{.13\linewidth} 
                 p{.14\linewidth} 
                 p{.13\linewidth} p{.10\linewidth}|} 
 \hline
 \multicolumn{4}{|c|}{Included Inputs} &  &  & & & \\ [0.5ex] 
  
 \hline
 Meteorology & Latitude, Longitude, Elevation & Discharge & GAGES-II attributes & n\_obs/n\_sites (training) & Median site RMSE (std\_dev) ($^{\circ}$C) &  Median site Bias (std\_dev) ($^{\circ}$C) & Median site RMSE (std\_dev) Warmest 10\% ($^{\circ}$C) & n\_sites $<$ 2$^{\circ}$C RMSE \\ [0.5ex] 
 \hline\hline
 \checkmark & &  & & 3.82mil/782 & 1.77($\pm$1.04) & -0.29($\pm$1.36) & 2.18($\pm$1.68) & 351 \\ 
 \rowcolor{gray!40}\checkmark & &  & & 6.13mil/1346 & 1.80($\pm$1.00) & -0.28($\pm$1.32) & 2.23($\pm$1.62) & 356 \\ 
 \hline\hline
 \checkmark & \checkmark & & & 3.82mil/782 & 1.78($\pm$1.19) & -0.17($\pm$1.40) & 2.14($\pm$1.97) & 333 \\
  \rowcolor{gray!40}\checkmark & \checkmark & & & 6.13mil/1346 & 1.77($\pm$1.02) & -0.10($\pm$1.29) & 1.99($\pm$1.69) & 349\\

 \hline\hline
  \checkmark & \checkmark & \checkmark & & 3.82mil/782 & 1.57($\pm$1.10) & -0.14($\pm$1.29) & 1.76($\pm$1.79) & 383\\
  \rowcolor{gray!40}\checkmark & \checkmark & \checkmark & & 4.65mil/1048 & 1.50($\pm$1.11) & -0.19($\pm$1.28) & 1.72($\pm$1.82) & 399 \\
 \hline\hline
  \checkmark & \checkmark & & \checkmark & 3.82mil/782 & 1.47($\pm$1.03) & -0.15($\pm$1.18) & 1.68($\pm$1.65) & 406\\
  \rowcolor{gray!40}\checkmark & \checkmark & & \checkmark & 4.13mil/833 & \textbf{1.43($\pm$1.05)} & -0.17($\pm$1.18) & 1.71($\pm$1.69) & 422\\
 \hline\hline
 \checkmark & \checkmark & \checkmark & \checkmark & 3.82mil/782 & 1.46($\pm$1.08)  & -0.17($\pm$1.29) & \textbf{1.64($\pm$1.79)} & 406\\
 \hline
\end{tabular}}
\caption{Performance of continental-scale models with specific categories of inputs removed including GAGES attributes, discharge data, and location and elevation data for 580 test sites. The final row is identical to the \textit{LSTM\_conus} model from Experiment 1. Rows with grey backgrounds indicate an extended training dataset with observations previously discarded due to lack of either discharge data or GAGES attributes.}
\label{tab:exp2_overall}
\end{table}

\begin{table}[]
    \centering
    \begin{tabular}{|p{.28\linewidth} p{.10\linewidth} p{.10\linewidth} p{.10\linewidth} p{.10\linewidth} p{.10\linewidth} p{.10\linewidth}|}
        \hline
         Model & \multicolumn{2}{c}{Significant (Better)} & \multicolumn{2}{c}{No Significance} & \multicolumn{2}{c|}{Significant (Worse)} \\
         \hline
          & n\_sites & $\Delta$ RMSE & n\_sites & $\Delta$ RMSE & n\_sites & $\Delta$ RMSE \\
         \hline
         \textbf{Experiment 1} & & & & & & \\
         \textit{LSTM\_regional} & 135 & -0.41 & 56 & 0.02 & 389 & 1.36 \\
         \textit{LSTM\_cluster} & 123 & 0.35 & 48 & -0.02 & 407 & 1.23 \\
         \textit{MTL} & 117 & -0.63 & 48 & 0.06 & 415 & 0.89 \\
         \hline
         \textbf{Experiment 2 Default Training Data} & & & & & &\\
         \textit{Meteo only} & 145 & -0.64 & 67 & 0.03 & 368 & 0.70 \\
         \textit{Meteo + Lat/Long/Alt} & 165 & -0.58 & 63 & 0.05 & 352 & 0.87 \\
         \textit{Meteo + Lat/Long/Alt + Flow} & 247 & -0.56 & 64 & 0.06 & 269 & 0.66 \\
         \textit{Meteo + Lat/Long/Alt + GAGES-II} & 226 & -0.26 & 88 & 0.01 & 266 & 0.28 \\
         \hline
         \textbf{Experiment 2 Extended Training Data} & & & & & & \\
         \textit{Meteo only (extended data)} & 141 & -0.71 & 52 & 0.03 & 387 & 0.71 \\
         \textit{Meteo + Lat/Long/Alt (extended data)} & 170 & -0.62 & 60 & 0.04 & 350 & 0.77 \\
         \textit{Meteo + Lat/Long/Alt + Flow (extended data)} & 247 & -0.55 & 64 & 0.04 & 269 & 0.69 \\
         \textit{Meteo + Lat/Long/Alt + GAGES-II (extended data)} & 237 & -0.29 & 83 & 0.01 &  260 & 0.29 \\
         \hline
         \textbf{Experiment 3} & & & & & & \\
         \textit{LSTM\_conus (expert-selected)} & 260 & -0.51 & 65 & 0.01 & 255 & 0.56 \\
         \textit{LSTM\_conus (z-score-based)} & 254 & -0.42 & 74 & -0.01 & 252 & 0.40 \\
         \hline
    \end{tabular}
    \caption{Significance counts of models for Experiments 1, 2, and 3 compared to the LSTM\_conus baseline using a two-sided Wilcoxon signed-rank test ($\alpha = 0.05$) for the 580 test sites. $\Delta$ RMSE indicates the difference in RMSE calculated by subtracting the baseline model RMSE from a given model RMSE, averaged over all sites in the respective category.}
    \label{tab:wilcoxon_results}
\end{table}

\subsubsection{Experiment 2 Feature Importance Results}
\label{subsubsec:results:exp2:featimp}
The feature importance values for all models in Experiment 2 also indicate that air temperature is the primary input influencing stream temperature predictions  (with a 4-5$^{\circ}$C importance, see Figures S5-S12 in the supplementary information). Both meteorology-only models have significant but less than 0.5$^{\circ}$C importance for solar radiation, vapor pressure, precipitation, and day length (in descending order of importance). For the models that add in the latitude, longitude, and altitude we see that altitude is the most important feature of the three, with an importance of 0.41$^{\circ}$C and 0.43$^{\circ}$C for the default and extended training data versions respectively. Other meteorological variables besides air temperature drop significantly in importance when these three variables are added in both cases. When discharge is added as an additional requirement without the GAGES-II attributes, its importance is between 0.74-76$^{\circ}$C for both models, with the log-transformed version being utilized far more than the raw version. When the GAGES-II attributes are added as a requirement without discharge, the total importance of the features are 1.11$^{\circ}$C for the model using the default training data and 0.54$^{\circ}$C for the model using the extended training data. However, no other single attribute was noticeable besides a small (0.31$^{\circ}$C) importance of slope percentage for the extended version. 

\subsection{Experiment 3}
\label{subsec:results:exp3}
Experiment 3 compared three different representations of catchment attributes with the \textit{LSTM\_conus} model using all 274 GAGES-II attributes and two other models using (i) features defined through a network-based clustering technique to reduce redundancy in the attributes, and (ii) a set of features that were selected by domain experts. 

\subsubsection{Experiment 3 prediction performance}
\label{subsubsec:results:exp3:perf}

\begin{figure}[h!]
\centering
\includegraphics[width=\textwidth]{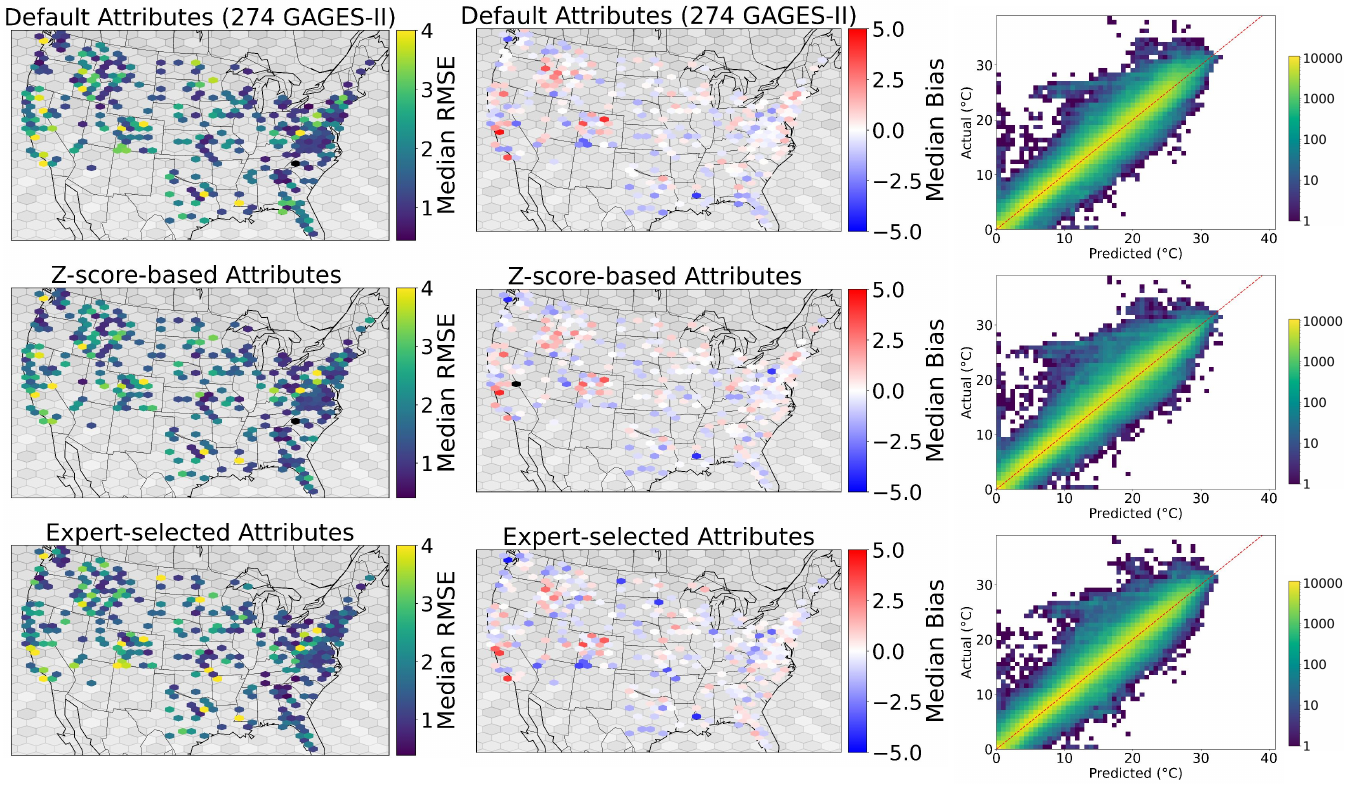}
\caption{9 panel plot comparing different representations of attributes, showing the spatial distribution and prediction performance of each model. The top row shows the default \textit{LSTM\_conus}, and the second and third rows show the models with the original attributes swapped for z-score-based attributes from \citeA{ciulla2023data} and expert-selected 21 GAGES-II attributes from \protect\citeA{rahmani2021deep} respectively. The left and middle columns show the hexbin spatial distribution of per-site RMSE and mean bias values in $^{\circ}$C, where the color represents the median within each hexbin. The right column is a two dimensional histogram showing the distribution of individual stream temperature predictions across all sites. The color represents the count within each bin.}
\label{fig:exp3_9panel}
\end{figure}

The results for Experiment 3 demonstrate that the default model (\textit{LSTM\_conus}) using all 274 GAGES-II attributes has the lowest median per-site RMSE of 1.43$^{\circ}$, but overall the performance is quite similar between methods (Table \ref{tab:exp3_overall}). Surprisingly, the model using the expert-selected attributes also performs well with a median RMSE of 1.46$^{\circ}$, and even has a slightly lower median RMSE for the warmest 10\% of observations  relative to the \textit{LSTM\_conus} model (1.64$^{\circ}$C versus 1.70$^{\circ}$C). The model using the z-score-based also has comparable performance (median per-site RMSE= 1.47$^{\circ}$). Based on the per-site RMSE, we find the best performing model to be the one with the default attributes for 186/580 test sites, the z-score-based attributes for 176/580 test sites, and the expert-selected attributes for 218/580 test sites. In general, spatial distributions of RMSE across sites are similar between the methods, although there are some differences in bias (Figure \ref{fig:exp3_9panel}. The number of sites that are significantly worse than the (\textit{LSTM\_conus}) are also split nearly equally across the different attribute combinations (Table \ref{tab:wilcoxon_results}), with the z-score approach having a lower $\Delta$ RMSE compared to the expert selected set of attributes. Hence, none of the three models is clearly  superior to the others. Additional results for Experiment 3 are included in the supplementary information (Tables S11, Figure S19 and S23).

\begin{table}[h!]
\centering
\begin{tabular}{|p{.25\linewidth} p{.22\linewidth} p{.16\linewidth} p{.16\linewidth} p{.1\linewidth}|} 
 \hline
 Input Set & Median per-site RMSE ($^{\circ}$C) (std\_dev) & Bias ($^{\circ}$C) & RMSE (warmest 10\%) ($^{\circ}$C) & n\_sites $<$ 2$^{\circ}$C RMSE\\ [0.5ex] 
 \hline
  Default (\textit{LSTM\_conus}; 274 GAGES-II attributes)& \textbf{1.43($\pm$0.98)} & -0.09($\pm$1.22) & 1.70($\pm$1.58) & 406 \\ \hline
  Expert-selected (21 GAGES-II attributes from \citeA{rahmani2021deep}) & 1.46($\pm$1.08)  & -0.17($\pm$1.29) & \textbf{1.64($\pm$1.79)} & \textbf{419}\\ \hline
 Aggregated Z-score  (25 attribute categories from \citeA{ciulla2023network})& 1.47($\pm$1.00) & -0.09($\pm$1.21) & 1.65($\pm$1.66) & 417\\

 \hline
\end{tabular}

\caption{Summary of performance metrics for the \textit{LSTM\_conus} model using three different sets of catchment attributes in Experiment 3. The 'Default' model includes all 274 GAGES-II attributes, the ‘Expert-selected’ model uses a subset of 21 attributes based on expert knowledge, and the ‘Z-score’ model employs 25 attribute categories derived from a network-based clustering method. Metrics include the median per-site RMSE, bias, RMSE for the warmest 10\% of observations, and the number of sites achieving RMSE below 2°C. Bold values represent the lowest errors for each performance metric.}
\label{tab:exp3_overall}
\end{table}

\subsubsection{Experiment 3 Feature Importance}
\label{subsubsec:results:exp3:featimp}

The feature importance results (Section \ref{subsubsec:results:exp1:featimp} for the default model, and Figures S13-S14 for the z-score and expert-selected attributes) show similar meteorological and streamflow dependencies as in the previous experiments. The z-score attribute model had a combined 0.77$^{\circ}$C importance for the attributes, with the category representing "Major Dams" being the most important at 0.16$^{\circ}$C, and lower importance values for other categories including "Higher Order Streams", "Lakes, Ponds, and Reservoirs", and "Shrublands" ranging from 0.03$^{\circ}$C to 0.06$^{\circ}$C. For the expert-selected model, the combined attribute importance was 0.66$^{\circ}$C, with the most important individual attributes being percent slope (PCT\_SLOPE) and drainage area (DRAIN\_SQKM) at 0.10$^{\circ}$C and 0.08$^{\circ}$C respectively. Less important attributes between 0.03$^{\circ}$C and 0.06$^{\circ}$C were maximum basin air temperature (T\_MAX\_BASIN), mean basin elevation (ELEV\_MEAN\_M\_BASIN), and straightline distance to nearest dam (RAW\_DIST\_NEAREST\_DAM).

\begin{figure}[h!]
\centering
\includegraphics[width=\textwidth]{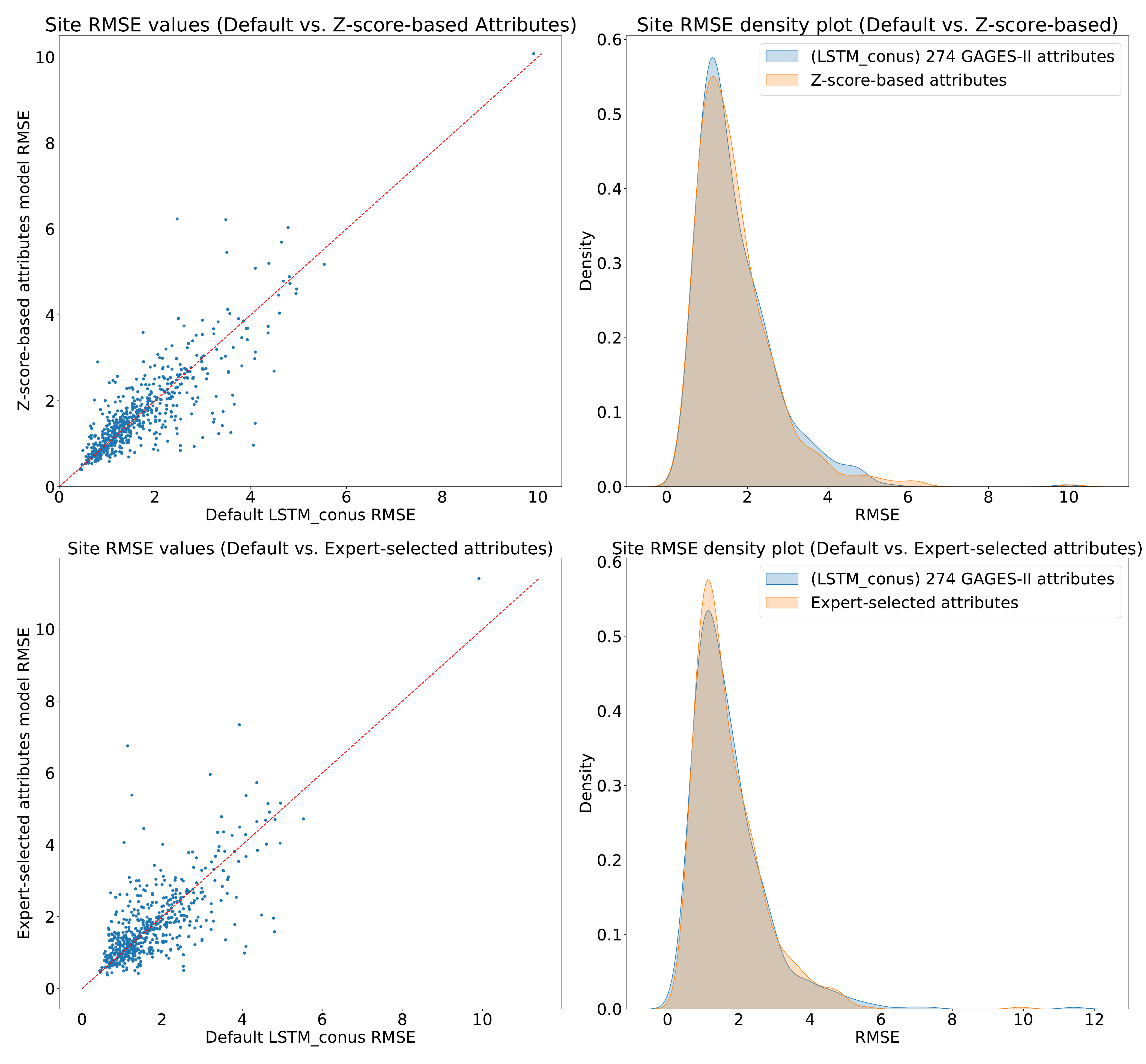}
\caption{Per-site RMSE comparisons between the models using the expert-selected attributes  and the z-score attribute categories with the default \textit{LSTM\_conus} using 274 GAGES-II attribute feature set.}
\label{fig:exp3_4panel_site_rmses}
\end{figure}

\subsubsection{Additional Error Analysis for Experiment 3}
Given the inconclusive results in Experiment 3, we conducted additional error analysis using a larger ensemble by expanding the set of model realizations from 5 to 20 to determine if increasing training ensemble size would lower errors. We found a slight improvement in model performance for all three models with updated median RMSE values of 1.39$^{\circ}$C for the default \textit{LSTM\_conus} model, 1.46$^{\circ}$C for the model using z-score attribute categories, and 1.41$^{\circ}$C for the expert-selected attributes. The results below are from further analysis conducted using the 20-member ensembles to explain the variance in per-site median errors for Experiment 3. A more detailed comparison of RMSE, bias, and performance on the warmest 10\% of data between the ensemble sizes are presented in the supplementary information (Table S12). 

In Figure \ref{fig:exp3_4panel_site_rmses}, we see that although the distribution of errors across sites (density plots in right panel) is roughly the same, there is still a large spread across locations, indicating that certain sites have better performance depending on which set of inputs are selected. For 97 of the 580 test sites, the standard deviation of the RMSE across the three methods is greater than 0.5$^{\circ}$C indicating significant performance differences, which is the same result as observed for the ensemble size of five model realizations. 

\begin{figure}[h!]
\centering
\includegraphics[width=0.8\textwidth]{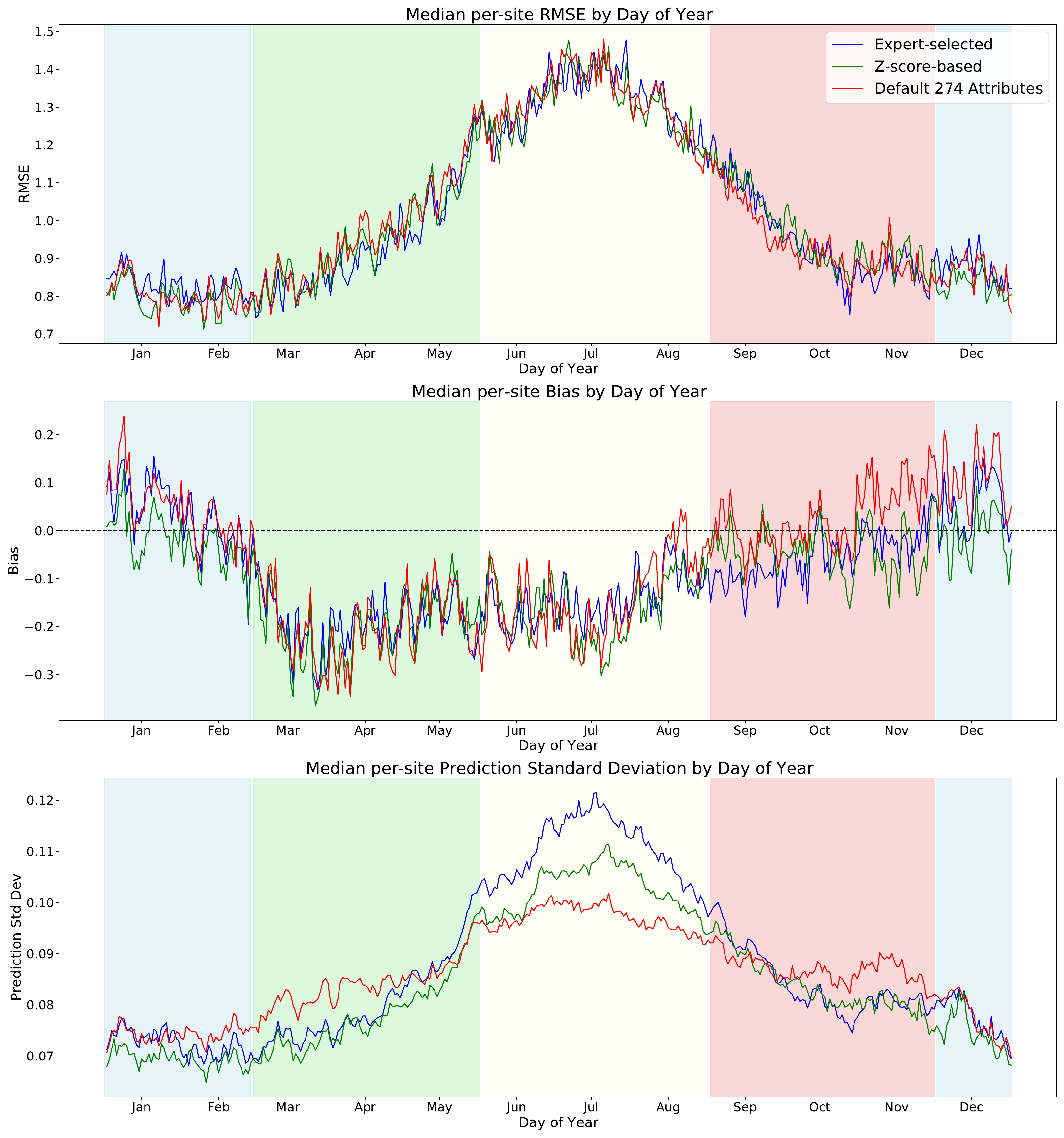}
\caption{Model performance per season and day of the year for the three models in Experiment 3 calculated as the per-site median RMSE for all sites that have data for each day. Background colors indicate seasons as  winter (blue), spring (green), summer (yellow), and autumn (red). Values are averaged over all 580 sites in the testing dataset.}
\label{fig:exp3_time_series_3panel}
\end{figure}

\begin{figure}[h!]
\centering
\includegraphics[width=\textwidth]{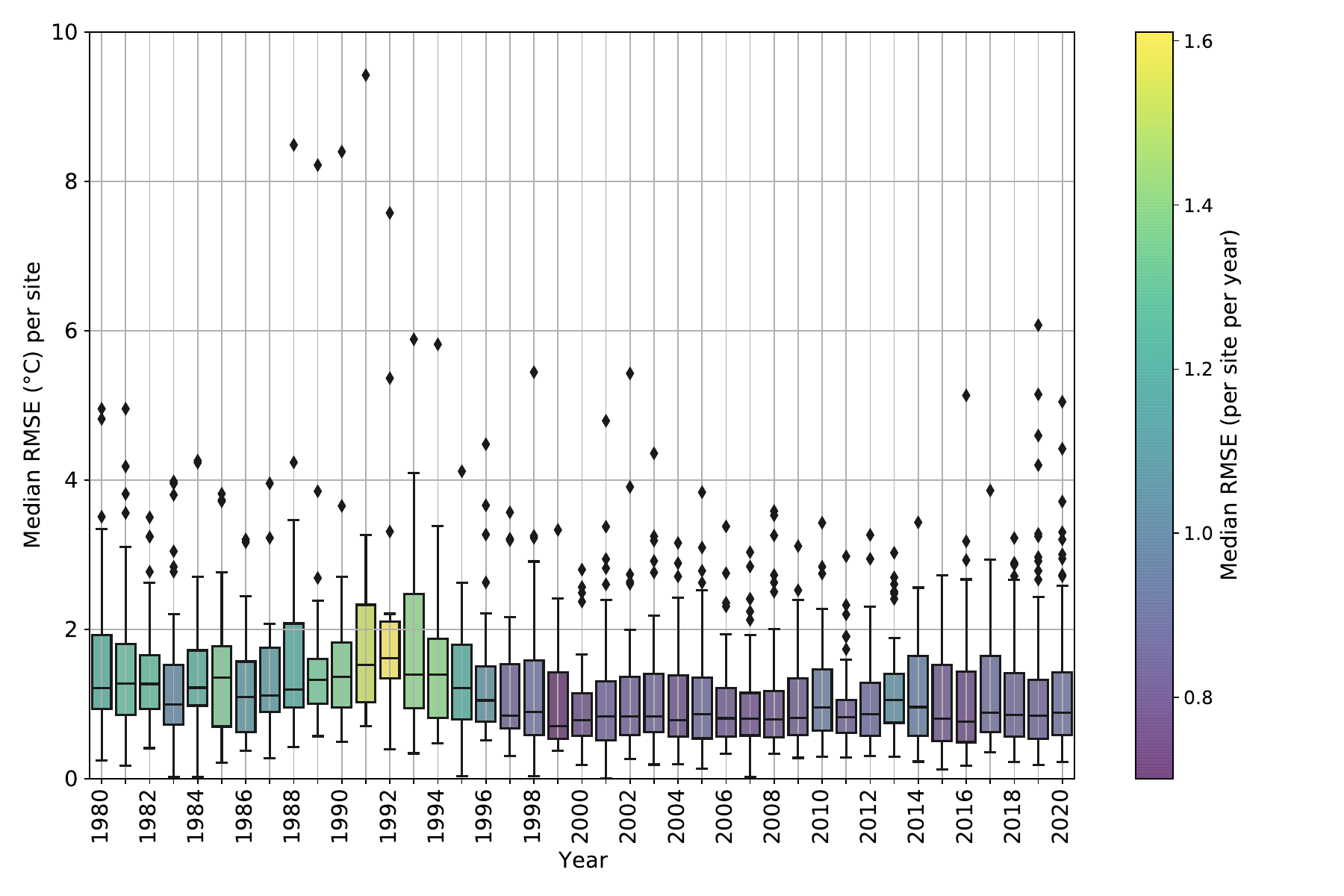}
\caption{Box plot of median per-site errors of the \textit{LSTM\_conus} model built using 274 GAGES-II attributes from 1980-2021. The 580 test sites contain a total 2236 total site years. Not pictured are the following outlier site years, huc\_id=03289000 had a 16.43 RMSE in 2012, huc\_id=13338500 had a 10.39 RMSE in 2020, and huc\_id=06200000 had RMSEs of 11.02 and 11.58 in 2018 and 2019 respectively.}
\label{fig:exp3_err_per_year}
\end{figure}

For all three models, prediction performance across the days of the year and different seasons is lowest in the summer period, which has the highest RMSE and standard deviations (Figure \ref{fig:exp3_time_series_3panel}). This mirrors the results from all the experiments where the highest 10\% of temperature values had greater RMSE values relative to overall performance. There is also an overall cold bias, most noticeable in the spring and summer across all models usually between -0.2 and -0.4 $^{\circ}$C. We also find that median per-site errors are lower for more recent years in the 1980-01-01 to 2021-06-30 prediction period, particularly after 1995 (Figure \ref{fig:exp3_err_per_year}).

\clearpage

\subsubsection{Attribute Error Analysis}

\begin{figure}[h]
\centering
\includegraphics[width=\textwidth]{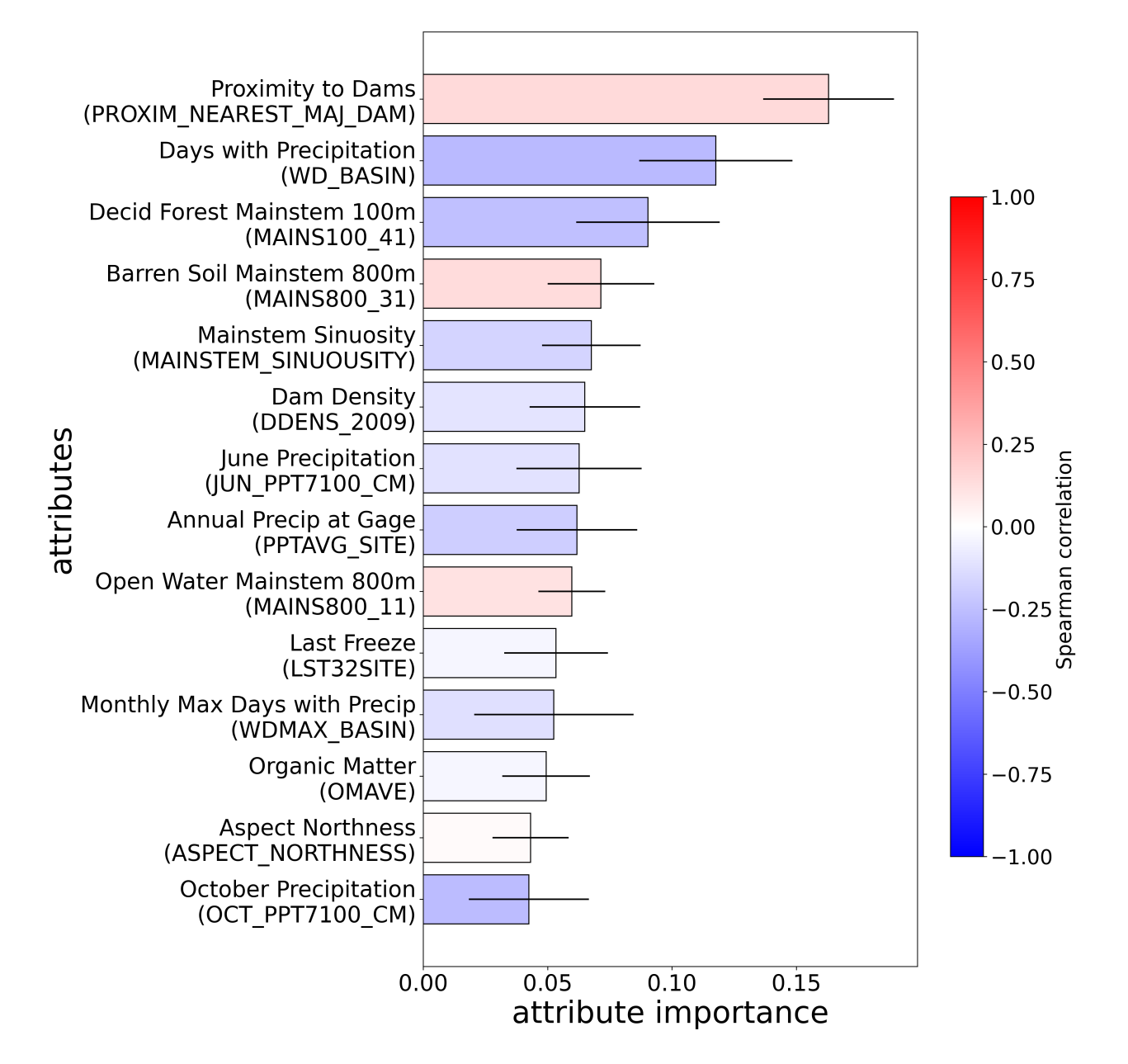}
\caption{Plot of the average feature importance, computed as Shapley values, for the top 14 attributes that are related to \textit{LSTM\_conus} model errors in Experiment 3. The color of the bars represents the Spearman correlation between the Shapley value of each attribute and the RMSE values. Black error lines are the standard deviation of 100 realizations of the XGBoost model.}
\label{fig:exp3_trait_error}
\end{figure}

The best predictions for the XGBoost model used to infer the attribute importance for the \textit{LSTM\_conus} model errors in Experiment 3 are achieved using the 14 attributes displayed in Figure \ref{fig:exp3_trait_error} ($RMSE=0.86$ and $R^2=0.19$). The primary attributes that have a weak positive correlation of the Shapley values with the RMSEs of the \textit{LSTM\_conus} model include: proximity to dams (Shapley value=0.16, Spearman correlation=0.15), the percentage of barren land defined in the National Land Cover dataset \cite{multi1992national} as "areas of bedrock, desert pavement, scarps, talus, slides, volcanic material, glacial debris, sand dunes, strip mines, gravel pits and other accumulations of earthen material" within a 800 m buffer of the stream mainstem (Shapley value=0.07, Spearman correlation=0.13), and percentage area of open water bodies with less than 25\% vegetation within an 800m  buffer of the stream mainstem (Shapley values = 0.06, Spearman correlation= 0.12). Conversely, the attributes that correspond to a weak negative correlation of the Shapley values with the RMSE of the \textit{LSTM\_conus} model are days with precipitation (Shapley values = 0.12, Spearman correlation= -0.27), the percent area of deciduous forests with in a 100m buffer of the stream mainstem (Shapley values = 0.09, Spearman correlation= -0.25), and the mainstem sinuosity (Shapley values = 0.07, Spearman correlation= -0.17). The attribute importance plots of the other two models in Experiment 3 also show a positive correlation of the Shapley values with the RMSE of proximity to dams, dam size and barren land as well as negative correlation of attributes related to high precipitation, river sinuosity and extent of deciduous forest cover in the catchment on model accuracy (Figure S24).


\subsubsection{Paired Air-Stream Temperature Analysis}
\label{subsubsec:groundwater_results}
Training and test sites were classified using paired air-stream temperature analysis (Section \ref{subsubsec:groundwater_analysis}). For the 580 testing sites analyzed, 50.5\% of sites were atmospherically driven, 5.0\% of sites were impacted by shallow groundwater, 7.8\% of sites were impacted by deep groundwater, and 36.7\% of sites were considered dam impacted. The proportion of sites in the atmospheric, shallow groundwater, deep groundwater, and dammed classes were consistent between training and test sites, with less than 4\% variation (Figure S15).

\begin{figure}[h]
\centering
\includegraphics[width=0.85\textwidth]{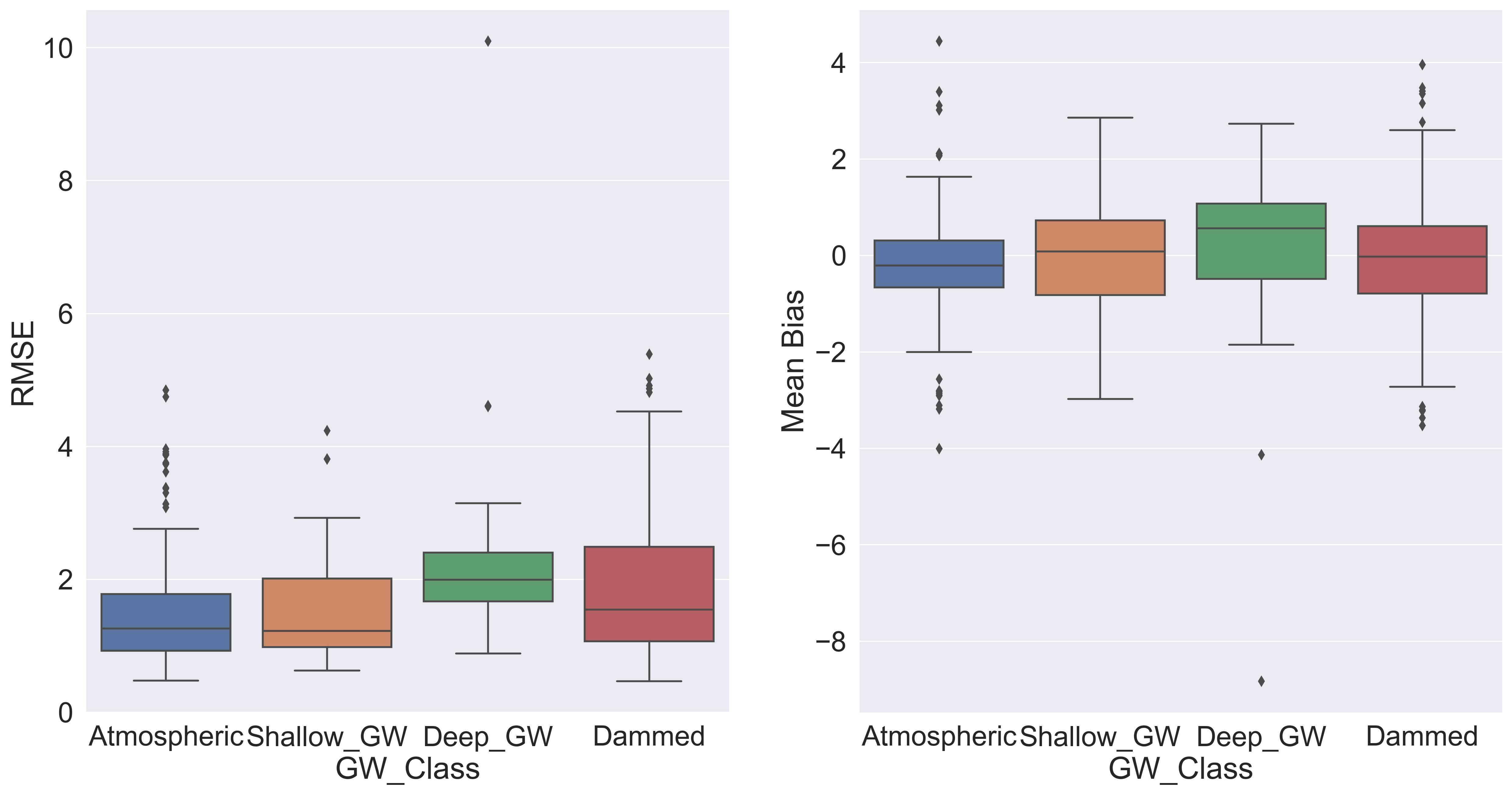}
\caption{Model RMSE (left) and mean bias (right) for the \textit{LSTM\_conus} model for the four classes identified using the paired air-stream temperature analysis.}
\label{fig:groundwater_rmse_bias}
\end{figure}

The median RMSE and mean bias show consistent results for the four air-stream temperature classes across the three different models in Experiment 3 (Table S13). For the \textit{LSTM\_conus} model, sites in the deep groundwater class had the highest median RMSE of 2.00 \textdegree C. This median RMSE (2.00 \textdegree C) is at least 0.45 \textdegree C higher than the all other classes (Fig \ref{fig:groundwater_rmse_bias}a). Dam-impacted sites have the second highest errors (median RMSE=1.55 \textdegree C), while the atmospheric-driven sites and shallow groundwater sites had similar median RMSE (1.26 \textdegree C and 1.22 \textdegree C respectively). In general the classes with the most sites (atmospheric-driven and dammed) had higher standard deviation in comparison to the smaller classes of shallow and deep groundwater impacted sites. Deep groundwater sites also had the highest mean warm bias (+0.56 \textdegree C), while the atmospheric class had a smaller mean cold bias (-0.21 \textdegree C, Fig \ref{fig:groundwater_rmse_bias}b). Additional median RMSE and mean bias metrics for the paired air-stream temperature classes across all the three models in Experiment 3 are presented in Table S13).

\section{Discussion}
\label{sec:disc}

\subsection{Superior performance of top-down approaches}

This work is the first to systematically compare top-down, bottom-up, and grouped modeling approaches for PUBs using deep learning approaches. While the top-down and grouped approaches follow the widely-used implementations of LSTMs for PUB modeling in hydrology \cite{kratzert_toward_2019, rahmani2021deep}, here we use a novel method based on meta transfer learning \cite{willard_predicting_2021}. This method makes predictions in unmonitored locations by transferring information from multiple source models built for similar sites, where site similarity is determined using the meta-features (Table S1). This ML-based bottom-up method is conceptually analogous to the traditional regionalization approach of building process-based models at a few intensively-monitored sites and extrapolating model parameters to unmonitored sites based on hydrological similarity or co-location \cite{guo2021regionalization}.  

Our results from Experiment 1 clearly demonstrate that the top-down modeling approach trained using data from all available sites significantly outperforms the transfer-based bottom-up and grouped modeling approaches. This superior performance is evident when comparing median per-site RMSE across all sites and also within the vast majority of regions and clusters (Tables \ref{tab:exp1_overall}, \ref{tab:regional_results}, \ref{tab:cluster_results}). Additionally, comparisons of the significance test results and $\Delta$ RMSE values in Table \ref{tab:wilcoxon_results} show that the errors in the \textit{LSTM\_regional} and \textit{MTL} are much worse relative to the \textit{LSTM\_conus} at locations where they underperform (also see Figure S16). Alternatively, for the ~20\% of sites  where the \textit{LSTM\_conus} underperforms the other models, the $\Delta$ RMSE values are much smaller, indicating that use of the top-down model at these locations only results increase in a minor increase in errors. 

Overall, these results suggest that deep learning models that leverage a diverse, large-sample dataset are more effective for predicting stream temperatures in unmonitored basins than any grouped or bottom-up approach. This indicates their ability to synthesize diverse information from several sites into a single model and produce realistic predictions for a large range of conditions . This result complements findings from prior studies predicting streamflow and soil moisture at monitored sites that also show the LSTMs benefiting from being trained on large, diverse datasets in comparison to training with smaller groups or with data of the same size but from regions that are more similar to the target sites \cite{fang_data_2022, kratzert2024hess}.  Another possibility for the poorer performance of the bottom-up MTL models is that LSTMs built for individual sites can be less accurate than top-down models trained on those same sites since the  latter has the ability to discriminate patterns in the large-sample, diverse training dataset  \cite{kratzert2024hess}. Thus transfer of these single-site LSTMs  by the XGBoost metamodel, which in the MTL framework is the part that handles the inter-site differences, may generally result in lower performance except in the cases where the single-site LSTMs outperform the top-down models. 

In our prior survey, we only found  three other studies that predict daily stream temperatures at unmonitored sites \cite{willard2023water}. A comparison of the performance of the \textit{LSTM\_conus} with these models show that its errors are comparable. The closest comparison to our study is that of \citeA{rahmani2021deep}, which uses a similar LSTM model for unnmonitored predictions at the continental scale, including both pristine and dammed sites. They report a median per-site RMSE of 1.13$^{\circ}$C, but their analysis tests fewer sites (n=40) from 2010-2016, while our study tests performance at 580 sites over approximately 40 years of data (Jan 01 1980 - June 30 2021). Our results also indicate better performance in more recent years (median RMSE=1.39$^{\circ}$C between 2010-2016 for 138 sites), partly explaining the performance difference. The continental-scale deep learning models also perform comparably to regional studies that used simpler models for unmonitored predictions. For example, \citeA{ouarda2022regional} used a statistical generalized additive model (GAM) that had median errors of 1.3$^\circ$C, 1.7$^\circ$C, and 1.9$^\circ$C for three regions in Eastern Canada. In \citeA{weierbach2022stream}, we compared the use of  multilinear regression (RMSE= 1.2$^{\circ}$C), support vector regression (RMSE=0.89$^{\circ}$C) and XGBoost models (RMSE=0.61$^{\circ}$C) for monthly unmonitored predictions in the Pacific Northwest (n=54 total sites) and mid-Atlantic regions (n=24 total sites) using a random subsampling approach over a 100-member ensemble that evaluated median RMSE for each individual test site. In comparison, the median RMSEs of the \textit{LSTM\_conus} for these two regions are 1.22$^{\circ}$C (n=208 total sites, n=70 test sites) and 1.27$^{\circ}$C  (n=212 total sites, n=121 test sites) respectively but for daily instead of monthly averages used in the prior study. 



Our findings also raise the issue of whether the data-driven, top-down modeling approach should be favored over traditional regionalization approaches that extrapolate parameters of localized process-based models to unmonitored sites, for applications where the primary purpose is to obtain the most accurate predictions (as opposed to other scientific applications such as testing hypotheses). We argue that the ML-based analog presents a best case scenario for regionalization for two reasons. First, numerous studies have demonstrated that LSTM models built for individual catchments are more accurate than process model counterparts for temporal predictions in monitored scenarios \cite{kratzert2018rainfall,ley2023intercomparing}. Secondly, the meta-transfer learning framework provides the ability to transfer and learn from multiple source models (n=782 in our study) using hundreds of features and past performance metrics. This could present a significant improvement over traditional approaches that group similar catchments based on much smaller sets of features \cite{rao2006regionalization_hyb,rami2022mixed}. We suggest that further comparisons of traditional approaches that use regionalization of site-specific, process-based models with top-down LSTM models should be conducted to evaluate the best method for making predictions at unmonitored sites. 

\subsection{Optimal selection of inputs considering tradeoffs between data availability, model complexity and performance}
Given the superior performance of top-down models, the question remains of how much data is needed to train an accurate  model that can be broadly applied to make predictions in any stream reach. Notably, the requirement of including stream discharge as an input to the \textit{LSTM\_conus} drastically reduces its applicability to gauged sites, which are also sparse (e.g., the approximately 11,800 gauges in the United States only monitor 1\% of stream reaches \cite{read2017water,rahmani2021deep}). In Experiment 2, we explored reducing input data requirements to make the model more generalizable to other regions as well as to increase the volume of data available for training. For example, meteorological data are universally available for most regions with acceptable resolution and accuracy from gridded products such as Daymet. Thus models that only need meteorological inputs can utilize the largest training dataset and be applied to obtain predictions any site, given the spatial resolution of the input meteorological product. In Experiment 2, the simpler \textit{Meteo\_only} models had a moderate 20-25\% increase in RMSE (Table \ref{tab:exp2_overall}) compared to the more complex models that also included discharge and watershed attributes, which may be acceptable performance for some applications. However, if higher accuracy is desired, we find that the models that dropped discharge or the GAGES-II attributes individually have comparable performance to the \textit{LSTM\_conus}. Notably, the LSTM can clearly perform well without streamflow inputs as evidenced by the results from the 2-sample statistical text comparing these models (Table \ref{tab:wilcoxon_results}, Figure S18) and the permutation feature importances (Figures S11, S12). This is a surprising but significant result, since streamflow has been shown to influence stream temperatures \cite{isaak2017norwest,leach2023primer}, but can help reduce input data constraints for stream temperature ML models more broadly. This could happen because air temperatures are the primary driver for water temperature dynamics for the majority of our testing sites (Section \ref{subsubsec:groundwater_analysis}), which is also reflected in the feature importance results where discharge variables only have small values (Figure \ref{fig:pfi_conus}). An additional reason may be the model's ability to learn streamflow behavior from meteorology and the GAGES-II attributes, as has been demonstrated in various studies predicting streamflows using LSTMs \cite{kratzert2019toward_ung,arsenault2023continuous,ayzel2020streamflow,koch2022long}.

Poor performance due to lack of data was most apparent in the grouping approach where certain regions or clusters of similar sites have clearly insignificant data for training a generalizable model. When only a few sites are available for training, we see that the grouped models do quite poorly in all regions with less than 15 training sites (Table \ref{tab:exp1_overall}). We believe the models are not seeing sufficient data and site diversity in these cases to be able to extrapolate to unseen locations. We note that there are ideas to alleviate this issue, for example by fine-tuning the top-down model to data from each target region \cite{oruche2021transfer,zhu2024advancing}. However, we found this technique yielded performance that mirrored the top-down model very closely when applied to both the grouped modeling and bottom-up approaches. Conversely, in the cases where an extended training dataset was available by dropping discharge or streamflow as inputs, there was surprisingly only a marginal improvement in performance (Table \ref{tab:exp2_overall}). This suggests that while increasing data volumes can partially offset the absence of certain input types, its effects may be limited beyond a certain training dataset size. 

The findings from this study underscore the need for careful consideration of training dataset construction when designing deep learning models. Ensuring broad applicability across diverse regions while maintaining a high level of accuracy requires balancing the number inputs selected with the scale of available data. To our knowledge, no other study has analyzed how performance varies based on the data availability of desired inputs to the model. Ultimately, the trade-offs between data availability and input selection must be considered within the context of the specific goals of the prediction task. For broad-scale applications, simpler models using universally available data may suffice, while more accurate predictions may require additional inputs. Importantly, our results from Experiments 2 and 3 suggest that it is not necessary to have the most complex model possible that includes all available data to make the most accurate predictions.

\subsection{Which site attributes have the most predictive power?}
Our study also highlights the complex relationship between the choice of static catchment attributes, model complexity, and overall performance. Here, model complexity increases as the number of attributes increases even if hyperparameters remain constant across experiments. This is because the number of weights connecting the input layer to the first hidden layer in the neural network will also increase, thereby adding more parameters. From Experiment 1, we see that using comprehensive input data at a continental scale generally enhances accuracy across regions and different clustered groupings of similar sites. However, Experiment 2 shows the selection of input features plays a significant role in predictive performance. Feature importance scores further support this, showing that while air temperature remains the most critical driver of stream temperatures, the combined set of catchment attributes also contributes significantly. However, the equivalent results from the three configurations in Experiment 3 indicate that we still cannot determine which individual features are the most important to include as inputs. We also find that the importance of different features is dependent on the size and redundancy of the attribute dataset. For example, individual feature importances of GAGES-II attributes in the expert-selected feature model are diminished when using the full 274 attribute set. This mirrors the findings from stream temperature predictions in \citeA{barclay2023train}, where they expanded inputs to their graph model with a small amount of attributes and found that the model transitioned from relying heavily on a few features to leaning more lightly on each member of a broader, more nuanced set of features. This shift in feature importance is reflective of the complexity introduced when models are tasked with interpreting a larger, more varied set of data.

Notably, the performance of the model using about 10\% of input features (either as individual features or aggregated z-scores) results in equivalent model performance. However, despite the similar performance of the different top-down models using various sets and transformations of site attributes, we observed that prediction accuracy varies across different sites, even as the ensemble of model realizations was expanded. We attribute this inconsistency to the inherent stochasticity of ML optimization, where the optimizer encounters different local minima shaped by the underlying feature space and dimensionality, even when the overall information content remains comparable \cite{sun2020optimization,geiger2021landscape}. Although the expert-selected and aggregated z-score attribute sets are less redundant and more interpretable, they still lead to unique, complex, and non-linear relationships that are difficult to untangle within a deep neural network. Reducing redundancy in the feature space may simplify the optimizer's trajectory across the loss surface during training, allowing more efficient convergence, but this does not always improve performance. These transformations can sometimes increase the number of local minima \cite{dauphin2014identifying}, guiding the optimizer through distinct regions of the loss surface and leading to varying outcomes across different attribute sets, even when the information content is similar.

While ensemble averaging helps to reduce these effects of variance between models, we see across different the performance analyses that it cannot fully eliminate the variability introduced by differing feature spaces. Even with 20 initial weight sets, models still diverge in how they navigate the feature space, especially when inputs exhibit varying levels of redundancy. As a result, different sites may benefit from different input configurations, which we believe is driven by complex, non-linear interactions between features and the site-specific dynamics. This complexity is compounded by additional sources of uncertainty and noise, such as measurement uncertainty in the input and targets, the static treatment of dynamic attributes, and the possibility of outdated information in the GAGES-II dataset. Our analysis of errors by catchment attributes yielded consistent patterns across all three sets of attributes used in Experiment 3 (Section \ref{subsubsec:groundwater_results}), indicating that the complexity of the problem transcends these classifications. To mitigate these issues, introducing more diverse ensemble members that capture different aspects of the feature space, rather than merely varying the initial model weights, could offer a more robust solution. Advanced ensemble techniques, as suggested by \cite{ganaie2022ensemble}, could further reduce variance and enhance the robustness of the models. 

The primary conclusion from our collective set of experiments is that the site attributes are clearly necessary to improve model performance for predictions in unmonitored regions. However, the question of how the attributes should be subset or aggregated for the best results is still an open question.

\subsection{Determining the predictability of different sites: Where and when can we trust the models?}
A key challenge for predictions in unmonitored locations is not having data to evaluate whether a model is performing well at a particular site. Despite the overall success of our top-down modeling approaches, there are specific scenarios where model predictions are poor regardless of the configurations and inputs used. Thus we needed to determine when and where to trust the model predictions based on their attributes, given the absence of direct data for comparison. In this context, both the paired air-stream temperature and attribute-based error analyses are useful tools. By analyzing errors across different sites and associating them with site-specific attributes (e.g., groundwater influence, meteorological conditions, watershed characteristics), we can identify patterns that signal where the model performs reliably and where it may struggle.

Overall, we find that the models perform best at sites where stream temperatures are primarily influenced by air temperatures. One significant challenge for these models is the prediction of stream temperatures in deep groundwater-influenced sites (Figure \ref{fig:groundwater_rmse_bias}). These sites exhibit a median RMSE of 2.00$^\circ$C, with a noticeable warm bias. This aligns with findings from \citeA{barclay2023train}, where they showed stream temperature predictions in  groundwater-influenced reaches of the Delaware River Basin were in general less accurate than air-temperature dominated reaches. They also found that that the predictions in groundwater-dominated reaches could be improved with some physics knowledge such as including additional attributes on subsurface properties or custom loss functions that included the amplitude and phase lags of the air-stream temperature analysis. Such modifications can possibly help improve the performance of  groundwater-influenced sites in our modeling approach as well. The attribute-based error analysis also provides insights into characteristics of sites where the models have poor performance. For instance, sites that are closer to dams, or have a higher fraction of barren lands or open water within 800 m of the river mainstem tend to have higher errors. The negative effect of barren lands or open water near the river on model skill may possibly be due to the presence of groundwater-influenced reaches, fed by infiltration and exchange from lakes or seeps near the mainstem, or from reservoir releases. In general, the combined error analyses indicate poorer prediction performance at dam-impacted sites. Improvement of the models for dam-impacted sites may require additional data such as reservoir release depths and volumes over time.

A potentially related aspect is the finding that certain regions in the mountainous west like the Upper and Lower Colorado (HUCs 14 and 15) and California (HUC 18) are more difficult for the top-down model, having slightly over 2$^{\circ}$C median RMSE. Mountainous headwater catchments can have substantial groundwater contributions to streamflow \cite{Somers2020Nov, Carroll2024May} and also tend to have more reservoirs to manage downstream flows \cite{cao2022novel,sambieni2023spatial}. However, we still see the top-down model is preferred in these scenarios compared with the grouped and bottom-up approaches. We also note there are a number of outliers with abnormally high RMSE values (Figure \ref{fig:boxplot}). We note that the highest RMSE site for all methods is "USGS 09416000 MUDDY RV NR MOAPA, NV", which is a likely hot spring with average annual water temperatures between 27.49-28.75 according to USGS NWIS (\url{https://waterdata.usgs.gov/nwis/inventory/?site_no=09416000&agency_cd=USGS}). In our dataset the water temperatures are on average 7.72$^{\circ}$C higher than the air temperatures on the same days for this site. This is an example of sites that will be predicted poorly regardless of method, unless input data is provided that would cause the model to learn that it is a hot spring.

Our study also confirms that stream temperature predictions are particularly difficult during the summer months with warmer temperatures and associated reservoir releases, which aligns with previous studies \cite{arismendi2014can,topp2023stream}. We note that in general there is a warm bias in the colder months (fall/winter) and a cold bias in the warmer months (spring/summer; Figure \ref{fig:exp3_time_series_3panel}). The former is also likely related to the warm bias observed in groundwater-impacted streams (Figure \ref{fig:groundwater_rmse_bias}). This indicates that model improvements are required to capture hot and cold spots or moments, although in some cases the prediction errors may result from observational artefacts (such as if the measurement were taken at mid-day instead of a daily average or if measurement depth varied substantially across streams). In the absence of detailed metadata and consistent measurement protocols, it would not be possible to distinguish between observational artefacts and model prediction errors.

To conclude, by correlating error patterns with specific attributes, we can infer the conditions under which the model's predictions are reliable, even in regions where monitoring data are not available to determine model performance. However, we also find that the presence of a small amount of monitoring data (between 1-5 years as was available for our test sites) can be highly valuable for evaluating how well the models perform at a new location. Our results provides a framework for water resource managers to anticipate and prioritize interventions or monitoring in regions where the model may be less trustworthy. 

\section{Conclusion}

This study advances the understanding of using deep learning for predictions in unmonitored basins with a set of model experiments with stream temperatures as a case study. First we systematically compare a top-down approach using a single deep learning model leveraging a diverse, large-sample dataset, with bottom-up and grouped modeling approaches. Our findings clearly demonstrate that the top-down approach consistently outperforms the other approaches across most metrics and scenarios. This challenges traditional methods that rely on building localized models with regional extrapolation when the purpose is to obtain the most accurate predictions of specific environmental variables at large scales. We also highlight the importance of input feature selection and data quantity in model performance. In general, we find that the deep learning models can be built with lower complexity, with a moderate but acceptable decrease in skill using fewer inputs. One of the advantages of lowering model complexity would be the ability to make widespread predictions at any stream reach using only meteorological inputs, with a loss of approximately 20-25\% median per-site accuracy. We find that the inclusion of river discharge as an input does not substantially improve model performance, even though streamflows are known to influence stream temperatures. This suggests that deep learning models may be able to infer streamflow behavior from the meteorology and catchment attributes, and that it is not necessary to include all inputs that may influence the target variable of interest. This is also reflected in the inclusion of catchment attributes as inputs, which in general can significantly improve model skill. However, the question of which (sub- or aggregated) set of attributes have the greatest predictive power remains open because their importance depends on the size and redundancy of the dataset. Surprisingly, we find that even if additional data are available, either for training (more stream temperature observations) or as inputs (more catchment attributes), it does not necessarily help improve model performance. The  interplay between these factors emphasizes the need for careful consideration of input data in deep learning models. Our detailed error analysis revealed that the models perform best at sites where stream temperatures are primarily controlled by air temperatures. The models tend to have greater errors for sites influenced by groundwater or dams, and during summer months. In general, the models have a cold bias in summer and spring as well as in groundwater-dominated sites. Our study shows that the top-down deep learning models are highly promising for spatiotemporal predictions of stream temperatures in unmonitored basins, and underscores the need for further model refinement to improve performance for certain sites and time periods.

\section*{Open Research Section}
The code, data, models, model outputs, and results presented in this study are available through the U.S. Department of Energy Environmental System Science Data Infrastructure for a Virtual Ecosystem (ESS-DIVE) data repository (\url{data.ess-dive.lbl.gov}) and can be accessed at https://doi.org/10.15485/2448016.

\section*{Author Contributions (CRediT system for contribution roles taxonomy)
}
\textbf{Jared Willard}: Writing – Original Draft (lead); Investigation (lead); Software (lead); Formal Analysis (lead); Writing – Review \& Editing (equal) Conceptualization (equal); Methodology (equal); Visualization (lead). \textbf{Fabio Ciulla}: Investigation (supporting); Writing - Review \& Editing (supporting); Formal Analysis (supporting); Visualization (supporting) \textbf{Helen Weierbach}: Investigation (supporting). Writing - Review \& Editing (supporting). Formal Analysis (supporting). Software (supporting)  \textbf{Vipin Kumar}: Funding Acquisition (equal); Project Administration (supporting); Resources (supporting); Supervision (supporting). \textbf{Charuleka Varadharajan}:  Data Curation (lead); Project Administration (lead); Resources (lead); Supervision (lead);  Funding Acquisition (equal); Writing – Review \& Editing (equal); Methodology (equal); Conceptualization (equal);.

\acknowledgments
This research is supported by the U.S. Department of Energy, Office of Science, Biological and Environmental Research Program for the iNAIADS DOE Early Career Award under contract no. DE-AC02-05CH11231. This research also used resources of the National Energy Research Scientific Computing Center (NERSC), a U.S. Department of Energy Office of Science User Facility located at Lawrence Berkeley National Laboratory, operated under Contract No. DE-AC02-05CH11231, and had additional support from the NESAP for Learning program. Additional computational facilities were provided by the Minnesota Supercomputing Institute. This work was also funded by the NSF grant \#193472. VK was supported from the NSF LEAP Science and Technology Center (award \#201962) and the NSF Advancing Deep Learning for Inverse Modeling grant (award \#2313174). This work was supported in part by the U.S. Department of Energy, Office of Science, Office of Workforce Development for Teachers and Scientists (WDTS) under the Office of Science Graduate Student Research Awards (SCGSR) program. We acknowledge the efforts of Valerie Hendrix and Danielle Christianson on developing the BASIN-3D software that supported the data acquisition for this paper. The U.S. Government retains, and the publisher, by accepting the article for publication, acknowledges, that the U.S. Government retains a non-exclusive, paid-up, irrevocable, world-wide license to publish or reproduce the published form of this manuscript, or allow others to do so, for U.S. Government purposes.

Though none of this paper was written by ChatGPT 4.0 (\url{https://www.openai.com/chatgpt}), it was used to provide feedback on written text and to assist in constructing the code to create plots.

\clearpage


%
\bibliography{main}
%




%
%
%
%
%

\renewcommand{\thefigure}{S\arabic{figure}}
\renewcommand{\thetable}{S\arabic{table}}
\setcounter{figure}{0}
\setcounter{table}{0}
\section*{Supporting Information for "Evaluating Deep Learning Approaches for Predictions in Unmonitored Basins with Continental-scale Stream Temperature Models}
\subsection*{S1: Long Short-Term Memory (LSTM) Details}

At each time step, given an input \( x^t \), the Long Short-Term Memory (LSTM) model generates a hidden representation, or embedding, denoted as \( h^t \), which is used for making predictions. The LSTM defines a transition for the hidden state \( h^t \) via an LSTM cell, which takes as input the features \( x^t \) at the current time step and inherited information from previous time steps.

Each LSTM cell contains a cell state \( c^t \), which serves as a memory mechanism, enabling the hidden states \( h^t \) to retain information from previous time steps. The cell state \( c^t \) is updated by combining the previous cell state \( c^{t-1} \), the previous hidden state \( h^{t-1} \), and the current input \( x^t \). This transition of the cell state over time allows the LSTM to capture long-term dependencies. Specifically, the candidate cell state \( \tilde{c}^t \) is generated by applying a \(\text{tanh}(\cdot)\) activation function to the combination of \( x^t \) and \( h^{t-1} \), as follows:

\begin{equation}
\footnotesize
\tilde{c}^t = \text{tanh}(W^c_h h^{t-1} + W^c_x x^t),
\end{equation}
where \( W^c_h\in \mathds{R}^{H\times H} \) and \( W^c_x\in \mathds{R}^{H\times D} \) are weight matrices that define the relationship between the inputs and the candidate cell state. For simplicity, bias terms are omitted, as they can be absorbed into the weight matrices.

Next, we compute three key gates: the forget gate \( f^t \), the input gate \( g^t \), and the output gate \( o^t \), as follows:
\begin{equation}
\footnotesize
\begin{aligned}
f^t &= \sigma(W^f_h h^{t-1} + W^f_x x^t),\\
g^t &= \sigma(W^g_h h^{t-1} + W^g_x x^t),\\
o^t &= \sigma(W^o_h h^{t-1} + W^o_x x^t),
\end{aligned}
\end{equation}
where \( \sigma(\cdot) \) represents the sigmoid activation function.

The new cell state \( c^t \) and hidden state \( h^t \) are then computed as follows:
\begin{equation}
\begin{aligned}
\small
c^t &= f^t \otimes c^{t-1} + g^t \otimes \tilde{c}^t,\\
h^t &= o^t \otimes \text{tanh}(c^t),
\end{aligned}
\end{equation}
where \( \otimes \) represents element-wise multiplication.

Finally, the output of the LSTM at each time step is generated from the hidden state \( h^t \). This output can be passed through a fully connected layer (or any other desired output layer) to make predictions, such as:

\begin{equation}
\hat{y}^t = W_o h^t + b_o,
\end{equation}
where \( W_o \in \mathds{R}^{O \times H} \) is a weight matrix that maps the hidden state \( h^t \) to the output space, and \( b_o \in \mathds{R}^O \) is the bias term for the output. This yields the final predicted value \( \hat{y}^t \).





\section*{Supplementary Figures}
\begin{figure}[h]
    \centering
    \includegraphics[width=0.8\textwidth]{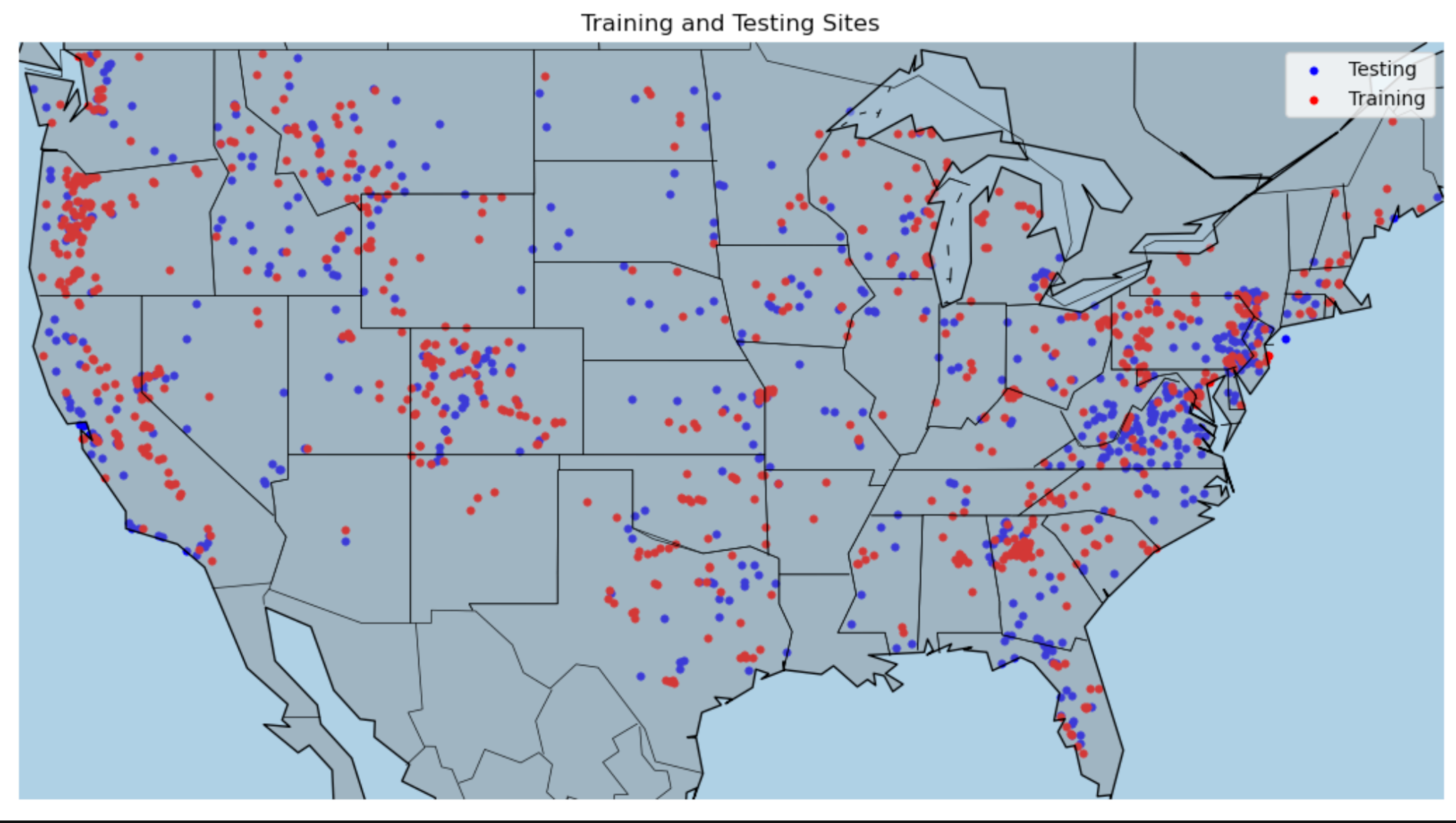}
    \caption{Spatial distribution of default training (n=782) and testing (n=580) stream sites}
    \label{fig:train_test_map}
\end{figure}

 \subsubsection*{Permutation Feature Importance Additional Results (Figures S2-S14)}
\begin{figure}
    \centering
    \includegraphics[width=0.95\textwidth]{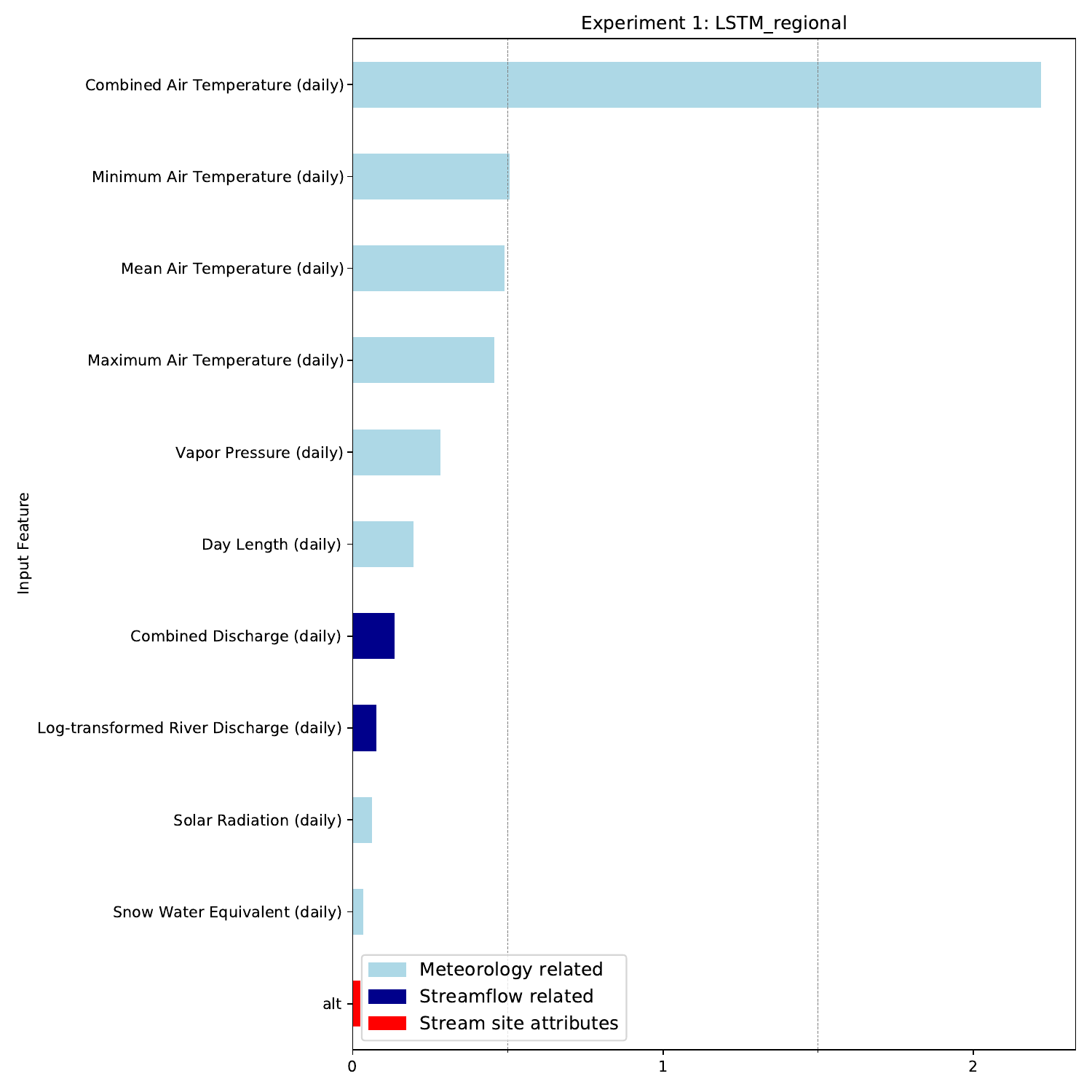}
    \caption{Permutation feature importance for various input features in the \textit{LSTM\_Regional} model from Experiment 1 measured by how much the overall RMSE increases compared to the baseline RMSE of 2.54$^{\circ}$C. Only importances greater than 0.09$^{\circ}$C are shown based on the standard deviation of the RMSE per individual member of the model ensemble.}
    \label{fig:pfi_regional_conus}
\end{figure}

\begin{figure}
    \centering
    \includegraphics[width=0.95\textwidth]{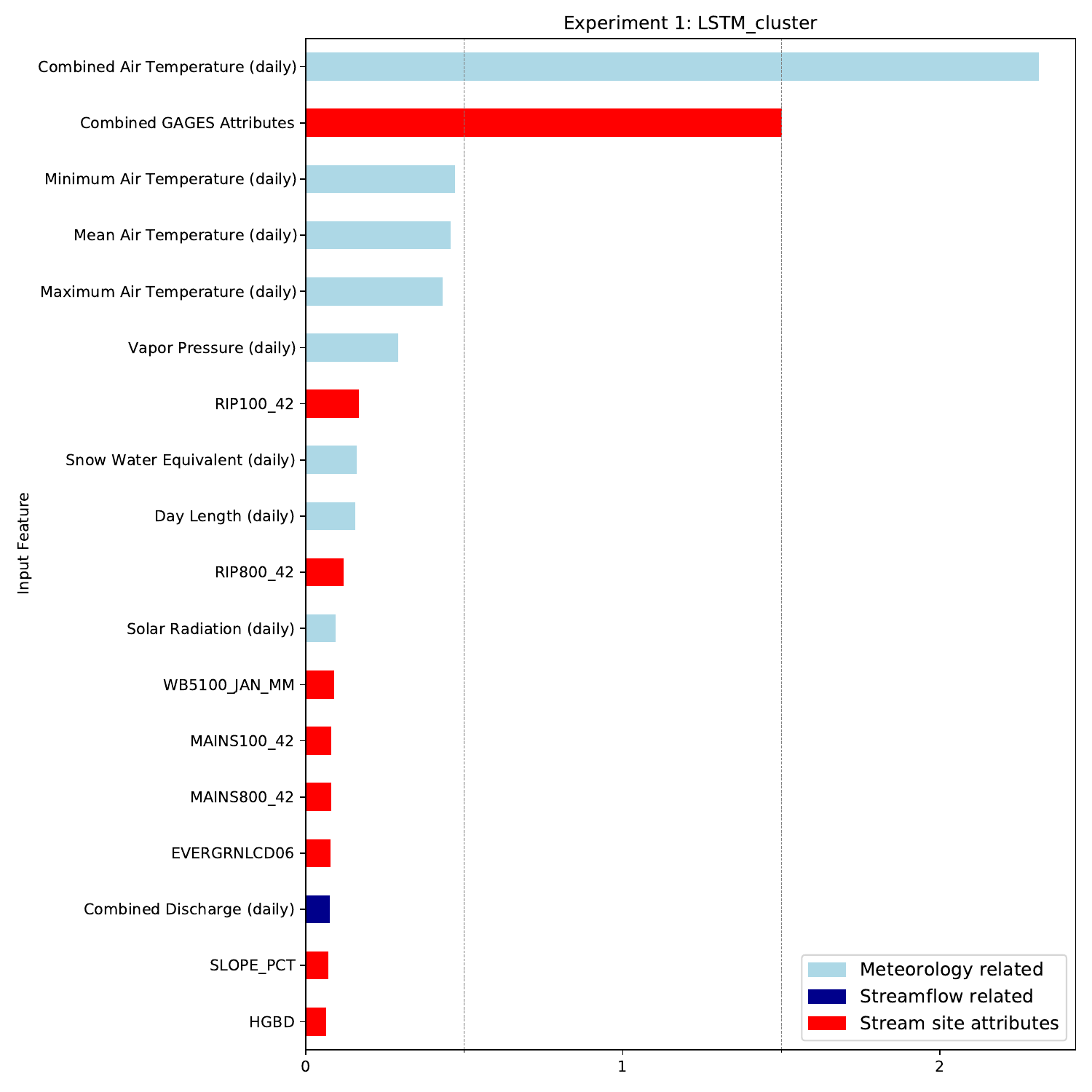}
    \caption{Permutation feature importance for various input features in the \textit{LSTM\_Cluster} model from Experiment 1 measured by how much the overall RMSE increases compared to the baseline RMSE of 2.37$^{\circ}$C. Only importances greater than 0.06$^{\circ}$C are shown based on the standard deviation of the RMSE per individual member of the model ensemble.}
    \label{fig:pfi_cluster_conus}
\end{figure}

\begin{figure}
    \centering
    \includegraphics[width=0.95\textwidth]{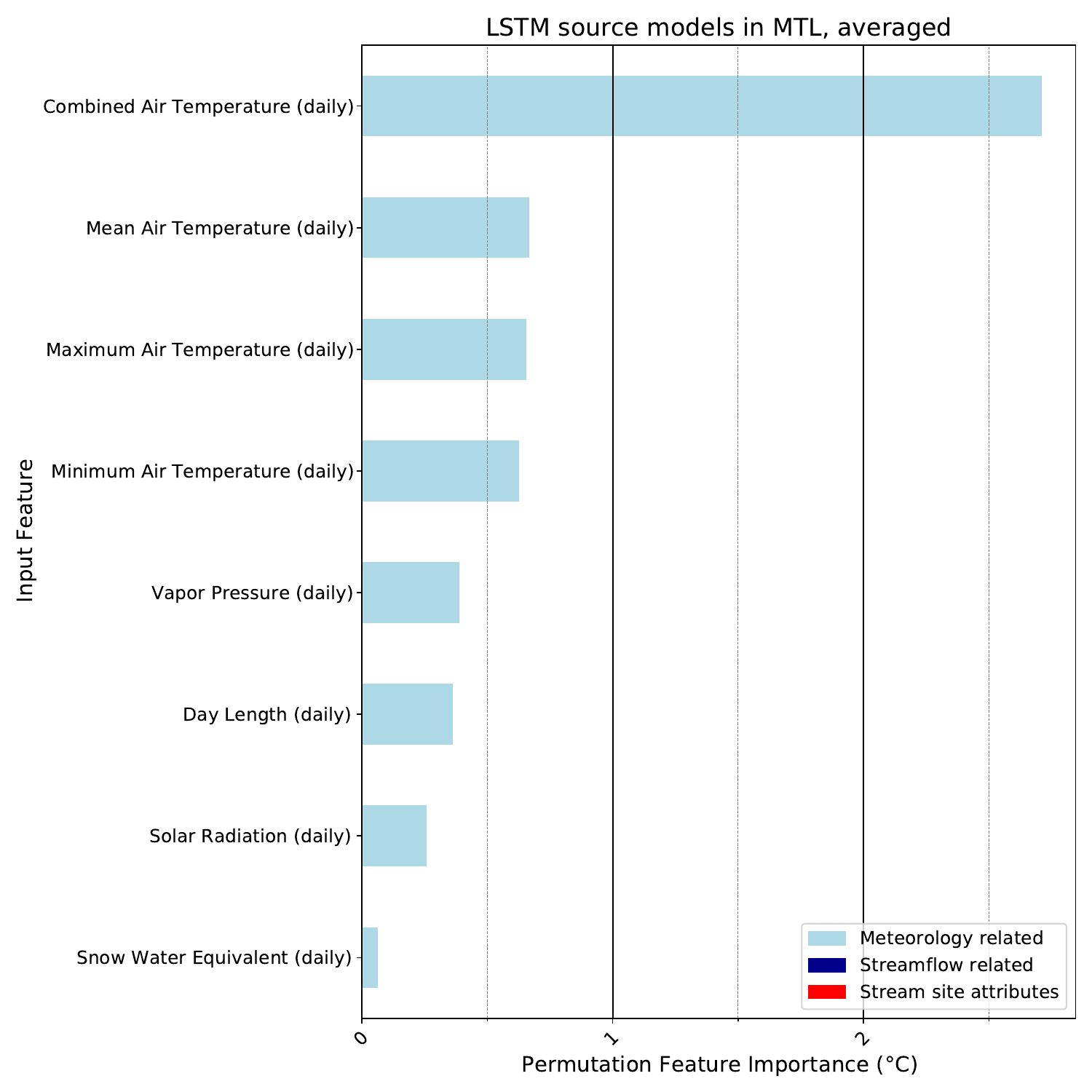}
    \caption{Permutation feature importance for various input features in the \textit{MTL} selected source models from Experiment 1 measured by how much the overall RMSE increases compared to the baseline RMSE of 2.37$^{\circ}$C. All importances greater than zero are shown.}
    \label{fig:pfi_mtl}
\end{figure}

\begin{figure}
    \centering
    \includegraphics[width=0.95\textwidth]{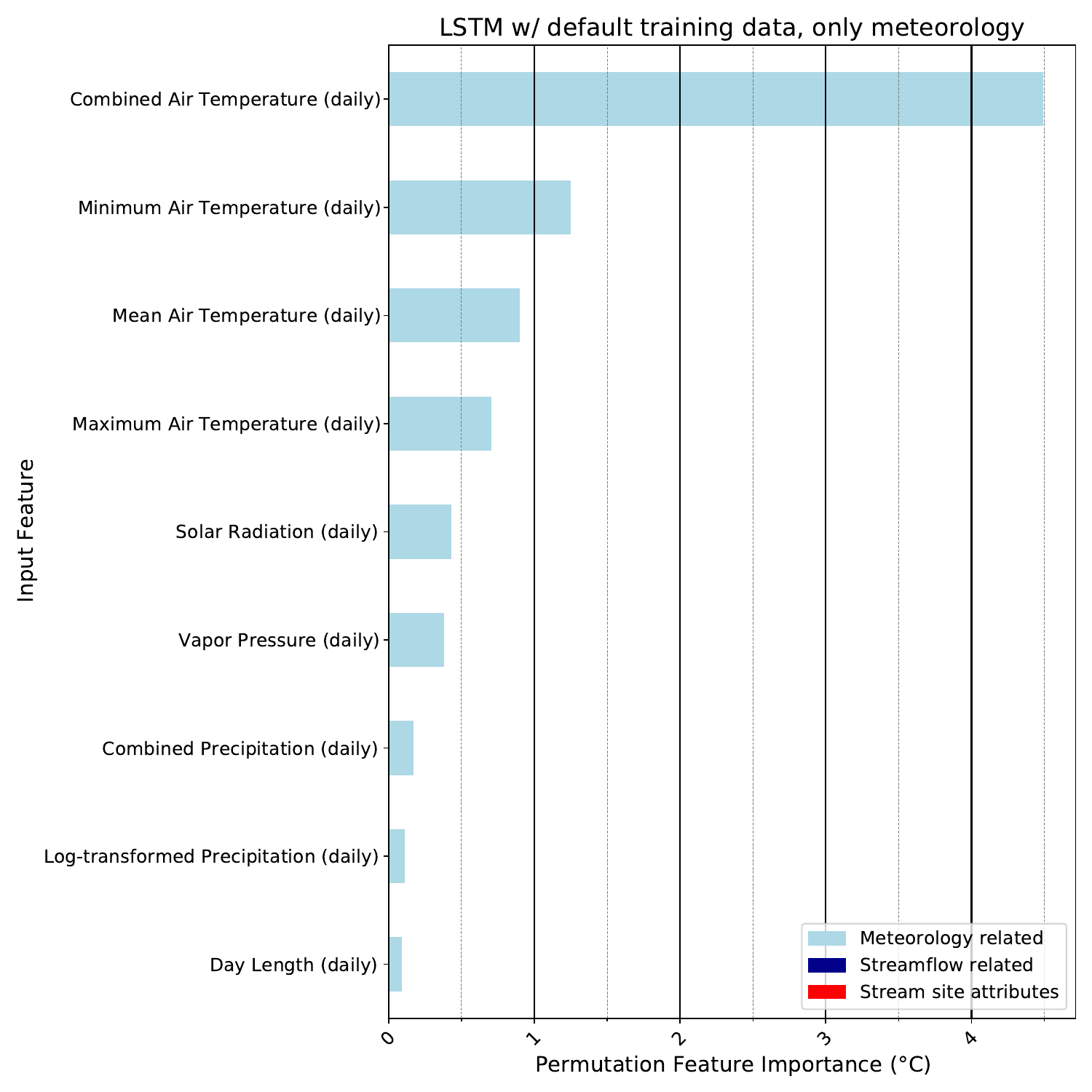}
    \caption{Permutation feature importance for various input features in the model using the default training data and only meteorological inputs from Experiment 2. Only importances greater than 0.05$^{\circ}$C are shown based on the standard deviation of the RMSE per individual member of the model ensemble.}
    \label{fig:pfi_defTrain_meteo_only}
\end{figure}

\begin{figure}
    \centering
    \includegraphics[width=0.95\textwidth]{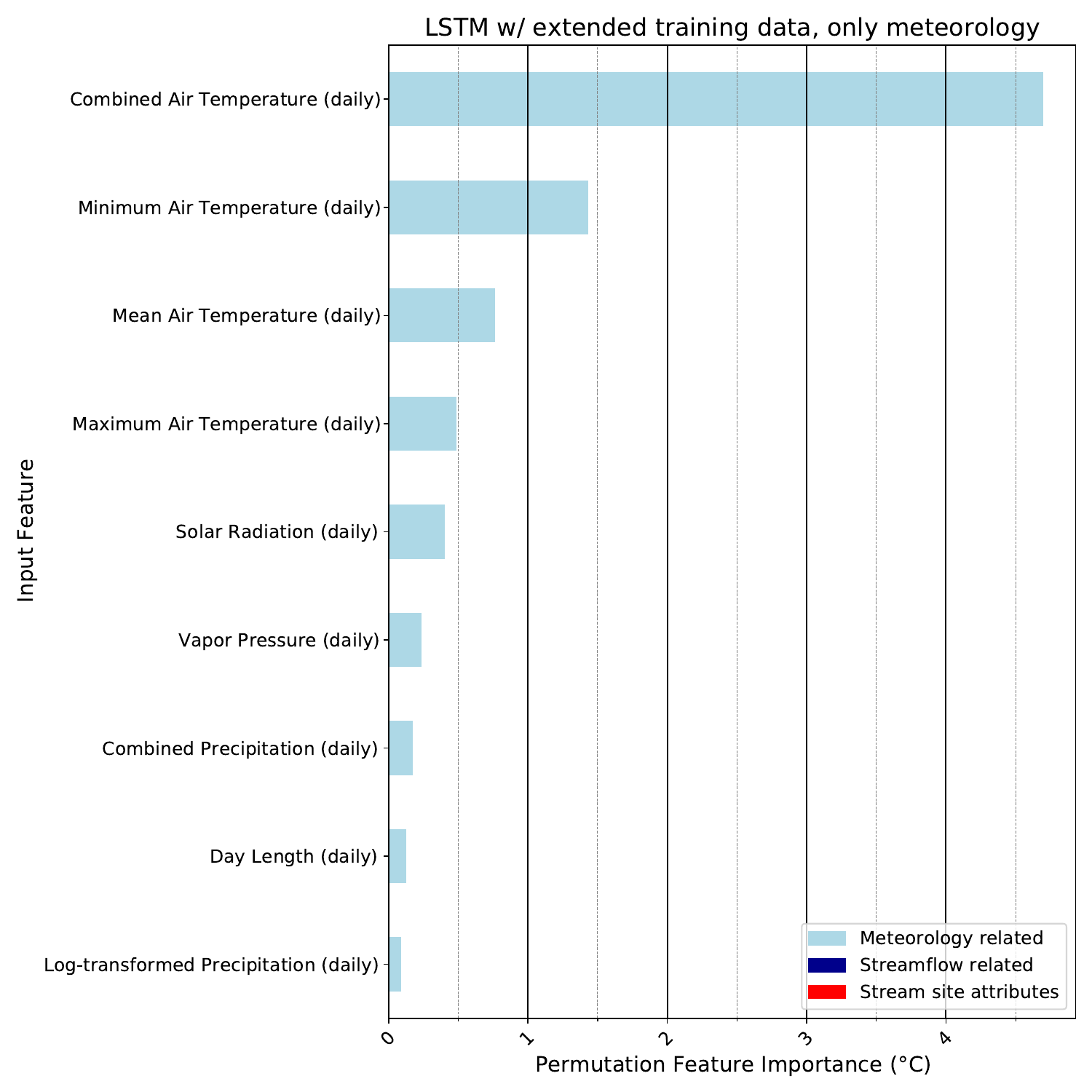}
    \caption{Permutation feature importance for various input features in the model using extended training data and only meteorological inputs from Experiment 2. Only importances greater than 0.04$^{\circ}$C are shown based on the standard deviation of the RMSE per individual member of the model ensemble.}
    \label{fig:pfi_extTrain_meteo_only}
\end{figure}

\begin{figure}
    \centering
    \includegraphics[width=0.95\textwidth]{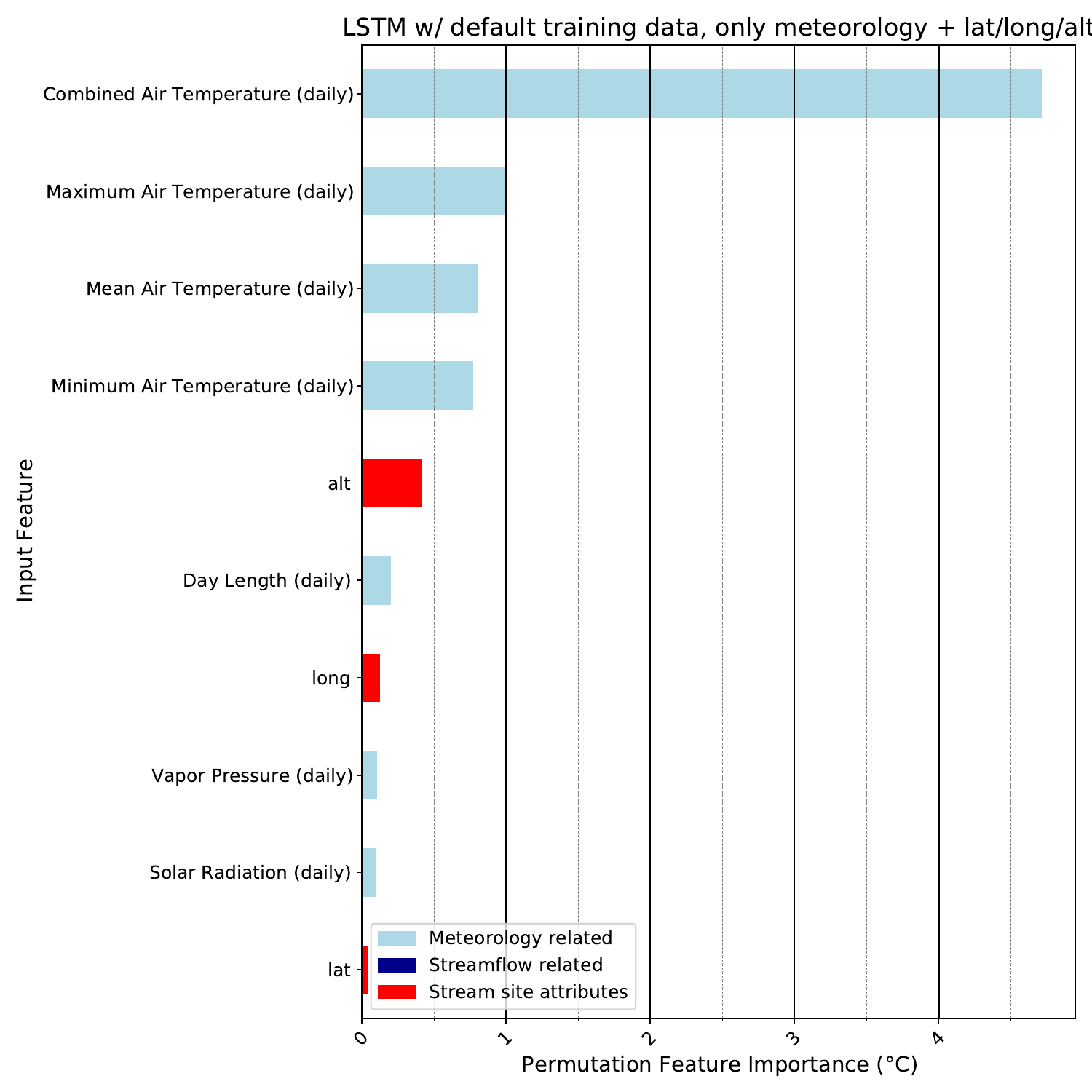}
    \caption{Permutation feature importance for various input features in the model using the default training data and only meteorological inputs and latitude/longitude/elevation from Experiment 2. Only importances greater than 0.01$^{\circ}$C are shown based on the standard deviation of the RMSE per individual member of the model ensemble.}
    \label{fig:pfi_defTrain_meteo_latlongalt}
\end{figure}

\begin{figure}
    \centering
    \includegraphics[width=0.95\textwidth]{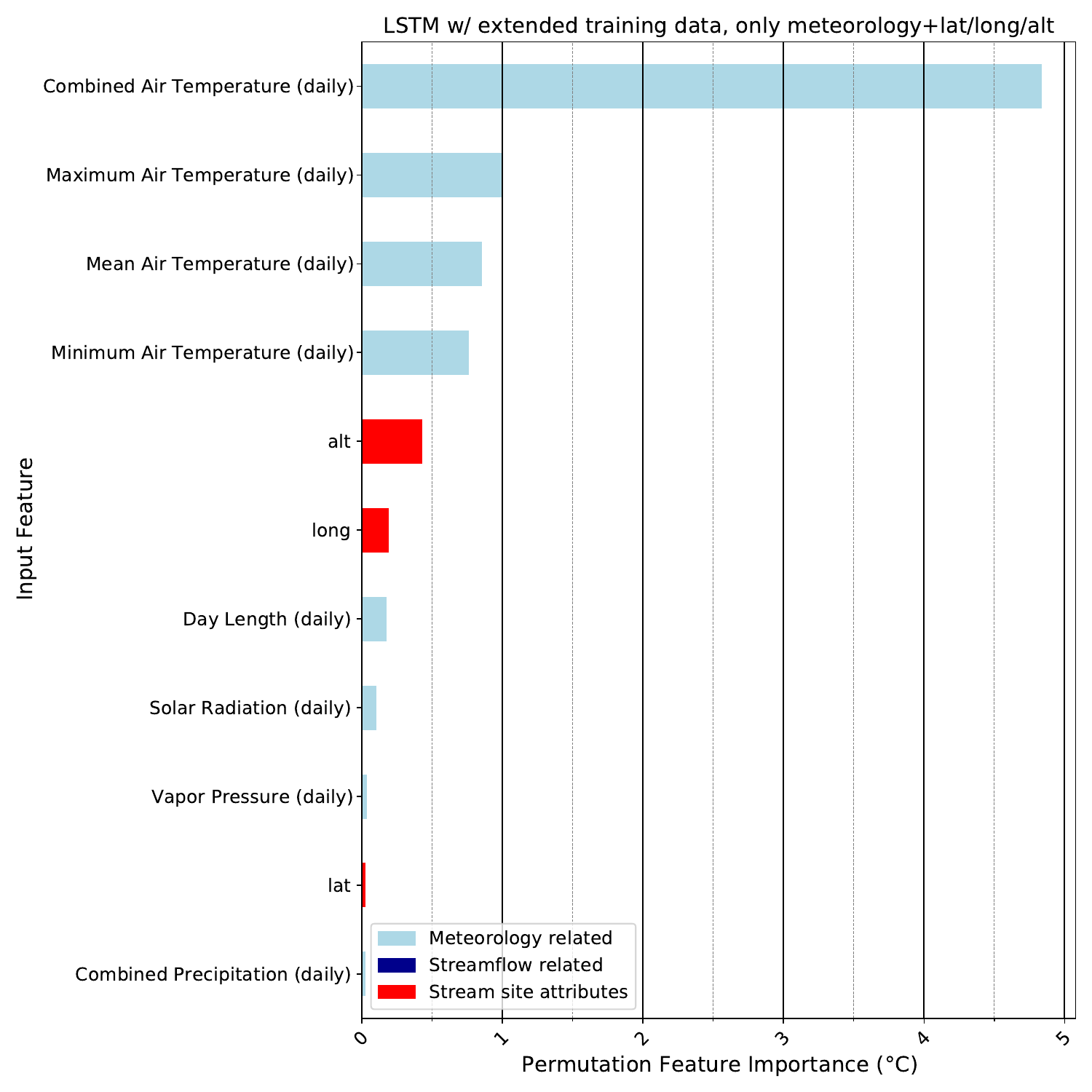}
    \caption{Permutation feature importance for various input features in the model using  extended training data and only meteorological inputs and latitude/longitude/elevation from Experiment 2. Only importances greater than 0.02$^{\circ}$C are shown based on the standard deviation of the RMSE per individual member of the model ensemble.}
    \label{fig:pfi_extTrain_meteo_latlongalt}
\end{figure}

\begin{figure}
    \centering
    \includegraphics[width=0.95\textwidth]{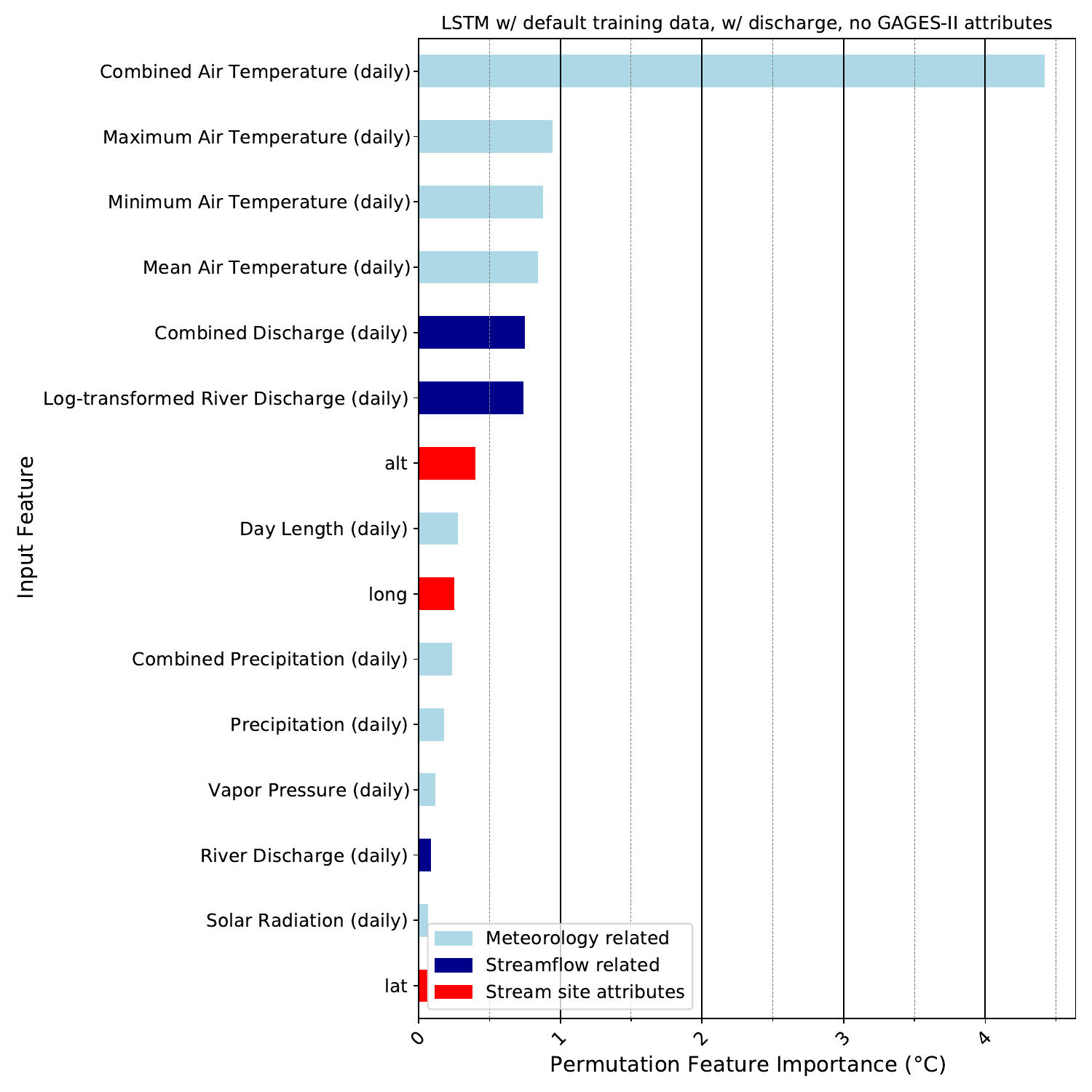}
    \caption{Permutation feature importance for various input features in the model using the default training data and with meteorological and streamflow drivers from Experiment 2. Only importances greater than 0.04$^{\circ}$C are shown based on the standard deviation of the RMSE per individual member of the model ensemble.}
    \label{fig:pfi_defTrain_meteo_wRDC_noGAGE}
\end{figure}

\begin{figure}
    \centering
    \includegraphics[width=0.95\textwidth]{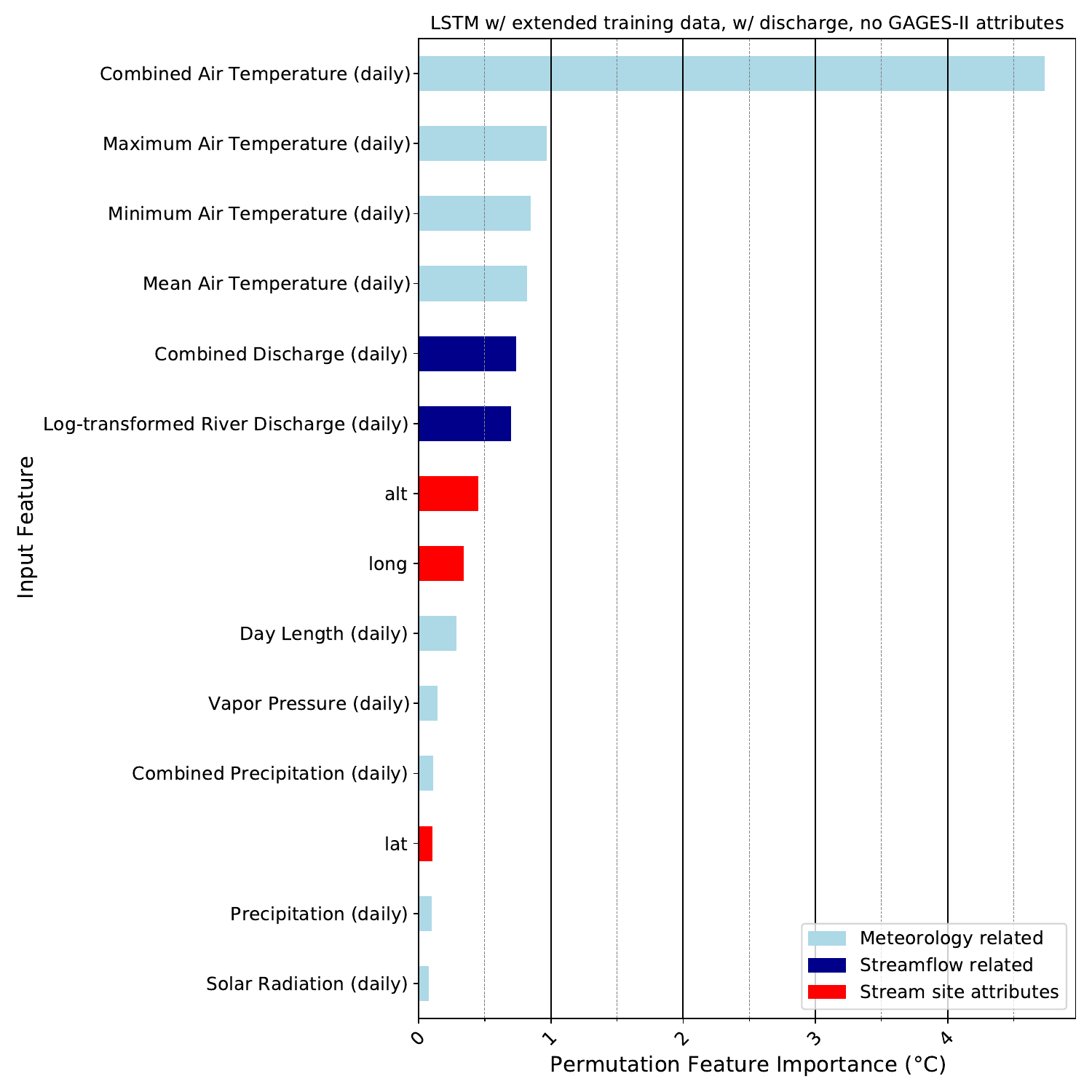}
    \caption{Permutation feature importance for various input features in the model using the extended training data and with meteorological and streamflow drivers from Experiment 2. Only importances greater than 0.06$^{\circ}$C are shown based on the standard deviation of the RMSE per individual member of the model ensemble.}
    \label{fig:pfi_extTrain_meteo_wRDC_noGAGE}
\end{figure}

\begin{figure}
    \centering
    \includegraphics[width=0.95\textwidth]{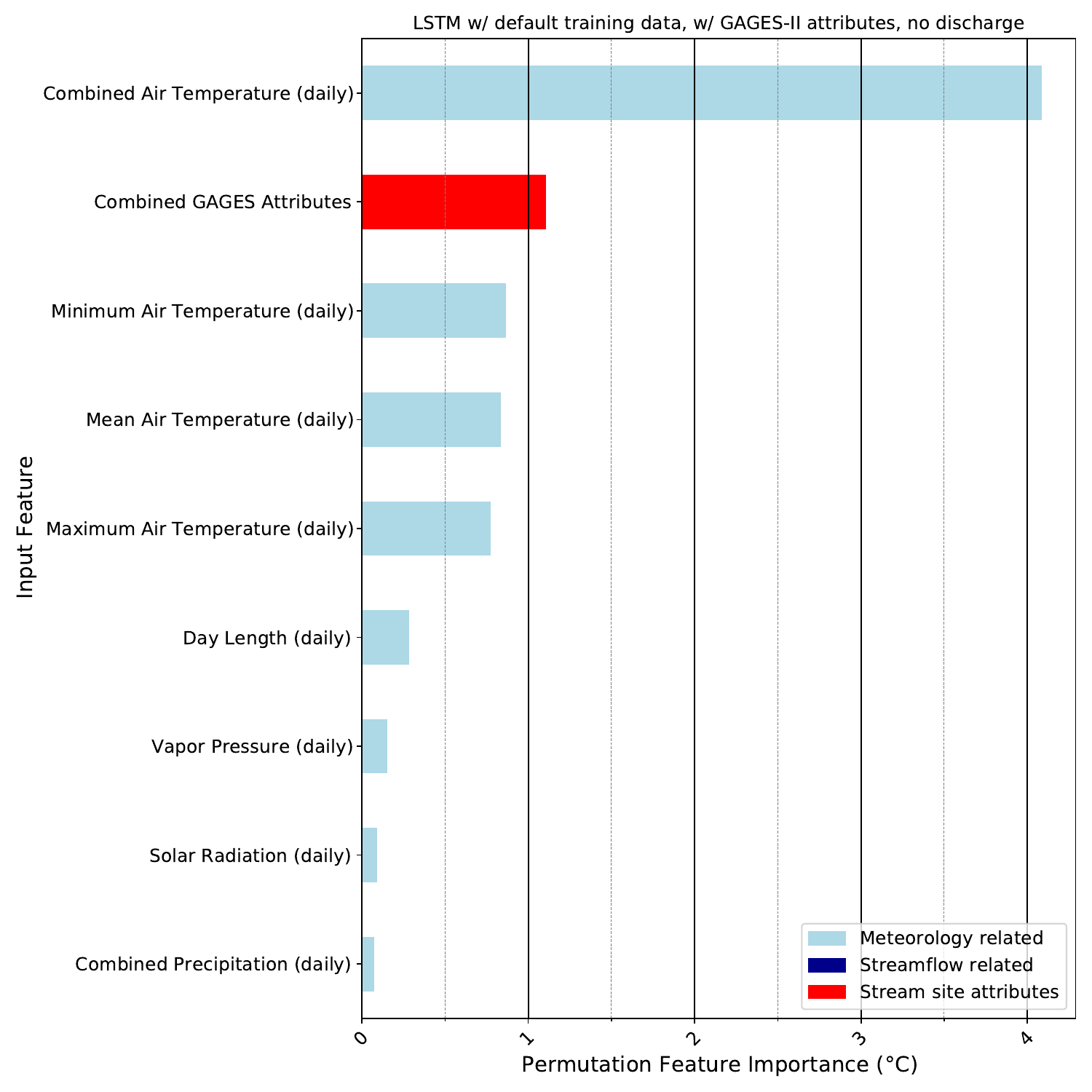}
    \caption{Permutation feature importance for various input features in the model using the default training data and with meteorological and  streamflow drivers as well as GAGES-II attributes from Experiment 2. Only importances greater than 0.01$^{\circ}$C are shown based on the standard deviation of the RMSE per individual member of the model ensemble.}
    \label{fig:pfi_defTrain_meteo_noRDC_wGAGE}
\end{figure}

\begin{figure}
    \centering
    \includegraphics[width=0.95\textwidth]{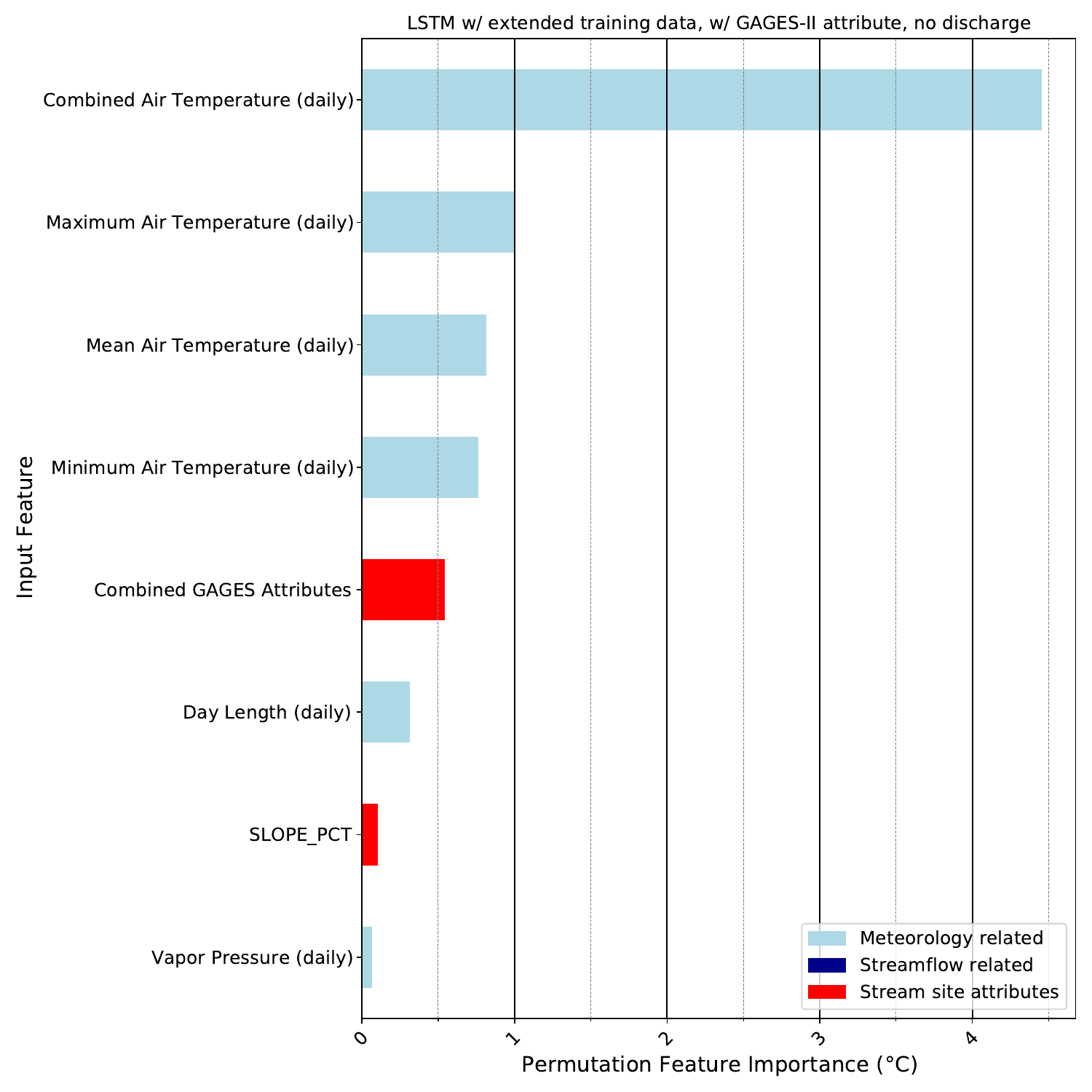}
    \caption{Permutation feature importance for various input features in the model using the extended training data and with meteorological and  streamflow drivers as well as GAGES-II attributes from Experiment 2. Only importances greater than 0.01$^{\circ}$C are shown based on the standard deviation of the RMSE per individual member of the model ensemble.}
    \label{fig:pfi_extTrain_meteo_noRDC_wGAGE}
\end{figure}

\begin{figure}
    \centering
    \includegraphics[width=0.95\textwidth]{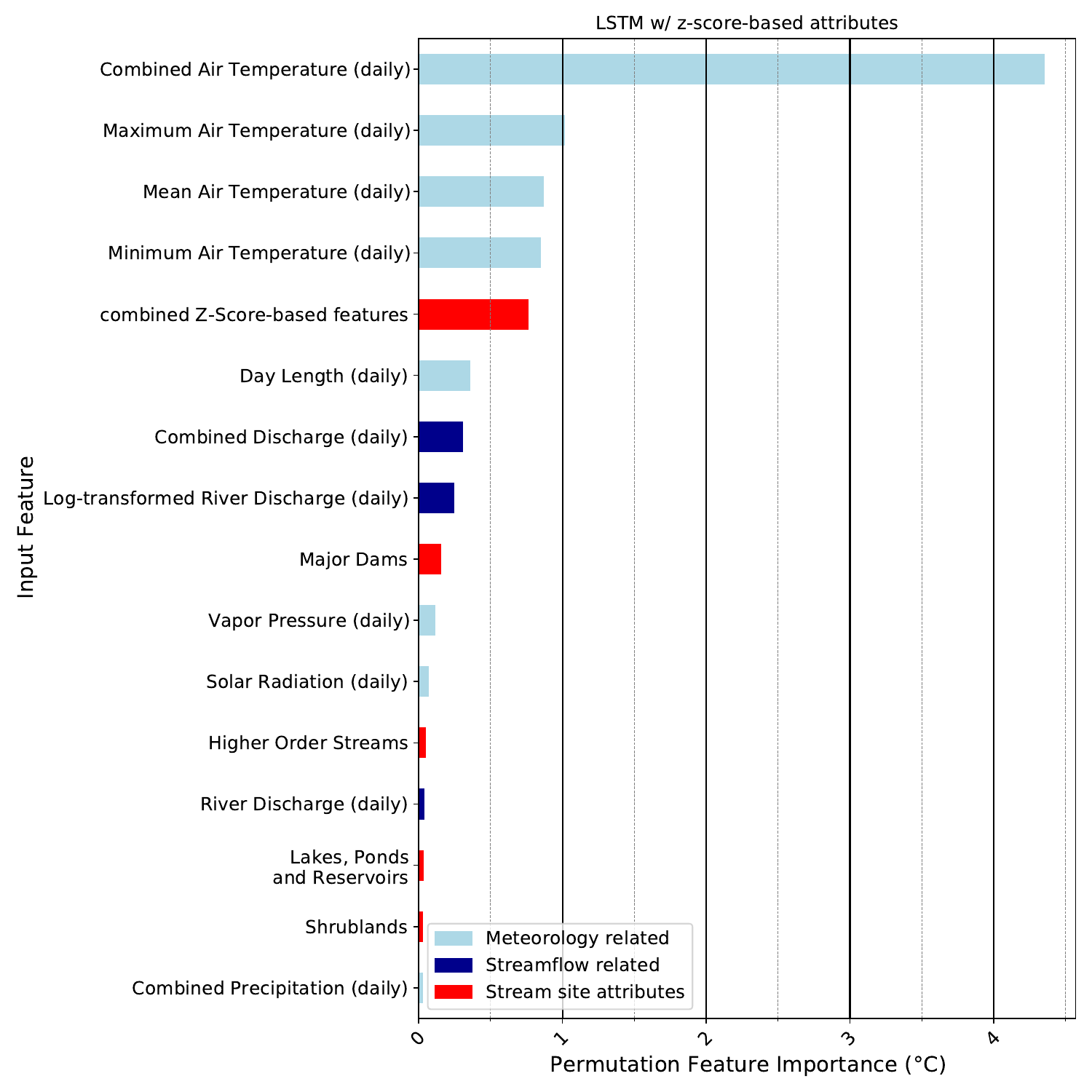}
    \caption{Permutation feature importance for various input features in the model using the z-score based attributes in addition to meteorological and streamflow drivers from Experiment 3 measured by how much the overall RMSE increases compared to the baseline RMSE of 1.93$^{\circ}$C. Only importances greater than 0.03$^{\circ}$C are shown based on the standard deviation of the RMSE per individual member of the model ensemble.}
    \label{fig:pfi_exp3_zscore}
\end{figure}

\begin{figure}
    \centering
    \includegraphics[width=0.95\textwidth]{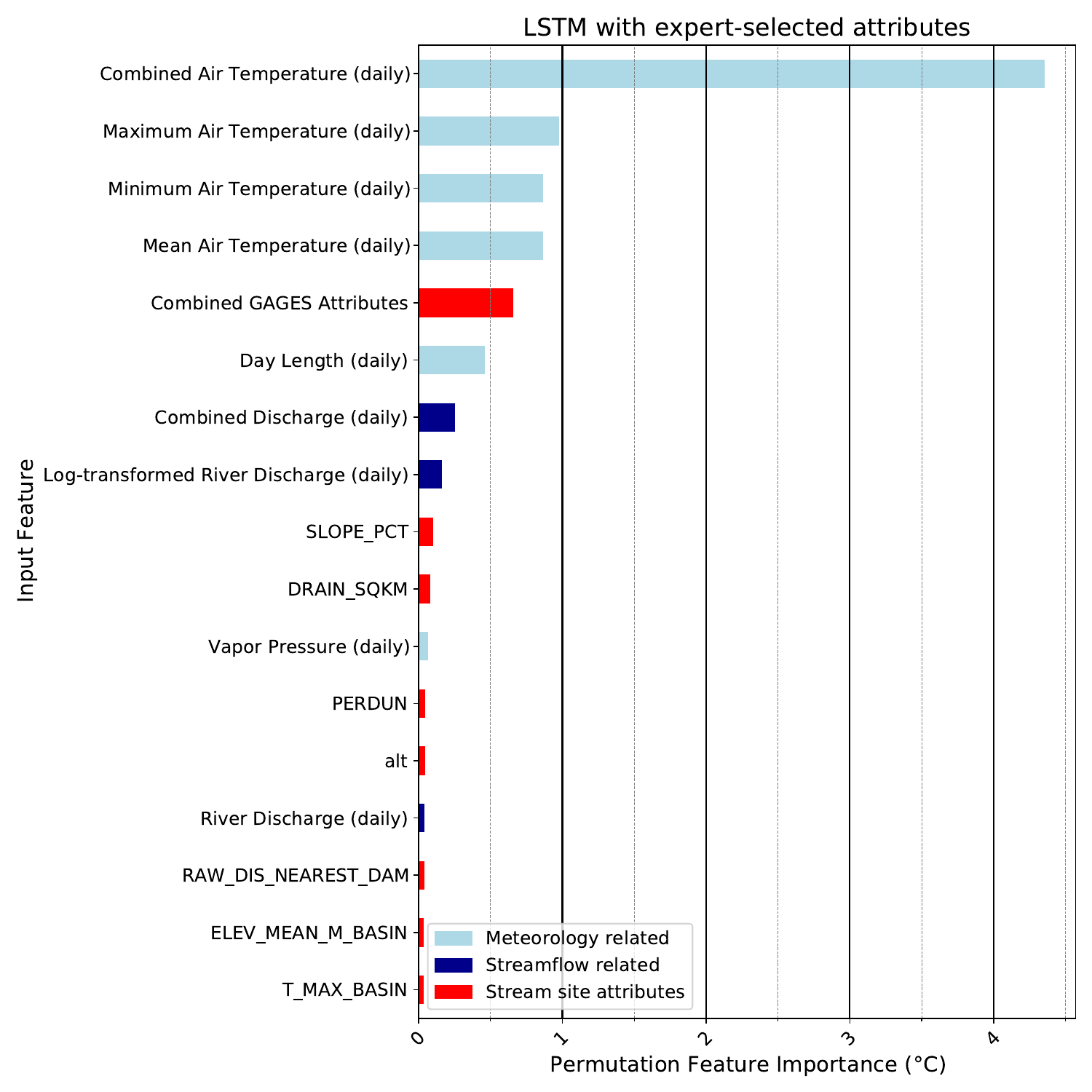}
    \caption{Permutation feature importance for various input features in the model using 21 expert-selected GAGES-II attributes in addition to meteorological and streamflow drivers from Experiment 3. Only importances greater than 0.01$^{\circ}$C are shown based on the standard deviation of the RMSE per individual member of the model ensemble. GAGES II variable descriptions can be found at \url{https://doi.org/10.5066/P96CPHOT} in \textit{variable\_descriptions.txt}}
    \label{fig:pfi_exp3_expert}
\end{figure}

\begin{figure}
    \centering
    \includegraphics[width=0.95\textwidth]{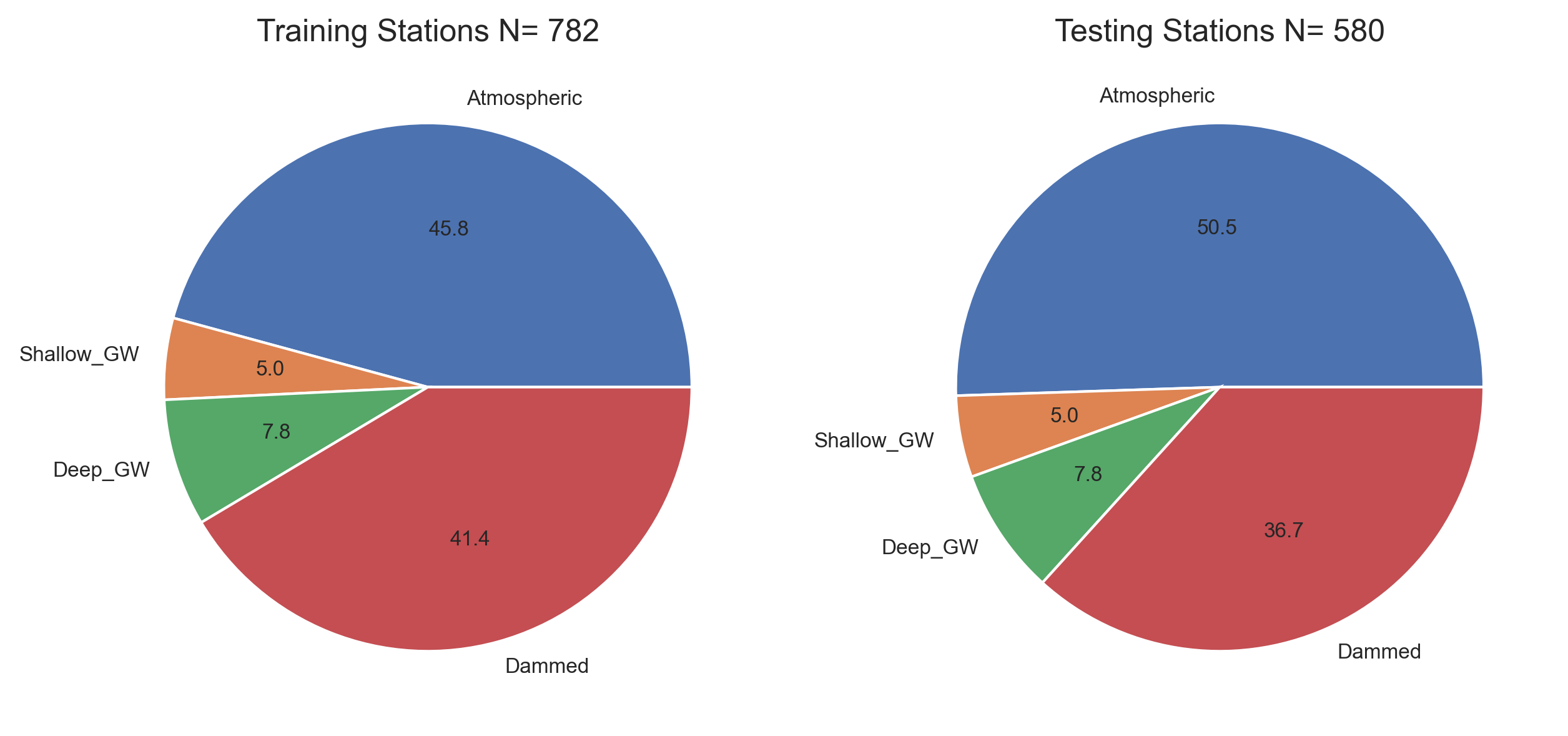}
    \caption{Paired air/stream temperature analysis classes for training stations (left) and testing stations (right).}
    \label{fig:airstream_traintest}
\end{figure}

 \subsubsection*{Comparison of Experiments with baseline \textit{LSTM\_conus} model (Figures S16-S19)}
\begin{figure}
    \centering
    \includegraphics[width=0.7\textwidth]{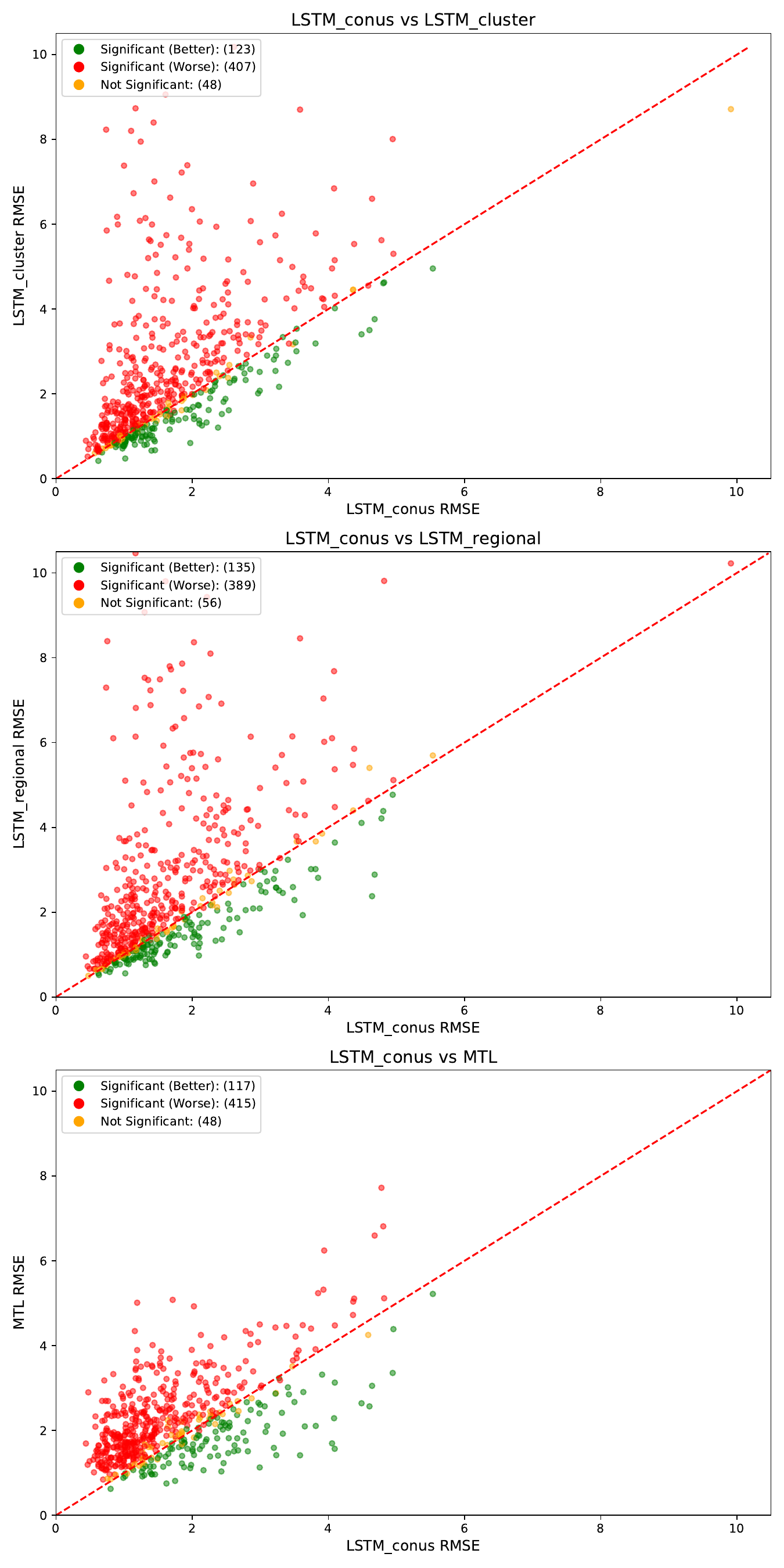}
    \caption{Scatter plots showing the individual RMSE values for the 580 test sites for each model in Experiment 1 compared with the top-down \textit{LSTM\_conus} model. Sites are colored according the result of the Wilcoxon significance test described in Section 3.3. Not pictured in third panel is huc\_id 09416000 which has an \textit{MTL} RMSE of 13.59°C and \textit{LSTM\_conus} RMSE of 9.91°C.}
    \label{fig:scatter_exp1}
\end{figure}

\begin{figure}
    \centering
    \includegraphics[width=.9\textwidth]{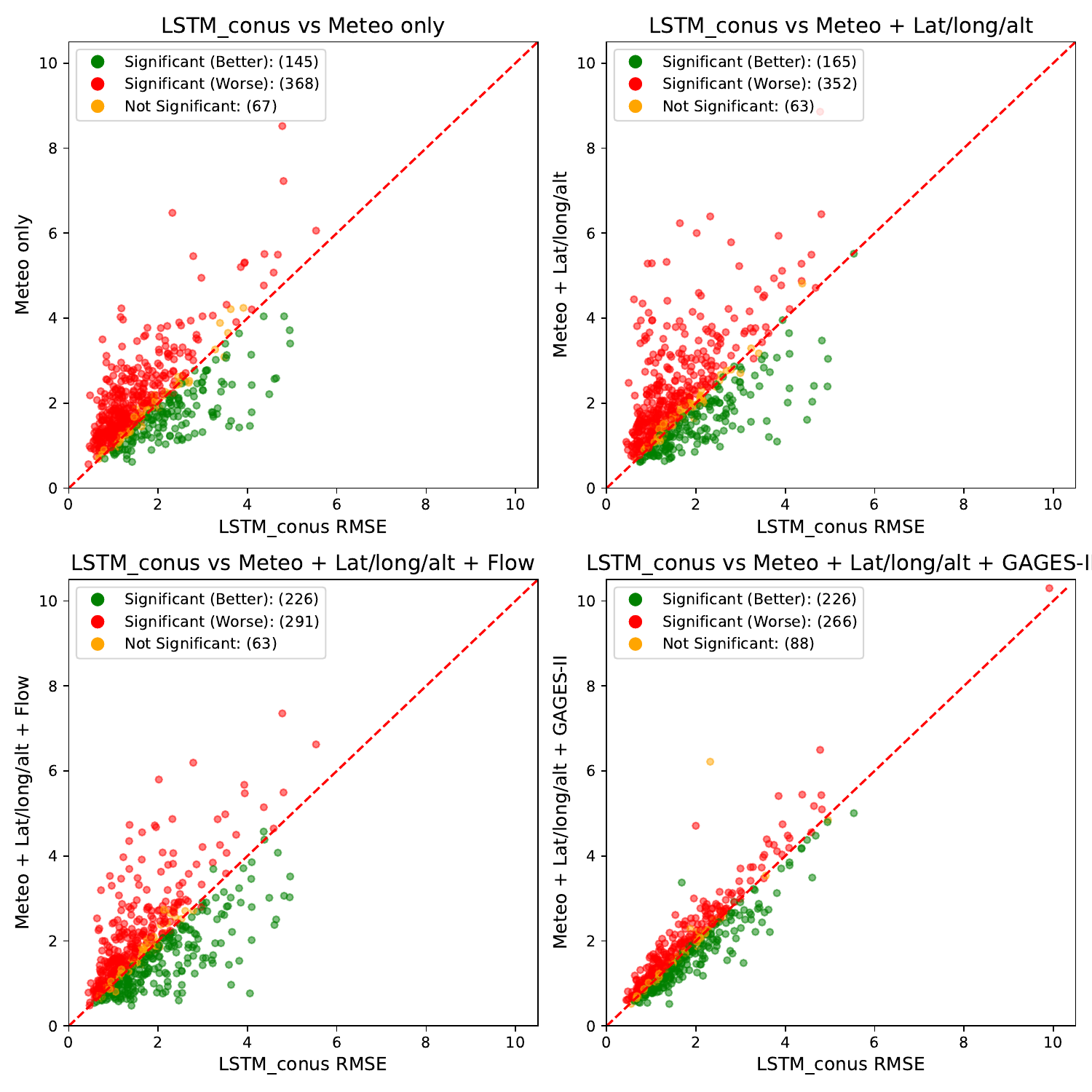}
    \caption{Scatter plots showing the individual RMSE values for the 580 test sites for each model in Experiment 2 that uses the default training data compared with the top-down \textit{LSTM\_conus} model. Sites are colored according the result of the Wilcoxon significance test described in Section 3.3. Not pictured in all the plots except the bottom right panel is huc\_id 09416000 which has \textit{LSTM\_conus} RMSE of 9.91°C, a \textit{Meteo only} RMSE of 11.67°C, a \textit{Meteo+Lat/Long/Alt} RMSE of 12.48°C, a \textit{Meteo+Lat/Long/Alt+Flow} RMSE of 11.56°C.}
    \label{fig:scatter_exp2a}
\end{figure}

\begin{figure}
    \centering
    \includegraphics[width=.9\textwidth]{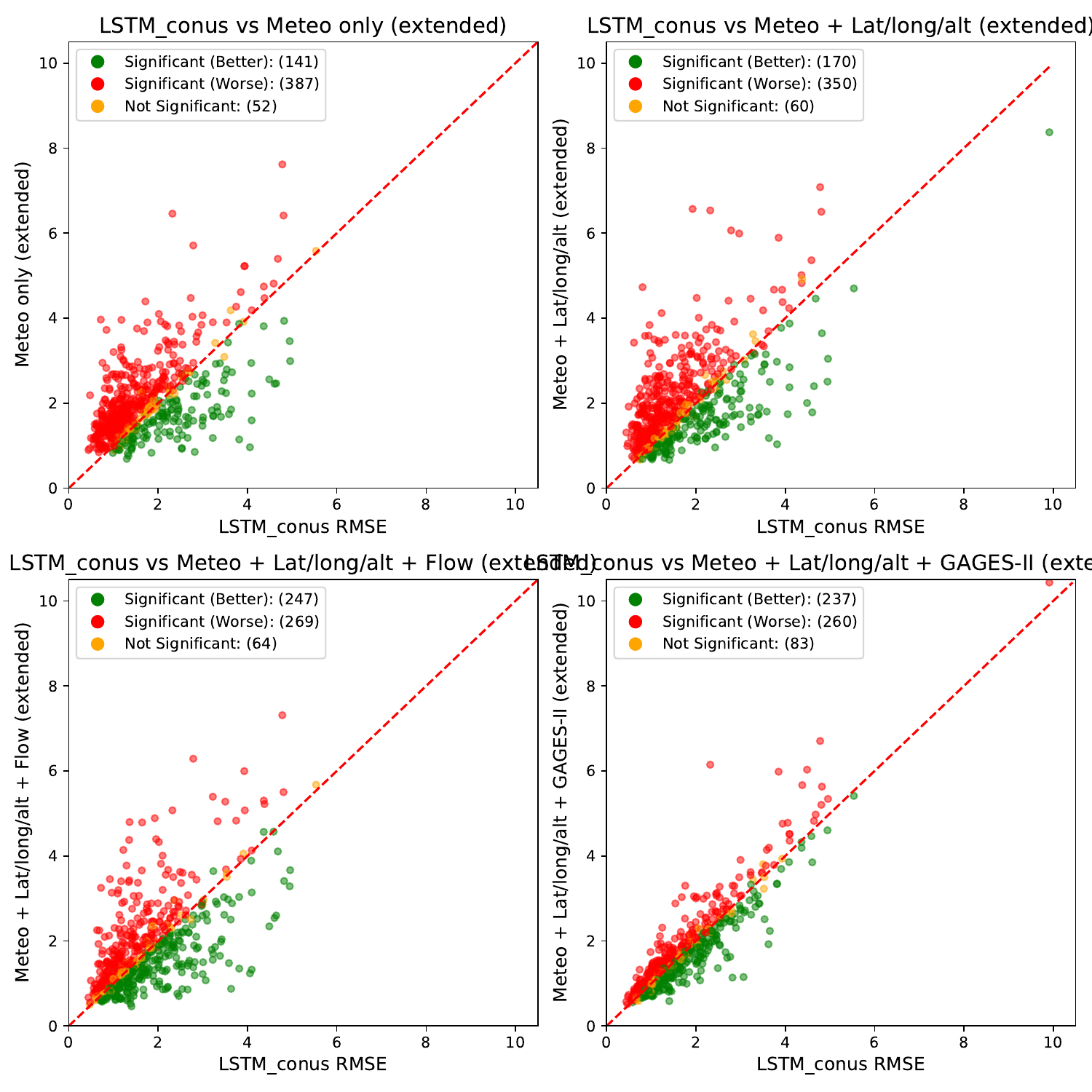}
    \caption{Scatter plots showing the individual RMSE values for the 580 test sites for each model in Experiment 2 that uses the extended training data compared with the top-down \textit{LSTM\_conus} model. Sites are colored according the result of the Wilcoxon significance test described in Section 3.3. Not pictured in left two panels is huc\_id 09416000 which has \textit{LSTM\_conus} RMSE of 9.91°C, a \textit{Meteo only} RMSE of 11.83°C and a \textit{Meteo+Lat/Long/Alt+Flow} RMSE of 11.76°C}
    \label{fig:scatter_exp2b}
\end{figure}

\begin{figure}
    \centering
    \includegraphics[width=0.7\textwidth]{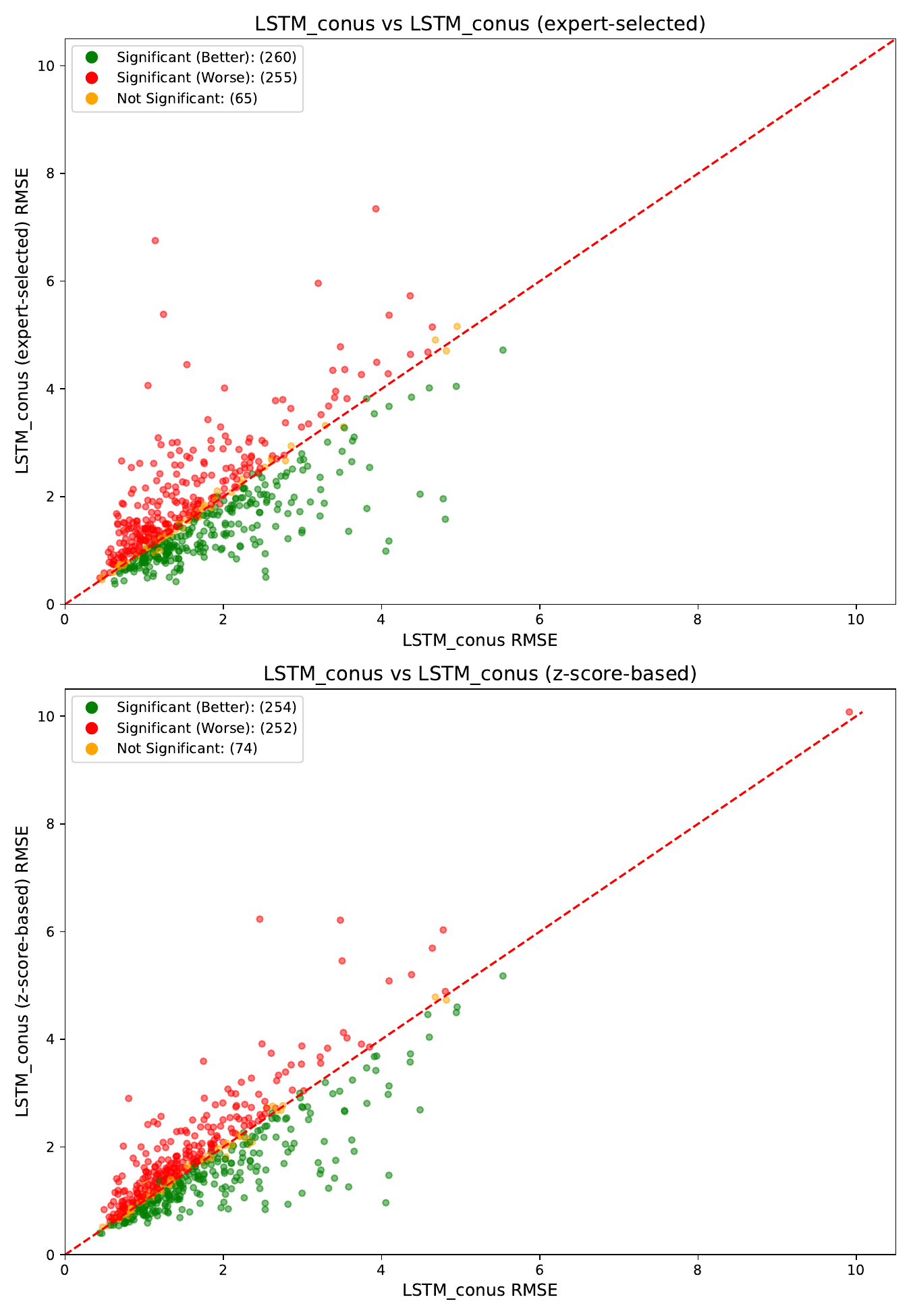}
    \caption{Scatter plots showing the individual RMSE values for the 580 test sites for each model in Experiment 3 compared with the top-down \textit{LSTM\_conus} model. Sites are colored according the result of the Wilcoxon significance test described in Section 3.3. Not pictured in the first panel is huc\_id 09416000 which has an \textit{LSTM\_conus (expert-selected)} RMSE of 11.41°C}
    \label{fig:scatter_exp3}
\end{figure}

\begin{figure}
    \centering
    \includegraphics[width=0.7\textwidth]{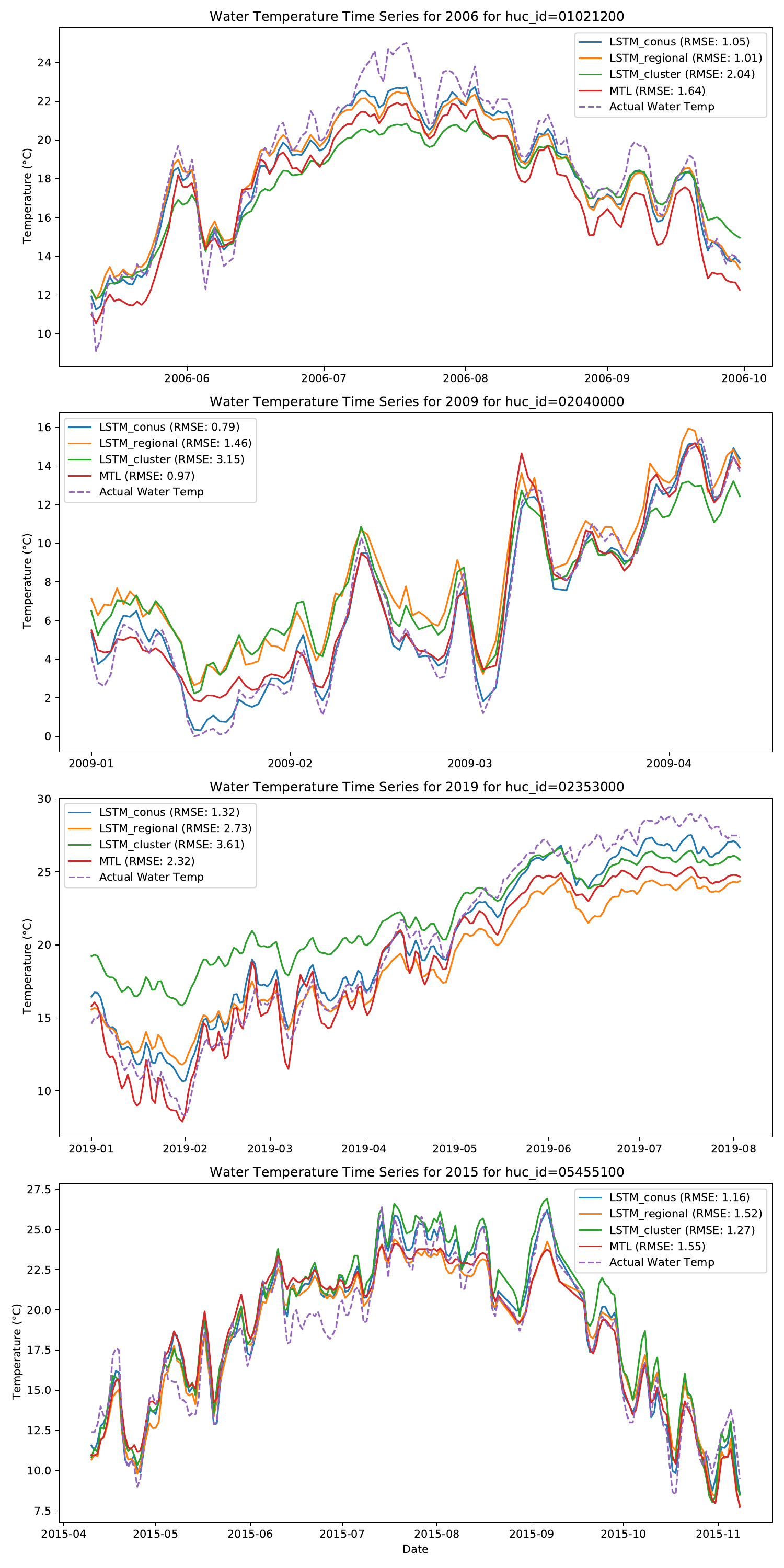}
    \caption{Time series stream temperature predictions for the 4 models used in Experiment 1 across four different sites. Each plot displays the most recent year of that site's data.}
    \label{fig:exp1_time_series}
\end{figure}

\begin{figure}
    \centering
    \includegraphics[width=0.7\textwidth]{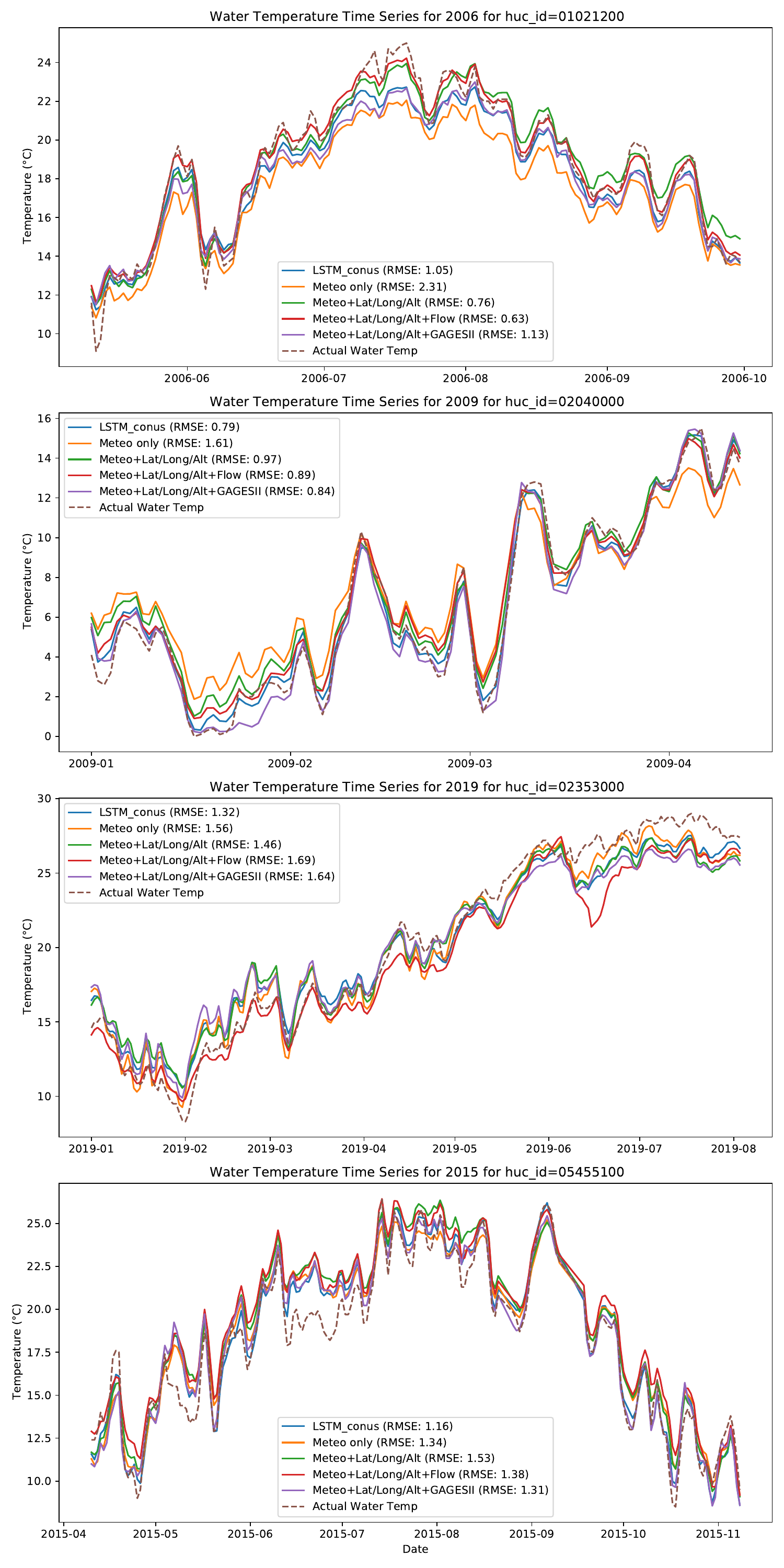}
    \caption{Time series stream temperature predictions for the 3 models used in Experiment 2 that use the default training data across four different sites compared with LSTM\_conus. Each plot displays the most recent year of that site's data.}
    \label{fig:exp2a_time_series}
\end{figure}

\begin{figure}
    \centering
    \includegraphics[width=0.7\textwidth]{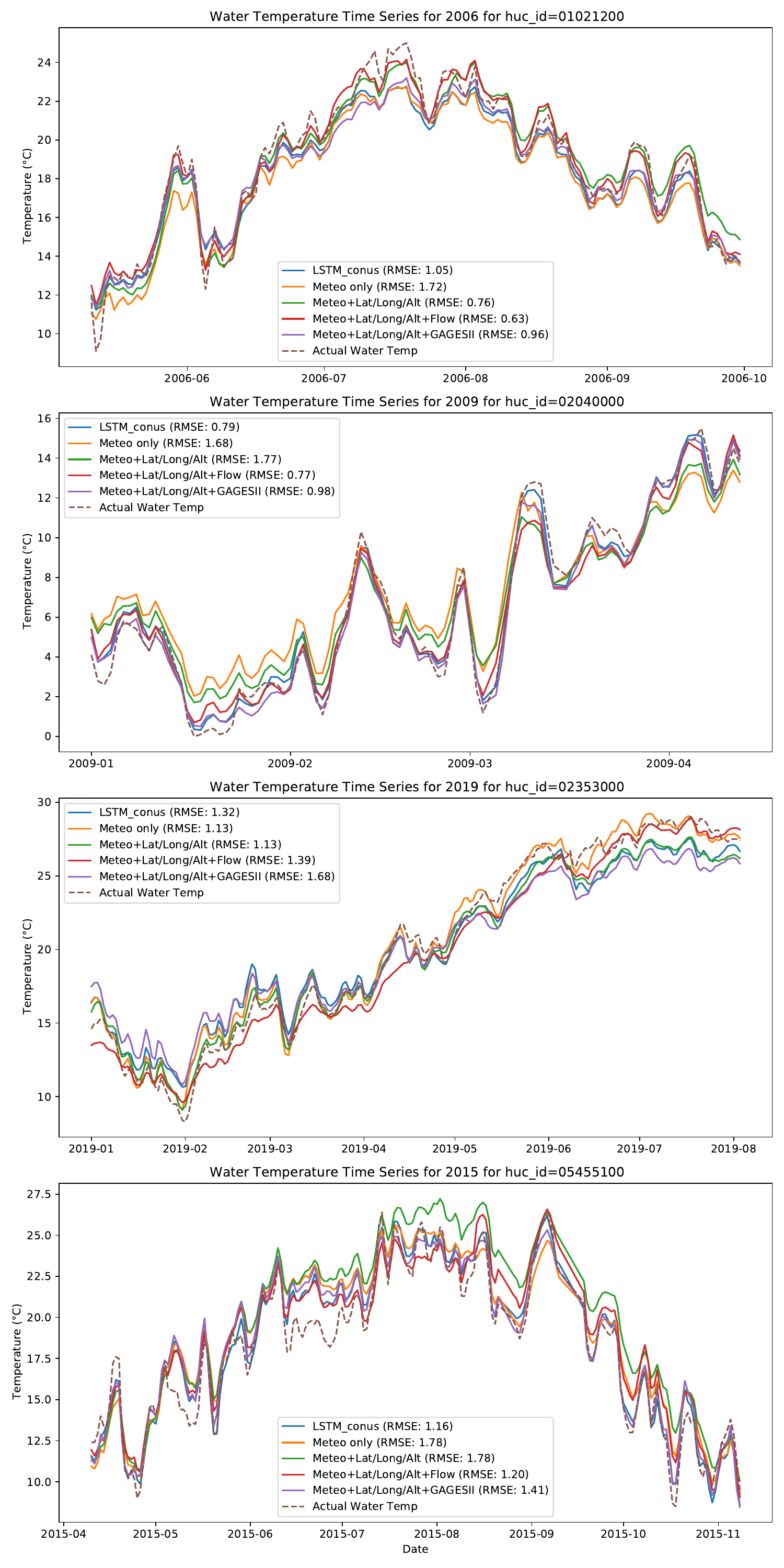}
    \caption{Time series stream temperature predictions for the 3 models used in Experiment 2 that use the extended training data across four different sites compared with LSTM\_conus. Each plot displays the most recent year of that site's data.}
    \label{fig:exp2b_time_series}
\end{figure}

\begin{figure}
    \centering
    \includegraphics[width=0.7\textwidth]{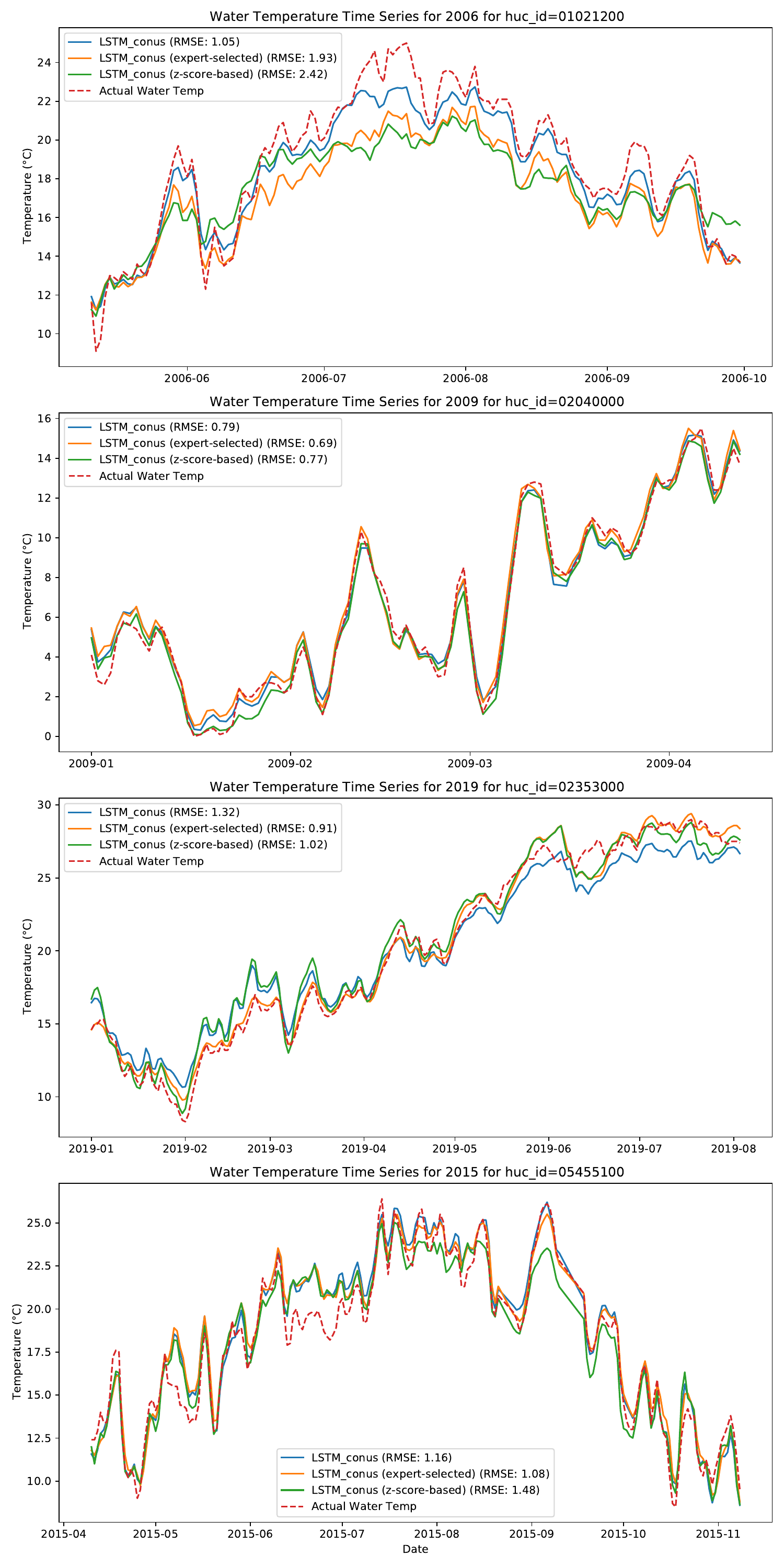}
    \caption{Time series stream temperature predictions for the 2 additional models used in Experiment 3 across four different sites compared with LSTM\_conus. Each plot displays the most recent year of that site's data.}
    \label{fig:exp3_time_series}
\end{figure}

\begin{figure}
    \centering
    \includegraphics[width=0.95\textwidth]{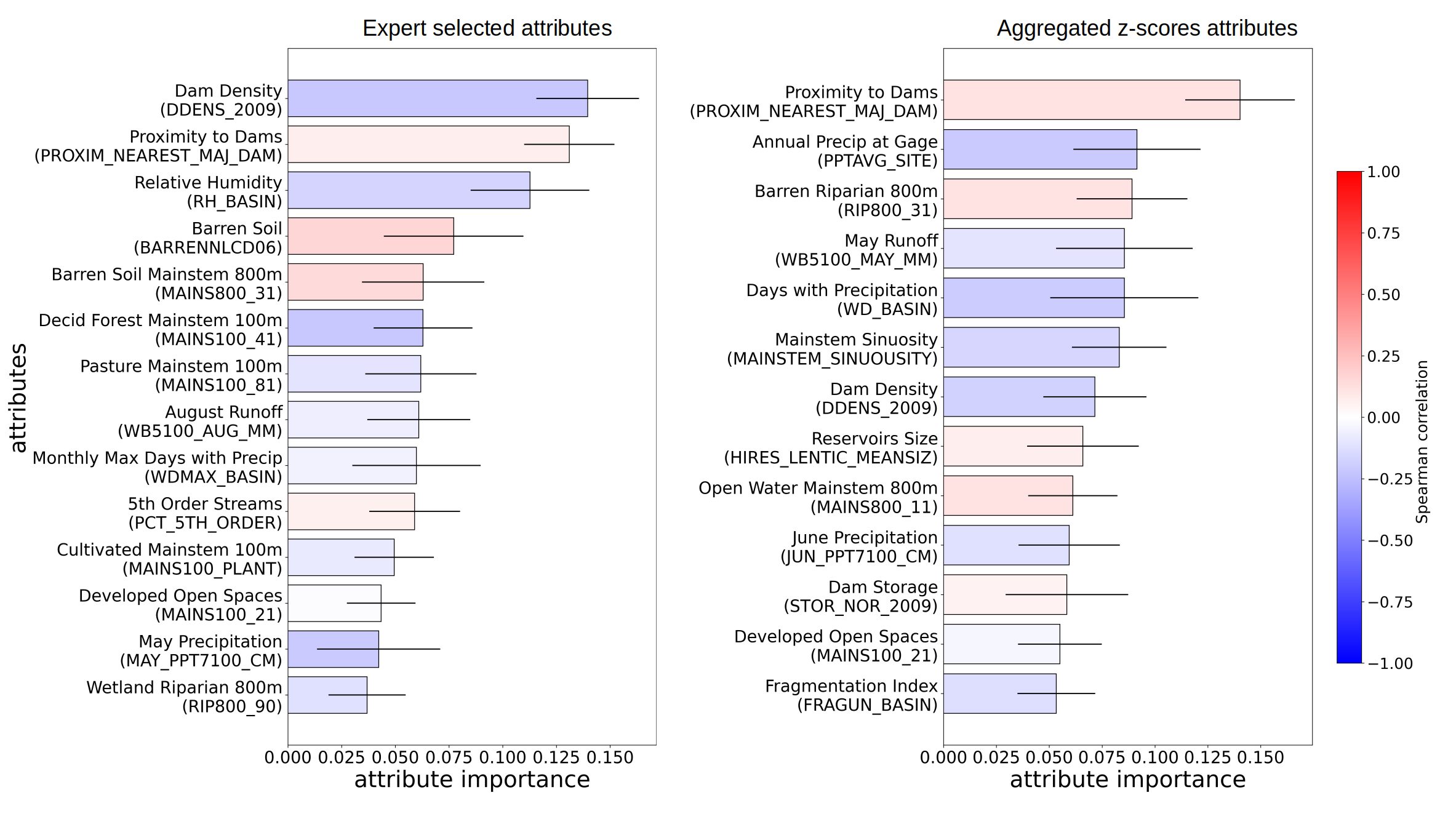}
    \caption{Plots of the average importance computed as Shapley values for the XGBoost models in Experiment 3 with the expert selected attributes (left) and aggregated z-scores attributes (right) following a recursive feature elimination. The color of the bars represents the Spearman correlation between each attribute and the RMSE values. Black error bars are the standard deviation of 100 realizations of the model.}
    \label{fig:exp3_rah_zscores}
\end{figure}

\clearpage
\section*{Supplementary Tables} 

\begin{table}
\centering
\fontsize{9}{11}\selectfont
\label{tab:metafeat}
\begin{tabular}{|p{0.4\linewidth}|p{0.15\linewidth}|p{0.4\linewidth}|}
\toprule
\textbf{Meta Feature} & \textbf{Units} & \textbf{Additional Description} \\
\midrule
\thead{\textbf{Attribute-based (Source/Target Differences)}} & & \\
Difference in Latitude & degrees & From USGS \\
Difference in Longitude & degrees & From USGS \\
Difference in Altitude & m & From USGS \\
Difference in DRAIN\_SQKM & km\(^2\) & Drainage area from GAGES-II \\
... continued for all 274 GAGES-II attributes. See \citeA{falcone_gages-ii_2011} for list & ... & ... \\
... & ... & ... \\
\midrule
\textbf{Observation Statistics (Source Only)} & & \\
Total Temperature Obs & samples & \\
Total Sampling Dates & days/profiles & \\
Number of Winter Obs & samples & Defined between solstice/equinox \\
Number of Spring Obs & samples & Defined between solstice/equinox \\
Number of Summer Obs & samples & Defined between solstice/equinox \\
Number of Autumn Obs & samples & Defined between solstice/equinox \\
Mean Temp of Obs & \(^\circ\)C & \\
5\% quantile Obs temp & \(^\circ\)C & \\
25\% quantile Obs temp & \(^\circ\)C & \\
75\% quantile Obs temp & \(^\circ\)C & \\
95\% quantile Obs temp & \(^\circ\)C & \\
Minimum Obs temp & \(^\circ\)C & \\
Maximum Obs temp & \(^\circ\)C & \\
Standard Deviation of Obs Temp & \(^\circ\)C & \\
Skew of Obs Temp & unitless & \\
Kurtosis of Obs Temp & unitless & \\
\midrule
\thead{\textbf{Meteorological (Source/Target Differences)}} & & \\
Solar Radiation (shortwave) & W/m\(^2\) & 2 Values: Difference in Mean/Std\_dev \\
Minimum Air Temp & \(^\circ\)C & 2 Values: Difference in Mean/Std\_dev \\
Mean Air Temp & \(^\circ\)C & 2 Values: Difference in Mean/Std\_dev \\
Max Air Temp & \(^\circ\)C & 2 Values: Difference in Mean/Std\_dev \\
Vapor Pressure & Pa & 2 Values: Difference in Mean/Std\_dev \\
Precipitation & mm/day & 2 Values: Difference in Mean/Std\_dev \\
Snow Water Equivalent & kg/m\(^2\) & 2 Values: Difference in Mean/Std\_dev \\
Log-transformed streamflow & ft\(^3\)/second & 2 Values: Difference in Mean/Std\_dev \\
\bottomrule
\end{tabular}
\caption{Candidate Meta Features for the MTL Metamodel.}
\end{table}

\begin{table}
\centering
\fontsize{8}{9.6}\selectfont
\label{tab:all_selected_metafeats}
\begin{tabular}{|p{0.3\linewidth}|p{0.7\linewidth}|}
\hline
\textbf{Meta Feature Category} & \textbf{Selected Meta Features (Differences)} \\
\hline
\textbf{Attribute-based (Source/Target Differences)} & 
Latitude, Longitude, DRAIN\_SQKM, STREAMS\_KM\_SQ\_KM, STOR\_NID\_2009, FORESTNLCD06, PLANTNLCD06, SLOPE\_PCT, RAW\_DIS\_NEAREST\_MAJ\_DAM, PERDUN, RAW\_DIS\_NEAREST\_DAM, RAW\_AVG\_DIS\_ALL\_MAJ\_DAMS, T\_MIN\_BASIN, T\_MAX\_BASIN, T\_MINSTD\_BASIN, RH\_BASIN, PPTAVG\_BASIN, HIRES\_LENTIC\_PCT, NDAMS\_2009, ELEV\_MEAN\_M\_BASIN, PPTAVG\_SITE, T\_AVG\_SITE, T\_MAX\_SITE, T\_MIN\_SITE, RH\_SITE, FST32F\_BASIN, LST32F\_BASIN, FST32SITE, WD\_BASIN, WD\_SITE, WDMAX\_BASIN, WDMIN\_BASIN, WDMAX\_SITE, WDMIN\_SITE, PET, SNOW\_PCT\_PRECIP, PRECIP\_SEAS\_IND, FEB\_PPT7100\_CM, MAY\_PPT7100\_CM, JUN\_PPT7100\_CM, SEP\_PPT7100\_CM, OCT\_PPT7100\_CM, NOV\_PPT7100\_CM, DEC\_PPT7100\_CM, JAN\_TMP7100\_DEGC, FEB\_TMP7100\_DEGC, MAR\_TMP7100\_DEGC, APR\_TMP7100\_DEGC, MAY\_TMP7100\_DEGC, JUN\_TMP7100\_DEGC, JUL\_TMP7100\_DEGC, SEP\_TMP7100\_DEGC, NOV\_TMP7100\_DEGC, DEC\_TMP7100\_DEGC, DDENS\_2009, STOR\_NOR\_2009, MAJ\_NDAMS\_2009, MAJ\_DDENS\_2009, CANALS\_PCT, RAW\_DIS\_NEAREST\_CANAL, RAW\_AVG\_DIS\_ALL\_CANALS, CANALS\_MAINSTEM\_PCT, NPDES\_MAJ\_DENS, RAW\_DIS\_NEAREST\_MAJ\_NPDES, RAW\_AVG\_DIS\_ALL\_MAJ\_NPDES, MINING92\_PCT, PCT\_IRRIG\_AG, POWER\_NUM\_PTS, POWER\_SUM\_MW, STRAHLER\_MAX, MAINSTEM\_SINUOUSITY, ARTIFPATH\_PCT, ARTIFPATH\_MAINSTEM\_PCT, BFI\_AVE, TOPWET, CONTACT, RUNAVE7100, WB5100\_JAN\_MM, WB5100\_FEB\_MM, WB5100\_MAR\_MM, WB5100\_APR\_MM, WB5100\_MAY\_MM, WB5100\_JUN\_MM, WB5100\_JUL\_MM, WB5100\_AUG\_MM, WB5100\_OCT\_MM, WB5100\_NOV\_MM, WB5100\_DEC\_MM, PCT\_2ND\_ORDER, PCT\_4TH\_ORDER, PCT\_5TH\_ORDER, PCT\_6TH\_ORDER\_OR\_MORE, FRAGUN\_BASIN, HIRES\_LENTIC\_NUM, HIRES\_LENTIC\_MEANSIZ, DEVNLCD06, WATERNLCD06, SNOWICENLCD06, DEVOPENNLCD06, DEVMEDNLCD06, BARRENNLCD06, DECIDNLCD06, EVERGRNLCD06, MIXEDFORNLCD06, SHRUBNLCD06, PASTURENLCD06, WOODYWETNLCD06, EMERGWETNLCD06, MAINS100\_11, MAINS100\_23, MAINS100\_31, MAINS100\_41, MAINS100\_42, MAINS100\_43, MAINS100\_52, MAINS100\_71, MAINS100\_81, MAINS100\_82, MAINS100\_95, MAINS800\_FOREST, MAINS800\_PLANT, MAINS800\_11, MAINS800\_12, MAINS800\_24, MAINS800\_31, MAINS800\_41, MAINS800\_42, MAINS800\_43, MAINS800\_52, MAINS800\_81, MAINS800\_90, MAINS800\_95, RIP100\_FOREST, RIP100\_PLANT, RIP100\_11, RIP100\_21, RIP100\_31, RIP100\_41, RIP100\_43, RIP100\_52, RIP100\_71, RIP100\_81, RIP100\_82, RIP100\_90, RIP100\_95, RIP800\_DEV, RIP800\_FOREST, RIP800\_PLANT, RIP800\_11, RIP800\_12, RIP800\_22, RIP800\_31, RIP800\_41, RIP800\_42, RIP800\_43, RIP800\_52, RIP800\_71, RIP800\_81, RIP800\_82, RIP800\_90, RIP800\_95, CDL\_CORN, CDL\_COTTON, CDL\_RICE, CDL\_SORGHUM, CDL\_SOYBEANS, CDL\_SUNFLOWERS, CDL\_BARLEY, CDL\_DURUM\_WHEAT, CDL\_SPRING\_WHEAT, CDL\_WINTER\_WHEAT, CDL\_WWHT\_SOY\_DBL\_CROP, CDL\_OATS, CDL\_ALFALFA, CDL\_OTHER\_HAYS, CDL\_DRY\_BEANS, CDL\_POTATOES, CDL\_FALLOW\_IDLE, CDL\_PASTURE\_GRASS, CDL\_ORANGES, CDL\_OTHER\_CROPS, CDL\_ALL\_OTHER\_LAND, NITR\_APP\_KG\_SQKM, PHOS\_APP\_KG\_SQKM, PESTAPP\_KG\_SQKM, PDEN\_2000\_BLOCK, PDEN\_DAY\_LANDSCAN\_2007, PDEN\_NIGHT\_LANDSCAN\_2007, NLCD01\_06\_DEV, PADCAT1\_PCT\_BASIN, PADCAT2\_PCT\_BASIN, PADCAT3\_PCT\_BASIN, HGA, HGAD, HGBD, HGCD, WTDEPAVE, ROCKDEPAVE, NO200AVE, SILTAVE, KFACT\_UP, RFACT, ELEV\_MAX\_M\_BASIN, ELEV\_MIN\_M\_BASIN, ELEV\_MEDIAN\_M\_BASIN, ELEV\_STD\_M\_BASIN, ELEV\_SITE\_M \\
\hline
\textbf{Observation Statistics (Source Only)} & 
Temperature Obs, Number of Winter Obs, Number of Spring Obs, Number of Summer Obs, Number of Autumn Obs, Mean Temp of Obs, Standard Deviation of Obs Temp, Skew of Obs Temp, Kurtosis of Obs Temp \\
\hline
\textbf{Meteorological Statistics (Source/Target Differences)} & 
Solar Radiation Mean+Stddev, Min Air Temp Mean+Stddev, Max Air Temp Mean+Stddev, Vapor Pressure Mean+Stddev, Precipitation Mean+Stddev, Snow Water Equivalent Mean+Stddev, Log-transformed Streamflow Mean+Stddev\\
\hline
\end{tabular}
\caption{Selected Meta Features for the MTL Metamodel chosen using RFECV}
\end{table}

\begin{table}[h]
\centering
\label{tab:stream_site_attributes}
\begin{tabular}{l l l}
\hline
\textbf{Feature Abbreviation} & \textbf{Description} & \textbf{Units} \\
\hline
DRAIN\_SQKM & Watershed drainage area & sq km \\
ELEV\_MEAN\_M\_BASIN & Mean watershed elevation & m \\
FORESTNLCD06 & Watershed percent forest & \% \\
HIRES\_LENTIC\_PCT & Percent of watershed area covered by lake/pond/reservoir & \% \\
NDAMS\_2009 & Number of dams in watershed & count \\
PERDUN & Dunne overland flow as percentage of total streamflow & \% \\
PLANTNLCD06 & Watershed percent agriculture (plant) & \% \\
PPTAVG\_BASIN & Mean annual precipitation for watershed (1971-2000) & mm \\
RAW\_AVG\_DIS\_ALL\_MAJ\_DAMS & Avg. distance to all major dams in watershed & m \\
RAW\_AVG\_DIS\_ALLDAMS & Avg. distance to all dams in watershed & m \\
RAW\_DIS\_NEAREST\_DAM & Distance to nearest dam in watershed & m \\
RAW\_DIS\_NEAREST\_MAJ\_DAM & Distance to nearest major dam in watershed & m \\
RH\_BASIN & Watershed average relative humidity & \% \\
SLOPE\_PCT & Mean watershed slope & \% \\
STOR\_NID\_2009 & Dam storage in watershed per area & m$^3$/sq km \\
STREAMS\_KM\_SQ\_KM & Stream density (length per area) & km/sq km \\
T\_AVG\_BASIN & Average annual temperature from 1971-2000 \\
T\_MAX\_BASIN & Avg. monthly max temperature (1971-2000) & °C \\
T\_MAXSTD\_BASIN & Std. dev. of monthly max temperature (1971-2000) & °C \\
T\_MIN\_BASIN & Avg. monthly min temperature (1971-2000) & °C \\
T\_MINSTD\_BASIN & Std. dev. of monthly min temperature (1971-2000) & °C \\
\hline
\end{tabular}
\caption{List of stream site attributes from \protect\citeA{rahmani2021deep} with descriptions}
\end{table}

\begin{table}[htbp]
    \centering
    \begin{tabular}{|l|}
    \hline
    \textbf{GAGES-II Aggregated Attributes} \\
    \hline
    Developed Areas \\
    Precipitation and Runoff \\
    Temperature \\
    Croplands \\
    Croplands and Canals \\
    Croplands and Dams \\
    Barren Soil and Deciduous Forests \\
    Elevation \\
    Evergreen Forests \\
    Woody Wetlands and Croplands \\
    Lakes, Ponds and Reservoirs \\
    Pastures and Grasslands \\
    Fine Soils \\
    Major Dams \\
    Summer Precipitation \\
    Herbaceous Wetlands \\
    Mixed Forests \\
    Coarse Soils \\
    Perennial Ice and Snow \\
    Shrublands \\
    Lower Order Streams \\
    Higher Order Streams \\
    Non Croplands \\
    Overland Flow \\
    Bulk Density \\
    \hline
    \end{tabular}
        \caption{25 aggregated z-score attribute categories used as a condensed representation of the 274 GAGES-II attributes as described in \cite{ciulla2023network}.}
    \label{tab:zscore_attr}
\end{table}

\begin{table}[h]
\centering
\begin{tabular}{|l l l l|}
\hline
Parameter      & Distribution          & Min   & Max   \\ \hline
batch\_size    & int\_uniform          & 50    & 1200  \\ \hline
n\_hidden      & int\_uniform          & 60   & 800   \\ \hline
num\_layers    & int\_uniform          & 1     & 8   \\ \hline
weight\_decay  & log\_uniform\_values  & 1e-7  & 1e-3  \\ \hline
dropout        & uniform               & 0.0   & 0.3   \\ \hline
\end{tabular}
\caption{Hyperparameter search ranges for LSTM}
\label{tab:my_label}
\end{table}

\begin{table}[h]
\centering
\begin{tabular}{|c c c c c|}
\hline
batch\_size & n\_hidden & num\_layers & weight\_decay & dropout \\ \hline
496         & 570       & 4           & 4.128e-06     & 2.216e-05 \\ \hline
236         & 594       & 3           & 4.753e-07     & 0.0138    \\ \hline
521         & 699       & 5           & 2.974e-07     & 0.0883    \\ \hline
201         & 760       & 3           & 9.684e-06     & 0.0518    \\ \hline
489         & 764       & 5           & 2.005e-04     & 0.0145    \\ \hline
\end{tabular}
\caption{Top 5 Hyperparameter Combinations for LSTM}
\label{tab:hyperparameters}
\end{table}

\begin{table}[h]
\centering
\begin{tabular}{|p{.25\linewidth} p{.22\linewidth} p{.16\linewidth} p{.16\linewidth}|} 
 \hline
 Method & RMSE ($^{\circ}$C) (std\_dev) & Mean Bias ($^{\circ}$C) & RMSE (warmest 10\%) ($^{\circ}$C)\\ [0.5ex] 
 \hline
 \textbf{LSTM\_conus} & \textbf{1.71}($\pm$0.98)  & -0.12($\pm$1.22) & \textbf{2.91($\pm$1.58))}\\ \hline
 LSTM\_regional & 2.53($\pm$1.78) & 0.00($\pm$1.76) & 4.77($\pm$2.90) \\ \hline
 LSTM\_cluster & 2.50($\pm$1.66) & 0.17($\pm$1.80) & 4.04($\pm$2.37) \\ \hline
 MTL & 2.23($\pm$1.07) & -0.55($\pm$1.57) & 3.73($\pm$1.75) \\ 
 \hline
\end{tabular}
\caption{Experiment 1 Overall RMSE as mean values for the top-down \textit{LSTM\_conus} model, the grouped \textit{LSTM\_regional} and \textit{LSTM\_cluster} models, and the bottom-up \textit{MTL} model to compare against "per-site medians" shown in the results section. }
\label{tab:exp1_overall_mean}
\end{table}

\begin{table}[!ht]
\centering
\scalebox{0.8}{
\begin{tabular}{|l l p{1.5cm} l l l l|}
\hline
Region & n\_obs (train/test) & n\_sites (train/test) & LSTM\_conus & LSTM\_regional & LSTM\_cluster & MTL \\ \hline
01 (New England) & 101,768/9,784 & 10/23 & \textbf{1.31($\pm$0.74)} & 2.89($\pm$2.37) & 1.93($\pm$0.96) & 2.18($\pm$0.77) \\ \hline
02 (Mid-Atlantic) & 477,381/88,224 & 81/121 & \textbf{1.54($\pm$0.88)} & 1.94($\pm$1.17) & 2.34($\pm$1.73) & 2.05($\pm$1.02) \\ \hline
03 (South Atlantic-Gulf) & 537,631/64,533 & 97/77 & \textbf{1.60($\pm$0.81)} & 2.08($\pm$1.27) & 2.40($\pm$1.44) & 1.88($\pm$0.79) \\ \hline
04 (Great Lakes) & 282,868/18,522 & 58/22 & \textbf{1.09($\pm$0.29)} & 1.56($\pm$0.78) & 2.37($\pm$2.04) & 1.53($\pm$0.41) \\ \hline
05 (Ohio) & 273,588/28,431 & 66/41 & \textbf{1.57($\pm$0.83)} & 2.70($\pm$1.75) & 2.10($\pm$1.13) & 2.40($\pm$0.91) \\ \hline
06 (Tennessee) & 48,359/3,174 & 15/5 & \textbf{0.89($\pm$0.35)} & 2.27($\pm$1.64) & 1.24($\pm$0.70) & 1.73($\pm$0.50) \\ \hline
07 (Upper Mississippi) & 122,889/28,053 & 32/29 & \textbf{1.92($\pm$0.83)} & 3.02($\pm$1.68) & 2.39($\pm$1.09) & 2.17($\pm$0.87) \\ \hline
08 (Lower Mississippi) & 18,467/3,813 & 5/5 & \textbf{2.34($\pm$1.40)} & 7.30($\pm$1.75) & 2.98($\pm$1.07) & 2.40($\pm$1.57) \\ \hline
09 (Souris-Red-Rainy) & 19,440/4,188 & 4/5 & 1.98($\pm$0.98) & 8.97($\pm$1.30) & 6.18($\pm$3.65) & \textbf{1.75($\pm$0.64)} \\ \hline
10 (Missouri) & 220,070/46,083 & 52/54 & \textbf{1.71($\pm$0.91)} & 2.82($\pm$2.17) & 2.40($\pm$1.29) & 2.26($\pm$0.97) \\ \hline
11 (Arkansas-White-Red) & 322,319/21,989 & 56/23 & \textbf{1.62($\pm$0.51)} & 2.85($\pm$1.36) & 3.33($\pm$2.49) & 2.18($\pm$0.50) \\ \hline
12 (Texas-Gulf) & 172,894/15,377 & 26/16 & \textbf{2.03($\pm$1.28)} & 2.31($\pm$1.21) & 2.93($\pm$1.72) & 2.93($\pm$1.14) \\ \hline
13 (Rio Grande) & 14,533/0 & 3/0 & N/A & N/A & N/A & N/A \\ \hline
14 (Upper Colorado) & 234,865/27,723 & 47/31 & \textbf{2.15($\pm$0.89)} & 2.83($\pm$1.36) & 2.26($\pm$1.14) & 2.53($\pm$1.10) \\ \hline
15 (Lower Colorado) & 5,609/5,742 & 2/7 & \textbf{3.14($\pm$3.07)} & 6.84($\pm$1.68) & 4.07($\pm$2.45) & 4.96($\pm$3.87) \\ \hline
16 (Great Basin) & 115,347/17,914 & 25/20 & \textbf{1.84($\pm$1.03)} & 3.29($\pm$1.92) & 2.21($\pm$1.27) & 2.59($\pm$0.73) \\ \hline
17 (Pacific Northwest) & 901,098/67,182 & 142/70 & \textbf{1.56($\pm$0.89)} & 1.79($\pm$1.17) & 1.97($\pm$1.30) & 2.32($\pm$0.96) \\ \hline
18 (California) & 138,220/36,043 & 54/44 & \textbf{2.39($\pm$1.06)} & 3.15($\pm$1.41) & 3.93($\pm$1.57) & 2.47($\pm$0.98) \\ \hline
\end{tabular}}
\caption{Experiment 1 regional RMSE as mean values per USGS-defined hydrological region (\url{https://water.usgs.gov/GIS/regions.html}) to compare against "per-site medians" shown in the results section. Values are the RMSE($^{\circ}$C) across sites in a region with the RMSE standard deviation in parenthesis. Lowest RMSE values per region are shown in bold.}
\label{tab:regional_results_mean_rmse}
\end{table}

\begin{table}[!ht]
    \centering
    \scalebox{1}{
    \begin{tabular}{|l l p{1.5cm} l l l l|}
        \hline
        Cluster & n\_obs (train/test) & n\_sites (train/test) & LSTM\_conus & LSTM\_regional & LSTM\_cluster & MTL \\ \hline
        0 & 882,382/105,762 & 183/109 & \textbf{1.94($\pm$0.98)} & 2.76($\pm$1.64) & 2.15($\pm$1.09) & 2.45($\pm$0.99) \\ \hline
        1 & 182,149/46,399 & 46/45 & \textbf{1.68($\pm$0.74)} & 2.63($\pm$1.70) & 2.48($\pm$1.76) & 2.00($\pm$0.89) \\ \hline
        2 & 152,682/54,531 & 38/59 & \textbf{1.81($\pm$0.86)} & 2.79($\pm$1.87) & 3.08($\pm$1.72) & 1.94($\pm$0.83) \\ \hline
        3 & 476,469/41,125 & 89/47 & \textbf{1.54($\pm$0.96)} & 2.13($\pm$1.53) & 1.83($\pm$1.33) & 2.11($\pm$0.96) \\ \hline
        4 & 193,885/59,760 & 40/73 & \textbf{1.35($\pm$0.68)} & 1.98($\pm$1.26) & 2.28($\pm$1.56) & 1.80($\pm$0.62) \\ \hline
        5 & 159,958/30,618 & 38/38 & \textbf{2.35($\pm$1.61)} & 3.49($\pm$1.99) & 4.09($\pm$1.77) & 2.92($\pm$2.07) \\ \hline
        6 & 662,196/47,206 & 99/44 & \textbf{1.50($\pm$0.86)} & 1.68($\pm$1.08) & 1.68($\pm$0.93) & 2.32($\pm$0.96) \\ \hline
        7 & 315,696/27,276 & 56/33 & \textbf{1.45($\pm$1.02)} & 2.21($\pm$1.93) & 1.86($\pm$1.38) & 2.22($\pm$1.41) \\ \hline
        8 & 246,997/42,897 & 57/51 & \textbf{1.63($\pm$0.97)} & 2.24($\pm$1.43) & 2.17($\pm$1.38) & 2.21($\pm$0.91) \\ \hline
        9 & 98,175/21,157 & 22/23 & \textbf{1.82($\pm$0.69)} & 2.99($\pm$2.39) & 2.98($\pm$1.52) & 2.23($\pm$0.57) \\ \hline
        10 & 187,683/10,136 & 34/12 & \textbf{1.48($\pm$0.75)} & 3.25($\pm$2.87) & 4.30($\pm$2.58) & 1.85($\pm$0.52) \\ \hline
        11 & 206,921/23,653 & 34/22 & \textbf{1.52($\pm$0.79)} & 2.51($\pm$2.22) & 2.34($\pm$1.37) & 2.66($\pm$0.99) \\ \hline
        12 & 49,148/9,582 & 9/9 & \textbf{2.33($\pm$1.18)} & 2.49($\pm$0.82) & 4.70($\pm$2.72) & 2.38($\pm$0.70) \\ \hline
        13 & 92,009/9,525 & 21/8 & \textbf{1.92($\pm$0.92)} & 2.95($\pm$1.41) & 3.10($\pm$1.31) & 2.67($\pm$1.14) \\ \hline
        14 & 22,750/5,491 & 6/5 & \textbf{2.00($\pm$0.95)} & 6.79($\pm$2.01) & 5.19($\pm$2.34) & 2.11($\pm$0.95) \\ \hline
    \end{tabular}}
    \caption{Mean RMSE statistics per cluster. Values are the mean RMSE($^\circ$C) $\pm$ standard deviation across sites in a cluster, with mean RMSE standard deviation in parentheses. Lowest RMSE values per cluster are shown in bold.}
\label{tab:cluster_results_mean_rmse}
\end{table}

\begin{table}[h!]
\centering
    \scalebox{0.85}{
\begin{tabular}{|>{\centering\arraybackslash}p{.10\linewidth} 
                 >{\centering\arraybackslash}p{.11\linewidth}
                 >{\centering\arraybackslash}p{.07\linewidth}
                 >{\centering\arraybackslash}p{.10\linewidth} |
                 p{.15\linewidth} 
                 p{.13\linewidth} 
                 p{.14\linewidth} 
                 p{.13\linewidth}||} 
 \hline
 \multicolumn{4}{|c|}{Included Inputs} &  &  & &  \\ [0.5ex] 
  
 \hline
 Meteorology & Latitude, Longitude, Elevation & Discharge & GAGES attributes & n\_obs/n\_sites (training) & RMSE (std\_dev) ($^{\circ}$C) &  Bias (std\_dev) ($^{\circ}$C) & RMSE (std\_dev) Warmest 10\% ($^{\circ}$C) \\ [0.5ex] 
 \hline\hline
 \checkmark & &  & & 3.82mil/782 & 2.01($\pm$1.04) & -0.23($\pm$1.36) & 3.14($\pm$1.68) \\ 
 \rowcolor{gray!40}\checkmark & &  & & 6.13mil/1346 & 2.02($\pm$1.00) & -0.25($\pm$1.32) & 3.18($\pm$1.62) \\ 
 \hline\hline
 \checkmark & \checkmark & & & 3.82mil/782 & 2.09($\pm$1.19) & -0.23($\pm$1.40) & 3.38($\pm$1.97) \\
  \rowcolor{gray!40}\checkmark & \checkmark & & & 6.13mil/1346 & 2.01($\pm$1.02) & -0.05($\pm$1.29) & 2.89($\pm$1.69) \\

 \hline\hline
  \checkmark & \checkmark & \checkmark & & 3.82mil/782 & 1.84($\pm$1.10) & -0.20($\pm$1.29) & 3.02($\pm$1.79)\\
  \rowcolor{gray!40}\checkmark & \checkmark & \checkmark & & 4.65mil/1048 & 1.80($\pm$1.11) & -0.22($\pm$1.28) & 3.03($\pm$1.82)\\
 \hline\hline
  \checkmark & \checkmark & & \checkmark & 3.82mil/782 & 1.76($\pm$1.03) & -0.16($\pm$1.18) & 3.05($\pm$1.65)\\
  \rowcolor{gray!40}\checkmark & \checkmark & & \checkmark & 4.13mil/833 & \textbf{1.74($\pm$1.05)} & -0.20($\pm$1.18) & 3.13($\pm$1.69)\\
 \hline\hline
 \checkmark & \checkmark & \checkmark & \checkmark & 3.82mil/782 & 1.72($\pm$1.08)  & -0.12($\pm$1.29) & \textbf{2.91($\pm$1.79)}\\
 \hline
\end{tabular}}
\caption{Performance of continental-scale models in Experiment 2 with specific categories of inputs removed including GAGES-II attributes, discharge data, and location and elevation data. The final row is identical to the \textit{LSTM\_conus} model from Experiment 1. Rows with grey background indicate an extended training dataset with observations previously discarded due to lack of either discharge data or GAGES attributes. Values shown are mean RMSE to compare with the medians used in the results section.}
\label{tab:exp2_overall}
\end{table}

\begin{table}[h!]
\centering
\begin{tabular}{|p{.35\linewidth} p{.22\linewidth} p{.16\linewidth} p{.16\linewidth}|} 
 \hline
 Input Set & RMSE ($^{\circ}$C) (std\_dev) & Bias ($^{\circ}$C) & RMSE (warmest 10\%) ($^{\circ}$C)\\ [0.5ex] 
 \hline
  Default (\textit{LSTM\_conus}; 274 GAGES-II attributes)& 1.72($\pm$0.98) & -0.12($\pm$1.22) & 2.91($\pm$1.58)\\ \hline
  Expert-selected (21 GAGES-II attributes from \cite{rahmani2021deep}) & 1.74($\pm$1.08)  & -0.27($\pm$1.29) & 3.14($\pm$1.79)\\ \hline
 Z-score-based Attributes& \textbf{1.71($\pm$1.00)} & -0.14($\pm$1.21) & \textbf{2.87($\pm$1.66)}\\

 \hline
\end{tabular}

\caption{Experiment 3 Overall Results, overall RMSE as opposed to median per-site used in Results section.}
\label{tab:exp3_overall}
\end{table}
\begin{table}[h!]
\centering
\begin{tabular}{|p{.25\linewidth} p{.10\linewidth} p{.22\linewidth} p{.16\linewidth} p{.16\linewidth}|} 
 \hline
 Input Set & Ensemble size & Median per-site RMSE ($^{\circ}$C) (std\_dev) & Bias ($^{\circ}$C) & RMSE (warmest 10\%) ($^{\circ}$C)\\ [0.5ex] 
 \hline
 Default (\textit{LSTM\_conus}) & 5 & 1.43($\pm$0.98)  & -0.09($\pm$1.22) & 1.70($\pm$1.58) \\
  & 20 & \textbf{1.39($\pm$0.98)}  & -0.09($\pm$1.20) & 1.57($\pm$1.56) \\
 Z-score-based attributes & 5& 1.47($\pm$1.00)  & -0.09($\pm$1.21) & 1.60($\pm$1.55)\\
 & 20 & 1.46($\pm$0.99)  & -0.14($\pm$1.20) & 1.61($\pm$1.64)\\
 Expert-selected attributes & 5 &1.46($\pm$1.08) & -0.17($\pm$1.29) & 1.64($\pm$1.79)\\
   & 20 &\textbf{1.41($\pm$1.04)} & -0.15($\pm$1.25) & \textbf{1.65($\pm$1.75)}\\
 \hline
\end{tabular}
\caption{Experiment 3 Extended Comparison with Ensemble Sizes of 5 and 20 compared}
\label{tab:exp3_ensemble_size_compare}
\end{table}

\begin{table}[h!]
\centering
\begin{tabular}{|p{.25\linewidth}| p{.10\linewidth} p{.10\linewidth} p{.10\linewidth} p{.10\linewidth}
p{.10\linewidth} p{.10\linewidth}
|} 
 \hline
 \textbf{Air-Stream Class} & RMSE Expert & Bias Expert & RMSE Z-score & Bias Z-score & RMSE LSTM\_conus & Bias LSTM\_conus \\ 
 \hline
 \\
Atmospheric & 1.32 & -0.21 & 1.32 & -0.17 & 1.26 & -0.21 \\
Shallow Groundwater & 1.48 & 0.06 & 1.50 & -0.15 & 1.22 & 0.08 \\
Deep Groundwater & 2.01 & 0.40 & 2.06 & 0.69 & 2.00 & 0.56 \\
Dammed & 1.60 & -0.17 & 1.57 & -0.20 & 1.55 & -0.02 \\
 \hline
\end{tabular}
\caption{Median RMSE and Mean Bias metrics by air-stream temperature class for all test sites}
\label{tab:airstream_class_metrics}
\end{table}
\end{document}